\documentclass{article}
\usepackage[utf8]{inputenc}

\usepackage[T1]{fontenc}    
\usepackage[colorlinks,citecolor=blue,urlcolor=red,linkcolor=red,bookmarks=false]{hyperref}       
\usepackage{url}            

\usepackage{amsmath,amsfonts,bm}

\def\eqref#1{equation~\ref{#1}}

\def\1{\bm{1}}

\DeclareMathAlphabet{\mathsfit}{\encodingdefault}{\sfdefault}{m}{sl}
\SetMathAlphabet{\mathsfit}{bold}{\encodingdefault}{\sfdefault}{bx}{n}

\usepackage{hyperref}
\usepackage{url}
\usepackage{gensymb}
\usepackage{graphicx}
\usepackage{authblk} %
\usepackage{amsmath,amsfonts,bm}
\usepackage{rotating}
\usepackage[toc,page]{appendix}
\usepackage{pgf, tikz}
\usetikzlibrary{positioning}
\usetikzlibrary{arrows, automata}
\usepackage{caption}
\usepackage{subcaption}
\usetikzlibrary{shapes,snakes}
\usepackage{makecell}
\usepackage{multirow}
\usepackage{amssymb}
\usepackage{enumerate}
\usepackage{amsthm}
\usepackage[authoryear,round,longnamesfirst]{natbib}
\usepackage{adjustbox}
\usepackage[ruled,vlined]{algorithm2e}
\usepackage{todonotes}

\newcommand*\circled[1]{\tikz[baseline=(char.base)]{
            \node[shape=circle,draw,inner sep=1pt,solid] (char) {#1};}}

\makeatletter
\newcommand*{\centernot}{%
  \mathpalette\@centernot
}
\def\@centernot#1#2{%
  \mathrel{%
    \rlap{%
      \settowidth\dimen@{$\m@th#1{#2}$}%
      \kern.5\dimen@
      \settowidth\dimen@{$\m@th#1=$}%
      \kern-.5\dimen@
      $\m@th#1\not$%
    }%
    {#2}%
  }%
}

\DeclareMathOperator*{\vol}{\text{vol}}

\makeatother\newcommand{\indep}{\perp\mkern-9.5mu\perp}
\newcommand{\notindep}{\centernot{\indep}}

\newtheorem{model}{Model}

\newtheorem{thm}{Theorem}

\newtheorem{ppn}{Proposition}
\newtheorem{dfn}{Definition}
\newtheorem{asm}{Assumption}

\newcommand\blfootnote[1]{%
  \begingroup
  \renewcommand\thefootnote{}\footnote{#1}%
  \addtocounter{footnote}{-1}%
  \endgroup
}

\title{\bf Nonlinear Invariant Risk Minimization: \\ A Causal Approach}

\author[1,2]{Chaochao Lu}
\author[3,5]{Yuhuai Wu}
\author[1,4]{Jo{\'s}e Miguel Hern{\'a}ndez-Lobato$^*$}
\author[2]{Bernhard Sch{\"o}lkopf$^*$}
\affil[1]{\small University of Cambridge, Cambridge, UK}
\affil[2]{\small Max Planck Institute for Intelligent Systems, T{\"u}bingen, Germany}
\affil[3]{\small University of Toronto, Toronto, Canada}
\affil[4]{\small Alan Turing Institute, London, UK}
\affil[5]{\small Vector Institute, Toronto, Canada}

\date{}

\begin{document}

\maketitle

\begin{abstract}\blfootnote{$^*$Equal supervision.}
Due to spurious correlations, machine learning systems often fail to generalize to environments whose distributions differ from the ones used at training time. Prior work addressing this, either explicitly or implicitly, attempted to find a data representation that has an invariant relationship with the target. This is done by leveraging a diverse set of training environments to reduce the effect of spurious features and build an invariant predictor. However, these methods have generalization guarantees only when both data representation and classifiers come from a linear model class. We propose invariant Causal Representation Learning (iCaRL), an approach that enables out-of-distribution (OOD) generalization in the nonlinear setting (i.e., nonlinear representations and nonlinear classifiers). It builds upon a practical and general assumption: the prior over the data representation (i.e., a set of latent variables encoding the data) given the target and the environment belongs to general exponential family distributions, i.e., a more flexible conditionally non-factorized prior that can actually capture complicated dependences between the latent variables. Based on this, we show that it is possible to identify the data representation up to simple transformations. We also prove that all direct causes of the target can be fully discovered, which further enables us to obtain generalization guarantees in the nonlinear setting. Extensive experiments on both synthetic and real-world datasets show that our approach outperforms a variety of baseline methods. Finally, in the concluding discussion, we further explore the aforementioned general assumption and propose a more general hypothesis, called \textit{the Agnostic Hypothesis}: there exist a set of hidden causal factors affecting both inputs and outcomes. The Agnostic Hypothesis can provide a unifying view of machine learning, be it supervised, unsupervised, or reinforcement learning, in terms of representation learning. More importantly, it can inspire a new direction to explore a general theory for identifying hidden causal factors, which is key to enabling the OOD generalization guarantees in machine learning.
\end{abstract}


\section{Introduction}

Modern machine learning algorithms still lack robustness, and may fail to generalize outside of a specific training distribution because they learn easy-to-fit spurious correlations which are prone to change between training and testing environments. We recall the widely used example of classifying images of camels and cows \citep{beery2018recognition}. Here, the training dataset has a selection bias, i.e., many pictures of cows are taken on green pastures, while most pictures of camels happen to be in deserts. After training, it is found that the model builds on spurious correlations, i.e., it relates green pastures with cows and deserts with camels, and fails to recognize images of cows on the beach. 
 
To address this problem, a natural idea is to identify which features of the training data present domain-varying spurious correlations with labels and which features describe true correlations of interest that are stable across domains. In the example above, the former are the features describing the context (e.g., pastures and deserts), whilst the latter are the features describing the animals (e.g., animal shape).
By exploiting the varying degrees of spurious correlation naturally present in training data collected from multiple environments, one can try to identify stable features and build invariant predictors.
Invariant risk minimization (IRM) seeks to find data representations \citep{arjovsky2019invariant} or features \citep{RojSchTurPet18} for which the optimal predictor is invariant across all environments. The general formulation of IRM is a challenging bi-leveled optimization problem, and theoretical guarantees require constraining both data representations and classifiers to be linear \citep[Theorem 9]{arjovsky2019invariant}, or considering the special case of feature selection \citep[Theorem 4]{RojSchTurPet18}.
\citet{ahuja2020invariant} study the problem from the perspective of game theory, with an approach termed invariant risk minimization games (IRMG). They show that the set of Nash equilibria for a proposed game is equivalent to the set of invariant predictors for any finite number of environments, even with nonlinear data representations and nonlinear classifiers. However, these theoretical results in the nonlinear setting only guarantee that one can learn invariant predictors from training environments, but do not guarantee that the learned invariant predictors can generalize well across all environments including unseen testing environments.

We propose invariant Causal Representation Learning (iCaRL), a novel approach that enables out-of-distribution (OOD) generalization in the nonlinear setting (i.e., nonlinear representations and nonlinear classifiers\footnote{In fact, we are not restricted to the classification case and allow the target to be either continuous or categorical, which will be formally defined in Section \ref{sect:irm}.}). We achieve this by extending and using methods from representation learning and graphical causal discovery.
In more detail, we first introduce our main \emph{general assumption}: when conditioning on the target (e.g., labels) and the environment (represented as an index), the prior over the data representation (i.e., a set of latent variables encoding the data) belongs to a \emph{general exponential family}. Unlike the conditionally factorized prior assumed in recent identifiable variational autoencoders (iVAE) \citep{khemakhem2020variational}, this is a more flexible conditionally non-factorized prior, which can actually capture complicated dependences between the latent variables. We then extend iVAE to the case in which the latent variable prior belongs to such a \emph{general exponential family}.
The combination of this result and the previous general assumption allows us to a guarantee that the data representation can be identified up to simple transformations. 
We then theoretically show that the direct causes of the target can be fully discovered by analyzing all possible graphs in a structural equation model setting. Once they are discovered, the challenging bi-leveled optimization problem in IRM and IRMG can be reduced to two simpler independent optimization problems, that is, learning the data representation and learning the optimal classifier can be performed separately. This leads to a practical algorithm and enables us to obtain generalization guarantees in the nonlinear setting.  

It is worth noting that in the concluding discussion, we further explore the assumption introduced above and propose a general hypothesis, called \textit{the Agnostic Hypothesis}: there exist a set of hidden causal factors affecting both inputs and outcomes. The Agnostic Hypothesis can provide a unifying view of machine learning, be it supervised, unsupervised, or reinforcement learning, in terms of representation learning. More importantly, it can inspire a new direction for achieving a general theory for hidden causal factor identification, which is key to enabling OOD generalization guarantees in machine learning.

Overall, we make a number of key contributions: (1) We propose a general framework for out-of-distribution generalization in the nonlinear setting with the theoretical guarantees on both identifiability and generalizability; (2) We propose a general assumption on the underlying causal diagram for prediction (Assumption \ref{eq:assumption1} and Fig. \ref{fig:general}), which covers many real-world scenarios (Section \ref{sect:asm_graph}); (3) We propose a general assumption on the prior over the latent variables (Assumption \ref{eq:assumption2}), i.e., a more flexible conditionally non-factorized prior; (4) We prove that an extended iVAE with this conditionally non-factorized prior is also identifiable (Theorems \ref{thm:1}, \ref{thm:2}\&\ref{thm:3}); (5) We prove that our framework has the theoretical guarantees for OOD  generalization in the nonlinear setting (Proposition \ref{prop:2}); (6) We propose the Agnostic Hypothesis which provides a unifying view of machine learning.


\section{Preliminaries}

\subsection{Identifiable Variational Autoencoders} \label{sect:ivae}

Variational autoencoders (VAEs, see Appendix \ref{appendix:vae}) \citep{kingma2013auto,rezende2014stochastic} lack identifiability guarantees. Consider a VAE model where $\boldsymbol{O} \in \mathbb{R}^d$ stands for the observed variables (data) and $\boldsymbol{X} \in \mathbb{R}^n$ for the latent variables. \citet{khemakhem2020variational} show that a VAE with an unconditional prior distribution $p_{\boldsymbol{\theta}}(\boldsymbol{X})$ over the latent variables is unidentifiable. However, they also show that it is possible to obtain an identifiable model if one posits a conditionally factorized prior distribution over the latent variables, $p_{\boldsymbol{\theta}}(\boldsymbol{X}|\boldsymbol{U})$, where $\boldsymbol{U} \in \mathbb{R}^m$ is an additional observed variable \citep{hyvarinen2019nonlinear}. Specifically, let $\boldsymbol{\theta}=(\boldsymbol{f}, \boldsymbol{T}, \boldsymbol{\lambda}) \in \Theta$ be the parameters of the conditional generative model
\begin{align} 
p_{\boldsymbol{\theta}}(\boldsymbol{O}, \boldsymbol{X}|\boldsymbol{U}) = p_{\boldsymbol{f}}(\boldsymbol{O}|\boldsymbol{X})p_{\boldsymbol{T}, \boldsymbol{\lambda}}(\boldsymbol{X}|\boldsymbol{U}),
\label{eq:ivae_gen}
\end{align} 
where $p_{\boldsymbol{f}}(\boldsymbol{O}|\boldsymbol{X})=p_{\boldsymbol{\epsilon}}(\boldsymbol{O}-\boldsymbol{f}(\boldsymbol{X}))$ in which $\boldsymbol{\epsilon}$ is an independent noise variable with probability density function $p_{\boldsymbol{\epsilon}}(\boldsymbol{\epsilon})$. Importantly, the prior $p_{\boldsymbol{T}, \boldsymbol{\lambda}}(\boldsymbol{X}|\boldsymbol{U})$ is assumed to be conditionally factorial, where each element of $X_i \in \boldsymbol{X}$ has a univariate exponential family distribution given $\boldsymbol{U}$. The conditioning on $\boldsymbol{U}$ is through an arbitrary function $\boldsymbol{\lambda}(\boldsymbol{U})$ (e.g., a neural net) that outputs the individual exponential family parameters $\boldsymbol{\lambda}_i(\boldsymbol{U})$ for each $X_i$. The prior probability density thus takes the form
\begin{align} 
p_{\boldsymbol{T}, \boldsymbol{\lambda}}(\boldsymbol{X}|\boldsymbol{U}) = \prod\nolimits_i {\mathcal{Q}_i(X_i)}/{\mathcal{Z}_i(\boldsymbol{U})} \exp\Big[\sum\nolimits_{j=1}^k T_{i,j}(X_i)\lambda_{i,j}(\boldsymbol{U})\Big], \label{eq:ivae_prior}
\end{align}
where $\mathcal{Q}_i$ is the base measure, $X_i$ the $i$-th dimension of $\boldsymbol{X}$, $\mathcal{Z}_i(\boldsymbol{U})$ the normalizing constant, $\boldsymbol{T}_i=(T_{i,1},\ldots, T_{i, k})$ the sufficient statistics, $\boldsymbol{\lambda}_i(\boldsymbol{U})=(\lambda_{i,1}(\boldsymbol{U}),\ldots, \allowbreak \lambda_{i, k}(\boldsymbol{U}))$ the corresponding natural parameters depending on $\boldsymbol{U}$, and $k$ the dimension of each sufficient statistic that is fixed in advance. It is worth noting that this prior is restrictive as it is factorial and therefore cannot capture dependencies. As in VAEs, the model parameters are estimated by maximizing the corresponding evidence lower bound,
\begin{multline} 
\mathcal{L}_{\text{iVAE}}(\boldsymbol{\theta}, \boldsymbol{\phi}) 
:= \mathbb{E}_{p_D}\big[\mathbb{E}_{q_{\boldsymbol{\phi}}(\boldsymbol{X}|\boldsymbol{O}, \boldsymbol{U})}\left[\log p_{\boldsymbol{f}}(\boldsymbol{O}|\boldsymbol{X})\right.\\ 
\left. + \log p_{\boldsymbol{T}, \boldsymbol{\lambda}}(\boldsymbol{X}|\boldsymbol{U}) -\log q_{\boldsymbol{\phi}}(\boldsymbol{X}|\boldsymbol{O}, \boldsymbol{U})\right]\big], \label{eq:ivae_loss}
\end{multline}
where we denote by $p_D$ the empirical data distribution given by the dataset $\mathcal{D}=\left\{\left(\boldsymbol{O}^{(i)}, \boldsymbol{U}^{(i)}\right)\right\}_{i=1}^N$ and
$q_{\boldsymbol{\phi}}(\boldsymbol{X}|\boldsymbol{O}, \boldsymbol{U})$ denotes an approximate conditional
distribution for $\boldsymbol{X}$  given
by a recognition network with parameters $\bm\phi$.
This approach is called identifiable VAE (iVAE). Most importantly, it can be proved that under the conditions stated in Theorem 2 of \citep{khemakhem2020variational}, iVAE can identify the latent variables $\boldsymbol{X}$ up to a permutation and a simple componentwise transformation, see Appendix \ref{appendix:identifiability}.

\subsection{Invariant Risk Minimization}
\label{sect:irm}

\citet{arjovsky2019invariant} introduced invariant risk minimization (IRM), whose goal is to construct an \textit{invariant predictor} $f$ that performs well across all environments $\mathcal{E}_{all}$ by exploiting data collected from multiple environments $\mathcal{E}_{tr}$, where $\mathcal{E}_{tr} \subseteq \mathcal{E}_{all}$. Technically, they consider datasets $D_e := \{(\boldsymbol{o}_i^e, \boldsymbol{y}_i^e)\}_{i=1}^{n_e}$ from multiple training environments $e \in \mathcal{E}_{tr}$, where $\boldsymbol{o}_i^e \in \mathcal{O} \subseteq \mathbb{R}^d$ is the input observation and its corresponding label is $\boldsymbol{y}_i^e \in \mathcal{Y} \subseteq \mathbb{R}^s$.\footnote{The setup applies to both continuous and categorical data. If any observation or label is categorical, we one-hot encode it.} The dataset $D_e$, collected from environment $e$, consists of examples identically and independently distributed according to some probability distribution $P(\boldsymbol{O}^e, \boldsymbol{Y}^e)$. The goal of IRM is to use these multiple datasets to learn a predictor $\boldsymbol{Y} = f(\boldsymbol{O})$ that performs well for all the environments. Here we define the risk reached by $f$ in environment $e$ as $R^e(f)=\mathbb{E}_{\boldsymbol{O}^e, \boldsymbol{Y}^e}\left[\ell(f(\boldsymbol{O}^e), \boldsymbol{Y}^e)\right]$, where $\ell(\cdot)$ is a loss function. Then, the invariant predictor can be formally defined as follows:
\begin{dfn}[Invariant Predictor \citep{arjovsky2019invariant}]
We say that a data representation $\Phi \in \mathcal{H}_{\Phi}: \mathcal{O} \rightarrow \mathcal{C}$ elicits an invariant predictor $w \circ \Phi$ across environments $\mathcal{E}$ if there is a classifier $w \in \mathcal{H}_{w}: \mathcal{C} \rightarrow \mathcal{Y}$ simultaneously optimal for all environments, that is, $w \in \arg \min_{\bar{w} \in \mathcal{H}_w} R^e(\bar{w} \circ \Phi)$ for all $e \in \mathcal{E}$, where $\circ$ means function composition. 
\end{dfn}

Mathematically, IRM can be phrased as the following constrained optimization problem:
\begin{align}\label{eq:irm}
\begin{split}
\underset{\Phi \in \mathcal{H}_{\Phi}, w \in \mathcal{H}_{w}}{\min} 
\sum\nolimits_{e \in \mathcal{E}_{tr}} R^e(w \circ \Phi) 
\quad\text{  s.t. } w \in \arg \min_{\bar{w} \in \mathcal{H}_w} R^e(\bar{w} \circ \Phi), \forall e \in \mathcal{E}_{tr}.
\end{split}
\end{align}
Since this is a generally infeasible bi-leveled optimization problem, \citet{arjovsky2019invariant} rephrased it as a tractable penalized optimization problem by transfering the inner optimization routine to a penalty term. The main generalization result (Theorem 9 in \cite{arjovsky2019invariant}) states that if both $\Phi$ and $w$ come from the class of linear models (i.e., $\mathcal{H}_{\Phi}=\mathbb{R}^{n \times n}$ and $\mathcal{H}_{w}=\mathbb{R}^{n \times 1}$), under certain conditions on the diversity of training environments (Assumption 8 in \cite{arjovsky2019invariant}) and the data generation, the invariant predictor $w \circ \Phi$ obtained by solving Eq.~(\ref{eq:irm}) remains invariant in $\mathcal{E}_{all}$.


\section{Problem Setup}


\subsection{A Motivating Example} \label{sect:motiv_example}

In this section, we extend the example which was introduced by \citet{wright1921correlation} and discussed by \citet{arjovsky2019invariant}, and provide a further in-depth analysis.

\begin{model} \label{model:toy}
Consider a structural equation model (SEM) with a discrete environment variable $E$ that modulates the noises in the structural assignments connecting the other variables (cf.\ Fig.~\ref{fig:toy_gt} below):
\begin{align*}
    X_1 &\leftarrow \text{Gaussian}(0, \sigma_1(E)), \\
    Y &\leftarrow X_1 + \text{Gaussian}(0, \sigma_2(E)), \\
    X_2 &\leftarrow Y + \text{Gaussian}(0, \sigma_3(E)),
\end{align*}
where $\text{Gaussian}(0, \sigma)$ denotes a Gaussian random variable with zero mean and standard deviation $\sigma$, and $\sigma_1,\ldots,\sigma_3$ are functions of the value $e \in \mathcal{E}_{\text{all}}$ taken by the environment variable $E$.
\end{model}

To ease exposition, here we consider the simple scenario in which $\mathcal{E}_{\text{all}}$ only contains all modifications varying the noises of $X_1$, $X_2$ and $Y$ within a finite range, i.e., $\sigma_i(e) \in [0, \sigma^2_{\text{max}}]$. Then, to predict $Y$ from $(X_1, X_2)$ using a least-square predictor $\hat{Y}^e = \hat{\alpha}_1X_1^e+\hat{\alpha}_2X_2^e$ for environment $e$, we can
\begin{itemize}
\item Case 1: regress from $X_1^e$, to obtain $\hat{\alpha}_1=1$ and $\hat{\alpha}_2=0$,
\item Case 2: regress from $X_2^e$, to obtain $\hat{\alpha}_1=0$ and $\hat{\alpha}_2=\frac{\sigma_1(e)+\sigma_2(e)}{\sigma_1(e)+\sigma_2(e)+\sigma_3(e)}$,
\item Case 3: regress from $(X_1^e, X_2^e)$, to obtain $\hat{\alpha}_1=\frac{\sigma_3(e)}{\sigma_2(e)+\sigma_3(e)}$ and $\hat{\alpha}_2=\frac{\sigma_2(e)}{\sigma_2(e)+\sigma_3(e)}$.
\end{itemize}

In the generic scenario (i.e., $\sigma_1(e)\neq0$, $\sigma_2(e)\neq0$, and $\sigma_3(e)\neq0$), the regression using $X_1$ in Case 1 is an invariant correlation: it is the only regression whose coefficients do not vary with $e$. By contrast, the regressions in both Case 2 and Case 3 have coefficients that depend on $e$. Therefore, only the invariant correlation in Case 1 will generalize well to new test environments.

From a practical perspective, let us take a closer look at Case 3. As we do not know in advance that regressing on $X_1$ alone will lead to an invariant predictor, in practice we may do the regression on all the available data $(X_1^e, X_2^e)$. As explained, when $\sigma_i(e)\neq 0$ for $i=1,2,3$, this fails to generalize. Indeed, no empirical risk minimization (ERM) algorithm (i.e., purely minimizing training error) \citep{Vapnik95} would work in this setting. Invariant Causal Prediction (ICP) methods \citep{peters2015causal} also do not work, since as argued by \citet{arjovsky2019invariant}, the noise variance in $Y$ may change across environments. To this end, \citet{arjovsky2019invariant} proposed IRM. As aforementioned, however, IRM and IRMG guarantee generalization to unseen environments only in the linear setting, while we consider the case where both $\Phi$ and $w$ are from the class of nonlinear models.

\begin{figure}[t]
\centering
\begin{subfigure}{.45\textwidth}
\centering
\begin{tikzpicture}[
	> = latex, 
    auto,
    observed/.style={circle, draw=black, fill=black!15, thick, inner sep=0pt, minimum size=6mm},
    unobserved/.style={circle, draw=black, thick, inner sep=0pt, minimum size=6mm},
    surrogate/.style={rectangle, draw=black, fill=black!15, thick, minimum size=5mm},
]        

        \node[observed] (X1) {\tiny $X_1$};
        \node[observed] (Y) [right=0.6cm of X1] {\tiny $Y$};
        \node[observed] (X2) [right=0.6cm of Y] {\tiny $X_2$};
        \node[surrogate] (E) [above=0.6cm of Y] {\tiny $E$};
       
        \path[->, very thick] (X1) edge (Y);
        \path[->, very thick] (Y) edge (X2);
        \path[->, very thick] (E) edge (X1);
        \path[->, very thick] (E) edge (Y);
		\path[->, very thick] (E) edge (X2);
        
\end{tikzpicture}
\vspace{1.2cm}
\caption{} 
\label{fig:toy_gt}
\end{subfigure}%
\begin{subfigure}{.45\textwidth}
\centering
\begin{tikzpicture}[
	> = latex, 
    auto,
    observed/.style={circle, draw=black, fill=black!15, thick, inner sep=0pt, minimum size=6mm},
    unobserved/.style={circle, draw=black, thick, inner sep=0pt, minimum size=6mm},
    surrogate/.style={rectangle, draw=black, fill=black!15, thick, minimum size=5mm},
]        
		
		\node[unobserved] (X1) {\tiny $X_1$};
        \node[observed] (Y) [right=0.6cm of X1] {\tiny $Y$};
        \node[unobserved] (X2) [right=0.6cm of Y] {\tiny $X_2$};
        \node[surrogate] (E) [above=0.6cm of Y] {\tiny $E$};
        \node[observed] (X) [below=0.6cm of Y] {\tiny $\boldsymbol{O}$};
       
        \path[->, very thick] (X1) edge (Y);
        \path[->, very thick] (Y) edge (X2);
        \path[->, very thick] (E) edge (X1);
        \path[->, very thick] (E) edge (Y);
		\path[->, very thick] (E) edge (X2);
		\path[->, very thick] (X1) edge (X);
		\path[->, very thick] (X2) edge (X);
        
\end{tikzpicture}
\caption{} 
\label{fig:toy_uc}
\end{subfigure} \\
\begin{subfigure}{.8\textwidth}
\centering
\begin{tikzpicture}[
	> = latex, 
    auto,
    observed/.style={circle, draw=black, fill=black!15, thick, inner sep=0pt, minimum size=6mm},
    unobserved/.style={circle, draw=black, thick, inner sep=0pt, minimum size=6mm},
    surrogate/.style={rectangle, draw=black, fill=black!15, thick, minimum size=5mm},
]        
		
		\node[unobserved] (Xp1) {\tiny $X_{p_1}$};
		\node (xpdots) [right=0.1cm of Xp1] {$\cdots$};
		\node[unobserved] (Xpr) [right=0.1cm of xpdots] {\tiny $X_{p_r}$};
        \node[observed] (Y) [right=0.6cm of Xpr] {\tiny $\boldsymbol{Y}$};
        \node[unobserved] (Xc1) [right=0.6cm of Y] {\tiny $X_{c_1}$};
        \node (xcdots) [right=0.1cm of Xc1] {$\cdots$};
		\node[unobserved] (Xck) [right=0.1cm of xcdots] {\tiny $X_{c_k}$};
        \node[surrogate] (E) [above=0.6cm of Y] {\tiny $\boldsymbol{E}$};
        \node[observed] (X) [below=0.6cm of Y] {\tiny $\boldsymbol{O}$};
       
        \path[->, very thick] (Xp1) edge [bend right=45] (Y);
        \path[->, very thick] (Xpr) edge (Y);
        \path[->, very thick, dashed, red] (Xp1) edge [-,bend left=30] (Xpr);
        \path[->, very thick, dashed, red] (Xp1) edge [-,bend left=30] (Xc1);
        \path[->, very thick, dashed, red] (Xp1) edge [-,bend left=20] (Xck);
        \path[->, very thick, dashed, red] (Xpr) edge [-,bend left=30] (Xc1);
        \path[->, very thick, dashed, red] (Xpr) edge [-,bend left=30] (Xck);
        \path[->, very thick, dashed, red] (Xc1) edge [-,bend left=30] (Xck);
        \path[->, very thick, dashed] (Y) edge (Xc1);
        \path[->, very thick, dashed] (Y) edge [bend right=45] (Xck);
        \path[->, very thick, dashed] (E) edge (Xp1);
        \path[->, very thick, dashed] (E) edge (Xpr);
		\path[->, very thick, dashed] (E) edge (Xc1);
		\path[->, very thick, dashed] (E) edge (Xck);
		\path[->, very thick] (Xp1) edge (X);
		\path[->, very thick] (Xpr) edge (X);
		\path[->, very thick] (Xc1) edge (X);
		\path[->, very thick] (Xck) edge (X);
        
\end{tikzpicture} 
\caption{} 
\label{fig:general}
\end{subfigure}
\caption{(a) Causal structure of Model \ref{model:toy}. (b) A more practical extension of Model \ref{model:toy}, where $X_1$ and $X_2$ are not directly observed and $\boldsymbol{O}$ is their observation. (c) A general version of (b), where we assume there exist multiple unobserved variables. Each of them could be either a parent, a child of $\boldsymbol{Y}$, or has no direct connection with $\boldsymbol{Y}$. We allow for arbitrary connections between the latent variables (red dashed lines) as long as the resulting causal diagram including $\boldsymbol{Y}$ is a directed acyclic graph (DAG). Grey nodes denote observed variables and white nodes represent unobserved variables. Dashed lines denote the edges which might vary across environments and even be absent in some scenarios, whilst solid lines indicate that they are invariant across all the environments.} 
\end{figure}
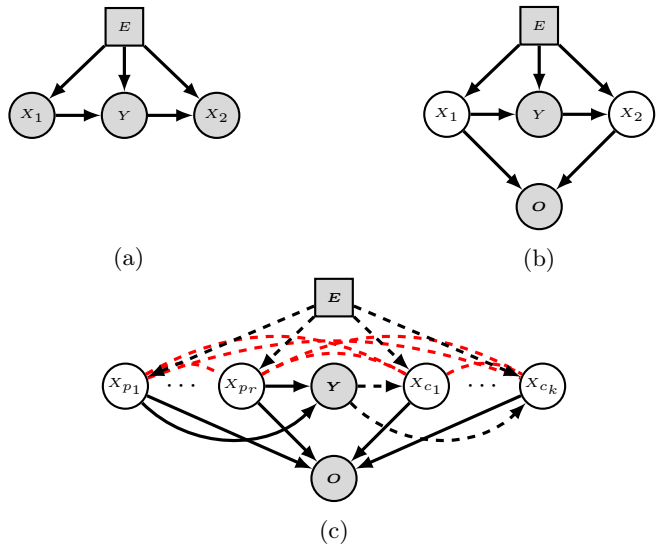


Another way to understand Model 1 is through its graphical representation\footnote{The relation between SEM and its graphical representation is formally defined in Appendix \ref{appendix:dfn}.}, as shown in Fig.~\ref{fig:toy_gt}. We treat the environment as a random variable $E$, where $E$ could be any information specific to the environment \citep{Storkey,peters2015causal,zhang2017causal,huang2020causal}. For simplicity, we let $E$ be the environment index, i.e., $E \in \{1, \ldots, N\}$, where $N$ is the number of training environments. A more realistic version appearing in many settings is shown in Fig.~\ref{fig:toy_uc}, where the true variables $\{X_1, X_2\}$ are unobserved and we  can only observe their transformation $\boldsymbol{O}$. In this case, Invariant Causal Prediction (ICP) \citep{peters2015causal} will fail when applied to $\boldsymbol{O}$, even when $Y$ is not affected by $E$ (i.e., the edge $E \rightarrow Y$ is removed). The reason is that each variable (i.e., each dimension) of $\boldsymbol{O}$ is jointly influenced by both $X_1$ and $X_2$ so that ICP is unable to find the variables containing the information only about $X_1$ by searching for a subset of variables $\boldsymbol{O}$. By contrast, both IRM and IRMG work, as long as the transformation is linear. These findings are also empirically illustrated in Section \ref{sect:syn_data}. We now go even further and consider a more general causal graph in which $\boldsymbol{Y}$ can have more than one parent or child.


\subsection{Assumptions on the Causal Graph}
\label{sect:asm_graph}

We extend the causal graph in Fig.~\ref{fig:toy_uc} to a more general setting\footnote{For simplicity, we do not explicitly consider unobserved confounders in this paper. In particular, we assume that there are no unobserved confounders between $\boldsymbol{X}, \boldsymbol{Y}, \boldsymbol{O}$, and $\boldsymbol{E}$.}, as encapsulated in Fig.~\ref{fig:general}. In particular, we now have $\boldsymbol{O} \in \mathcal{O} \subseteq \mathbb{R}^d$, $\boldsymbol{Y} \in \mathcal{Y} \subseteq \mathbb{R}^s$, $\boldsymbol{X}=(X_{p_1}, \ldots, X_{p_r}, X_{c_1}, \ldots, X_{c_k}) \in \mathcal{X} \subseteq \mathbb{R}^n$, where $n=r+k$, and $\{X_i\}_{i \in I_{p}=\{p_1, \ldots, p_r\}}$ and $\{X_j\}_{j \in I_{c}=\{c_1, \ldots, c_k\}}$ are multiple scalar \textit{causal factors} and \textit{non-causal factors}\footnote{This means that $X_{j \in I_c}$ could be either an effect of $\boldsymbol{Y}$, independent of $\boldsymbol{Y}$, or spuriously correlated with $\boldsymbol{Y}$ via a third set of confounders (i.e., both $X_j$ and $\boldsymbol{Y}$ are affected by a subset of $\{X_i\}_{i \neq j}$ and $\boldsymbol{E}$).} of $\boldsymbol{Y}$, respectively. We denote $\boldsymbol{X}_p \doteq (X_{p_1}, \ldots, X_{p_r})$ and $\boldsymbol{X}_c \doteq (X_{c_1}, \ldots, X_{c_k})$ for the ease of clarification. We also assume that $\boldsymbol{X}$ is of lower dimension than $\boldsymbol{O}$, that is, $n \leq d$. We allow for arbitrary connections between the latent variables $\boldsymbol{X}$ as long as the resulting causal diagram including $\boldsymbol{Y}$ is a directed acyclic graph (DAG). We use dashed lines to indicate the \textit{causal mechanisms} which might vary across environments and even be absent in some scenarios, whilst solid lines indicate that they are invariant across all the environments. To sum up, we assume that the underlying causal graph encapsulated in Fig.~\ref{fig:general} satisfies the following assumption\footnote{For generality, we replace $Y$ and $E$ with $\boldsymbol{Y}$ and $\boldsymbol{E}$ to allow for multi-dimensional variables.}:

\begin{asm}\label{eq:assumption1} 
(a) $X_i$ depends on one or both of $\boldsymbol{Y}$ and $\boldsymbol{E}$ for any $i$; (b) The causal graph containing $\boldsymbol{X}$ and $\boldsymbol{Y}$ is a DAG; (c) $\boldsymbol{O} \indep \boldsymbol{Y}, \boldsymbol{E} | \boldsymbol{X}$, implying that $p(\boldsymbol{O}|\boldsymbol{X})$ is invariant across all the environments; (d) $\boldsymbol{Y} \indep \boldsymbol{E} | \boldsymbol{X}_p$, implying that $p(\boldsymbol{Y}|\boldsymbol{X}_p)$ is invariant across all the environments.
\end{asm}

One may be wondering how practical Assumption \ref{eq:assumption1} is in real world applications. Let us explore this in more detail.
Assumption \ref{eq:assumption1}a rules out all the useless $X_i$ in the task of predicting $\boldsymbol{Y}$. This is because if Assumption \ref{eq:assumption1}a is violated, meaning that $X_i$ is independent of $\boldsymbol{Y}$ and $\boldsymbol{E}$ and has no influence in predicting $\boldsymbol{Y}$, then such $X_i$ should be viewed as noise and thus eliminated during learning. Assumption \ref{eq:assumption1}b is a common assumption in causal discovery \citep{spirtes2000causation,pearl2009causality,peters2015causal}. It also makes sense in Assumption \ref{eq:assumption1}c that the generative mechanism $p(\boldsymbol{O}|\boldsymbol{X})$ is invariant across all the environments. Otherwise, it is impossible to infer $\boldsymbol{X}$ from $\boldsymbol{O}$ in any unseen environment. Assumption \ref{eq:assumption1}d is a widely-used default assumption in OOD generalization \citep{peters2015causal,arjovsky2019invariant}. In fact, Assumption \ref{eq:assumption1}d can be further relaxed to the more practical one that $\mathbb{E}[\boldsymbol{Y}|\boldsymbol{X}_p]$ is invariant across all the environments. That is, given $\boldsymbol{X}_p$, we allow $\boldsymbol{E}$ to only affect the amount of noise in the distribution of $\boldsymbol{Y}$, because that would not change the expected value of $\boldsymbol{Y}$ since the mean of the noise would be zero. Apparently, Assumption \ref{eq:assumption1}, together with the causal graph in Fig.~\ref{fig:general}, covers most scenarios (e.g., the ones of \citet{zhang2013domain,Julius,sun2020latent,ahuja2021invariance,von2021self}) and is a very flexible model for causal analysis when predicting $\boldsymbol{Y}$ from $\boldsymbol{O}$.

\subsection{Assumptions on the Prior}
\label{sect:asm_on_prior}

When the underlying causal graph satisfies Assumption \ref{eq:assumption1}, our primary assumption leading to identifiability in this general setting is that the conditional prior $p(\boldsymbol{X} | \boldsymbol{Y}, \boldsymbol{E})$ belongs to a general exponential family. This is formalized as follows:

\begin{asm} \label{eq:assumption2}
$p(\boldsymbol{X} | \boldsymbol{Y}, \boldsymbol{E})$
 belongs to a general exponential family  with parameter vector given by an
arbitrary function $\boldsymbol{\lambda}(\boldsymbol{Y}, \boldsymbol{E})$ and sufficient statistics $\boldsymbol{T}(\boldsymbol{X})=[ \boldsymbol{T}_{f}(\boldsymbol{X})^\text{T}, \boldsymbol{T}_{NN}(\boldsymbol{X})^\text{T}]^\text{T}$ given by the concatenation of a) the sufficient statistics $\boldsymbol{T}_{f}(\boldsymbol{X})=[\boldsymbol{T}_{1}(X_1)^T,\cdots,\boldsymbol{T}_{n}(X_n)^T]^T$ of a factorized exponential family, where all the $\boldsymbol{T}_{i}(X_i) $ have dimension larger or equal to 2, and b) the output $\boldsymbol{T}_{NN}(\boldsymbol{X})$ of a neural network with ReLU activations. The resulting density function is thus given by
\begin{align} 
p_{\boldsymbol{T}, \boldsymbol{\lambda}}(\boldsymbol{X}|\boldsymbol{Y}, \boldsymbol{E}) = {\mathcal{Q}(\boldsymbol{X})}/{\mathcal{Z}(\boldsymbol{Y}, \boldsymbol{E})} \exp\big[ \boldsymbol{T}(\boldsymbol{X})^\text{T} \boldsymbol{\lambda}(\boldsymbol{Y}, \boldsymbol{E}) ],\label{eq:asm4_prior}
\end{align}
where $\mathcal{Q}$ is the base measure and $\mathcal{Z}$ the normalizing constant.
\end{asm}

A neural network with ReLU activation has universal approximation power. Therefore, the term $\boldsymbol{T}_{NN}(\boldsymbol{X})$ in the above prior distribution will allow us to capture arbitrary dependencies between the latent variables.
 
The distribution in Eq. (\ref{eq:asm4_prior}) is more flexible than the conditionally factorized prior  assumed by iVAEs. However,  surprisingly, the identifiability results of iVAEs also hold when using the more flexible prior in Eq. (\ref{eq:asm4_prior}), as we  will show in Section \ref{sect:phase1}. This motivates using an extended iVAE model with the above prior to model data generated by the ground truth  model in Fig.~\ref{fig:general}. However, in this case, the data generating model and the learned model might  have different priors. For example, in the ground truth  model, the prior for each $X_{i \in I_p}$ 
might be $p(X_i|\boldsymbol{E})$, when $X_i$ is only caused by $\boldsymbol{E}$. By contrast, in the extended iVAE model the prior is  $p(\boldsymbol{X}|\boldsymbol{Y}, \boldsymbol{E})$. Is this going to affect the identifiability of the latent variables? Well, in practice not because the posterior distribution for $\boldsymbol{X}$ given $\boldsymbol{O}$, $\boldsymbol{Y}$ and $\boldsymbol{E}$ would be equivalent in both models (up to identifiability guarantees).


\section{Invariant Causal Representation Learning}

We now introduce our algorithm, invariant Causal Representation Learning (iCaRL), which consists of 3 phases as summarized in Algorithm \ref{alg:icrl}. The basic idea is that we first identify  $\boldsymbol{X}$ by using an extended iVAE model under Assumptions \ref{eq:assumption1}\&\ref{eq:assumption2} (Phase \ref{sect:phase1}), then discover direct causes of $\boldsymbol{Y}$ among the identified $\boldsymbol{X}$  (Phase \ref{sect:phase2}), and finally learn an invariant predictor for $\boldsymbol{Y}$ from the discovered  causes (Phase \ref{sect:phase3}). 

\begin{algorithm}[bt]
\SetAlgoLined
\textbf{Phase 1:} We first learn a NF-iVAE model, including the decoder  and its corresponding encoder, by optimizing the objective function in Eq.~(\ref{eq:phase1_vaesm}) on the data $\{\boldsymbol{O}, \boldsymbol{Y}, \boldsymbol{E}\}$. Then, we use the mean of the NF-iVAE encoder to infer the latent variables $\boldsymbol{X}$ from observations  $\{\boldsymbol{O}, \boldsymbol{Y}, \boldsymbol{E}\}$. The latent variables are guaranteed to be identified up to a permutation and simple transformation. \\
\textbf{Phase 2:} After inferring $\boldsymbol{X}$, we can discover direct causes (parents) of $\boldsymbol{Y}$ by testing all pairs of latent variables with independence testing and conditional independence testing, i.e., finding a set of latent variables in which each pair of $X_i$ and $X_j$ satisfies that the dependency between them increases after conditioning on $\boldsymbol{Y}$.\footnote{The highly unlikely single-cause case is left in Appendix \ref{appendix:phase2_single}.} \\
\textbf{Phase 3:} Having obtained $\text{\normalfont Pa}(\boldsymbol{Y})$, we can solve the optimization problems in Eq.~(\ref{eq:phase3_w}) to learn the invariant classifier $w$. When in a new environment, we first infer $\text{\normalfont Pa}(\boldsymbol{Y})$ from $\boldsymbol{O}$ by solving Eq.~(\ref{eq:phase3_phi}) and then leverage the learned $w$ for prediction.
\caption{Invariant Causal Representation Learning (iCaRL)} \label{alg:icrl}
\end{algorithm}


\subsection{Phase 1: Identifying Latent Variables}
\label{sect:phase1}

In this section, we describe our identifiable model, namely NF-iVAE, which is an extended iVAE with a general non-factorized prior that is able to capture complex dependences between the latent variables. Technically, in the general setting under Assumption \ref{eq:assumption1}, it is straightforward to obtain a corresponding generative model by directly substituting $\boldsymbol{U}$ with $(\boldsymbol{Y}, \boldsymbol{E})$ in Eq.~(\ref{eq:ivae_gen}):
\begin{align} 
p_{\boldsymbol{\theta}}(\boldsymbol{O}, \boldsymbol{X}|\boldsymbol{Y}, \boldsymbol{E}) &= p_{\boldsymbol{f}}(\boldsymbol{O}|\boldsymbol{X})p_{\boldsymbol{T}, \boldsymbol{\lambda}}(\boldsymbol{X}|\boldsymbol{Y}, \boldsymbol{E}), \label{eq:phase1_gen}\\
p_{\boldsymbol{f}}(\boldsymbol{O}|\boldsymbol{X}) &= p_{\boldsymbol{\epsilon}}(\boldsymbol{O}-\boldsymbol{f}(\boldsymbol{X})). \label{eq:phase1_likelihood}
\end{align} 
The corresponding lower bound is
\begin{multline}
\mathcal{L}^{\text{VAE}}_{\text{phase1}}(\boldsymbol{\theta}, \boldsymbol{\phi}) 
:=  \mathbb{E}_{p_D}\big[\mathbb{E}_{q_{\boldsymbol{\phi}}(\boldsymbol{X}|\boldsymbol{O}, \boldsymbol{Y}, \boldsymbol{E})}\left[\log p_{\boldsymbol{f}}(\boldsymbol{O}|\boldsymbol{X})\right. \\ \left. + \log p_{\boldsymbol{T}, \boldsymbol{\lambda}}(\boldsymbol{X}|\boldsymbol{Y}, \boldsymbol{E})-\log q_{\boldsymbol{\phi}}(\boldsymbol{X}|\boldsymbol{O}, \boldsymbol{Y}, \boldsymbol{E})\right]\big]. \label{eq:phase1_loss}
\end{multline}
To obtain an identifiability result, we assume that the prior $p_{\boldsymbol{T}, \boldsymbol{\lambda}}(\boldsymbol{X}|\boldsymbol{Y}, \boldsymbol{E})$ satisfies Assumption \ref{eq:assumption2} (i.e., Eq. (\ref{eq:asm4_prior})). Since the prior is a general multivariate exponential family distribution with unknown normalization constant, we cannot learn its parameters $(\boldsymbol{T}, \boldsymbol{\lambda})$ by directly maximizing Eq. (\ref{eq:phase1_loss}). Instead, we use  score matching, a well-known method for training unnormalized probabilistic models \citep{hyvarinen2005estimation,vincent2011connection}, and learn $(\boldsymbol{T}, \boldsymbol{\lambda})$ by minimizing
\begin{multline}  
\mathcal{L}^{\text{SM}}_{\text{phase1}}(\boldsymbol{T}, \boldsymbol{\lambda}) 
:=  \mathbb{E}_{p_D}\big[\mathbb{E}_{q_{\boldsymbol{\phi}}(\boldsymbol{X}|\boldsymbol{O}, \boldsymbol{Y}, \boldsymbol{E})}\left[||\nabla_{\boldsymbol{X}} \log q_{\boldsymbol{\phi}}(\boldsymbol{X}|\boldsymbol{O}, \boldsymbol{Y}, \boldsymbol{E}) \right. \\ \left. - \nabla_{\boldsymbol{X}} \log p_{\boldsymbol{T}, \boldsymbol{\lambda}}(\boldsymbol{X}|\boldsymbol{Y}, \boldsymbol{E})||^2
\right]\big]. \label{eq:phase1_sm}
\end{multline}
In practice, we can use a simple trick of partial integration to simplify the evaluation of Eq. (\ref{eq:phase1_sm}), see Appendix \ref{appendix:derivation}. Furthermore, we can jointly learn $(\boldsymbol{\theta}, \boldsymbol{\phi})$ by combining Eq. (\ref{eq:phase1_loss}) and Eq. (\ref{eq:phase1_sm}) in the following objective:
\begin{align}
\mathcal{L}_{\text{phase1}}(\boldsymbol{\theta}, \boldsymbol{\phi}) = \mathcal{L}^{\text{VAE}}_{\text{phase1}}(\boldsymbol{f}, \hat{\boldsymbol{T}}, \hat{\boldsymbol{\lambda}}, \boldsymbol{\phi}) - \mathcal{L}^{\text{SM}}_{\text{phase1}}(\hat{\boldsymbol{f}}, \boldsymbol{T}, \boldsymbol{\lambda}, \hat{\boldsymbol{\phi}}), \label{eq:phase1_vaesm}
\end{align}
where $\hat{\boldsymbol{f}}, \hat{\boldsymbol{T}}, \hat{\boldsymbol{\lambda}}, \hat{\boldsymbol{\phi}}$ are
copies of $\boldsymbol{f}, \boldsymbol{T}, \boldsymbol{\lambda}, \boldsymbol{\phi}$
that are treated as constants and whose
gradient is not calculated during learning. More details can be found in Appendix \ref{appendix:implementation}.

We now state our main theoretical results:
\begin{thm} \label{thm:1}
Assume that we observe data sampled from a generative model defined according to Eqs. (\ref{eq:asm4_prior}-\ref{eq:phase1_likelihood}), with parameters $\boldsymbol{\theta}:=(\boldsymbol{f}, \boldsymbol{T}, \boldsymbol{\lambda})$, where $p_{\boldsymbol{T}, \boldsymbol{\lambda}}(\boldsymbol{X}|\boldsymbol{Y}, \boldsymbol{E})$
satisfies Assumption \ref{eq:assumption2}. Furthermore, assume the following holds: (i) The set $\{\boldsymbol{O} \in \mathcal{O}|\varphi_{\boldsymbol{\epsilon}}(\boldsymbol{O})=0\}$ has measure zero, where $\varphi_{\boldsymbol{\epsilon}}$ is the characteristic function of the density $p_{\boldsymbol{\epsilon}}$ defined in Eq. (\ref{eq:phase1_likelihood}). (ii) Function $\boldsymbol{f}$ in Eq. (\ref{eq:phase1_likelihood}) is injective, and has all second-order cross derivatives. (iii) The sufficient statistics in
$\boldsymbol{T}_f$ are all twice differentiable.
(iv) There exist $k+1$ distinct points $(\boldsymbol{Y}, \boldsymbol{E})^0, \ldots, (\boldsymbol{Y}, \boldsymbol{E})^{k}$ such that the matrix $L=\left(\boldsymbol{\lambda}((\boldsymbol{Y}, \boldsymbol{E})^1)-\boldsymbol{\lambda}((\boldsymbol{Y}, \boldsymbol{E})^0),\ldots,\boldsymbol{\lambda}((\boldsymbol{Y}, \boldsymbol{E})^{k})-\boldsymbol{\lambda}((\boldsymbol{Y}, \boldsymbol{E})^0)\right)$ of size $k \times k$ is invertible, where $k$ is the dimension of $\boldsymbol{T}$. Then the parameters $\boldsymbol{\theta}$ are identifiable up to a permutation and a {\bf  ``simple transformation"}
of the latent variables $\boldsymbol{X}$, defined as a componentwise nonlinearity making  each recovered $\boldsymbol{T}_{i}(X_i) $ in $\boldsymbol{T}_{f}(\boldsymbol{X})$ equal to the original up to a linear
operation.
\end{thm}
Note that, this theorem is inspired by but beyond the main results of iVAEs in that the former is predicated on Assumption \ref{eq:assumption2} which is more flexible than the conditionally factorized prior assumed in iVAEs. It results in several key changes in the proof, clarified in Appendix \ref{appendix:proof}. Interestingly, from (iv) we can further see that $\boldsymbol{E}$ is unnecessary when there exist $k+1$ distinct points $\boldsymbol{Y}^0, \ldots, \boldsymbol{Y}^{k}$ such that the matrix $L=\left(\boldsymbol{\lambda}(\boldsymbol{Y}^1)-\boldsymbol{\lambda}(\boldsymbol{Y}^0),\ldots,\boldsymbol{\lambda}(\boldsymbol{Y}^{k})-\boldsymbol{\lambda}(\boldsymbol{Y}^0)\right)$ of size $k \times k$ is invertible. Not requiring $\boldsymbol{E}$ would make our approach even more applicable.

We further have the following consistency result for the estimation.

\begin{thm} \label{thm:2}
Assume that the following holds: (i) The family of distributions $q_{\boldsymbol{\phi}}(\boldsymbol{X}|\boldsymbol{O}, \allowbreak \boldsymbol{Y}, \boldsymbol{E})$ contains $p_{\boldsymbol{\theta}}(\boldsymbol{X}|\boldsymbol{O}, \boldsymbol{Y}, \boldsymbol{E})$, and $q_{\boldsymbol{\phi}}(\boldsymbol{X}|\boldsymbol{O}, \allowbreak \boldsymbol{Y}, \boldsymbol{E})>0$ everywhere. (ii) We maximize $\mathcal{L}_{\text{phase1}}(\boldsymbol{\theta}, \boldsymbol{\phi})$ with respect to both $\boldsymbol{\theta}$ and $\boldsymbol{\phi}$. Then in the limit of infinite data, we learn the true parameters $\boldsymbol{\theta}^{\ast}$ up to a permutation and simple transformation of the latent variables $\boldsymbol{X}$.
\end{thm}
As a consequence of Theorem \ref{thm:1}\&\ref{thm:2}, we have:
\begin{thm} \label{thm:3}
Assume the hypotheses of Theorem \ref{thm:1} and Theorem \ref{thm:2} hold, then in the limit of infinite data, we identify the true latent variables $\boldsymbol{X}^{\ast}$ up to a permutation and simple transformation.    
\end{thm}
Theorem \ref{thm:3} states that we can use NF-iVAE to infer the true  $\boldsymbol{X}$ up to a permutation and simple transformation. We use the mean of $q_{\boldsymbol{\phi}}(\boldsymbol{X}|\boldsymbol{O}, \allowbreak \boldsymbol{Y}, \boldsymbol{E})$ for this task. Note that the noise $\boldsymbol{\epsilon}$ may introduce
uncertainty in the estimation of  $\boldsymbol{X}$. However, when $\boldsymbol{O}$ is high dimensional and $\boldsymbol{X}$ is low dimensional (as common in real world applications),
$q_{\boldsymbol{\phi}}(\boldsymbol{X}|\boldsymbol{O}, \allowbreak \boldsymbol{Y}, \boldsymbol{E})$ will be highly concentrated and we will still be able to estimate $\boldsymbol{X}$ with high accuracy. The good results obtained by our method in various experiments seem to corroborate this. All three theorems are proven in Appendix \ref{appendix:proof}.


\subsection{Phase 2: Discovering Direct Causes}
\label{sect:phase2}

After estimating  $\boldsymbol{X}$ for each data point, the next step is to determine which components of $\boldsymbol{X}$ are a direct cause of $\boldsymbol{Y}$. 
We denote these components by $\text{\normalfont Pa}(\boldsymbol{Y})$. From Assumption \ref{eq:assumption1}, one observation is that for any two latent variables $X_i$ and $X_j$, only when both are causes of $\boldsymbol{Y}$ do we have that the dependency between them increases after conditioning on $\boldsymbol{Y}$. Thus, when there exist at least two causal latent variables, it is trivial to test all pairs of latent variables with independence testing \citep{gretton2007kernel} and conditional independence testing \citep{zhang2012kernel} to discover all of them\footnote{These conditional independence tests can be performed in parallel to largely accelerate the testing procedure. Note that, in practice, it might occur that there exist some $X_i$ which is independent of any other $X_j$ when conditioning on $\boldsymbol{Y}$ and $\boldsymbol{E}$. It is probably because such $X_i$ is a deterministic transformation of $\boldsymbol{Y}$. In this special case, we can use IGCI \citep{daniusis2012inferring,janzing2012information} to determine whether or not $X_i$ is a cause of $\boldsymbol{Y}$. Also note that, in some scenarios in which the dependence signals between a pair of causal latent variables might be weak due to the data issue, we can test the conditional independence of such a latent variable with all the other causal latent variables. If it is conditionally dependent on more than half of them or its average p-value is larger than the pre-specified threshold, we will select it as one cause of $\boldsymbol{Y}$.} by comparing $p$-values from these two tests. Conversely, if no such a pair is found, it implies that there is at most one causal latent variable. This is a highly unlikely case in real world applications, which we leave to Appendix \ref{appendix:phase2_single}.


\subsection{Phase 3: Learning an Invariant Predictor} \label{sect:phase3}

After having obtained the causal latent variables $\text{\normalfont Pa}(\boldsymbol{Y})$ for $\boldsymbol{Y}$ across training environments, we can learn $w$ by solving the following optimization problem:
\begin{align}\small
\underset{w \in \mathcal{H}_{w}}{\min} 
\sum\nolimits_{e \in \mathcal{E}_{tr}} R^e(w) 
=\underset{w \in \mathcal{H}_{w}}{\min} \sum\nolimits_{e \in \mathcal{E}_{tr}}\mathbb{E}_{\text{\normalfont Pa}(\boldsymbol{Y}^e), \boldsymbol{Y}^e}\left[\ell_1(w(\text{\normalfont Pa}(\boldsymbol{Y}^e)), \boldsymbol{Y}^e)\right],
\label{eq:phase3_w} 
\end{align}
where $\ell_1(\cdot)$ could be any loss. Since we assume that $\mathbb{E}(\boldsymbol{Y}|\text{\normalfont Pa}(\boldsymbol{Y}))$ is invariant across $\mathcal{E}_{all}$ (the relaxed version of Assumption \ref{eq:assumption1}d), the learned $w$ is guaranteed to perform well across $\mathcal{E}_{all}$. 

The remaining question is how to infer $\text{\normalfont Pa}(\boldsymbol{Y})$ (i.e., $\boldsymbol{X}_p$) from $\boldsymbol{O}$ in a new environment. This can be implemented by leveraging the learned $p(\boldsymbol{O}|\boldsymbol{X})$. The rationale behind is that $p(\boldsymbol{O}|\boldsymbol{X})$ is assumed to be invariant across $\mathcal{E}_{all}$ (Assumption \ref{eq:assumption1}c). In light of this idea, we follow \citet{sun2020latent} and infer $\boldsymbol{X}_p$ from $\boldsymbol{O}$ in any new testing environment by solving the following optimization problem: 
\begin{align}\small
    \underset{\boldsymbol{X}_p, \boldsymbol{X}_{c}}{\max} \log p_{\boldsymbol{f}}(\boldsymbol{O}|\boldsymbol{X}_p, \boldsymbol{X}_{c}) + \lambda_1 ||\boldsymbol{X}_p||_2^2 + \lambda_2 ||\boldsymbol{X}_{c}||_2^2, \label{eq:phase3_phi} 
\end{align}
where the hyperparameters $\lambda_1 > 0$ and $\lambda_2 >0$ control the learned $\boldsymbol{X}_p$ and $\boldsymbol{X}_{c}$ have a reasonable scale. For optimization, we follow \citet{schott2018towards} to first use values of $\boldsymbol{X}$ sampled from the training set as initial points and then use Adam to optimize for several iterations. Note that 
the noise $\boldsymbol{\epsilon}$ will introduce
uncertainty in the estimation of 
$\boldsymbol{X}_p$ and $\boldsymbol{X}_{c}$ from $\boldsymbol{O}$.
However, as we mentioned before (below Theorem \ref{thm:3}),
this noise is not going to affect the estimation much because the likelihood will be highly concentrated around the ground truth values, as corroborated by 
our good empirical results.

A key question is if iCaRL performs well across $\mathcal{E}_{all}$ even though it uses only data from $\mathcal{E}_{tr}$. That is, does iCaRL enable OOD generalization, as defined by  \citet{arjovsky2019invariant}?
The answer is positive since Theorem A.1 in 
\citep{DBLP:journals/corr/abs-2103-02667} indicates that i) any predictor 
$w \circ \Phi$ with optimal OOD generalization uses only
$\text{\normalfont Pa}(\boldsymbol{Y})$
to compute $\Phi$ and ii) the 
classifier $w$ in this optimal predictor
 can be estimated using data from any environment $e$ for which the distribution of $\text{\normalfont Pa}(\boldsymbol{Y}^e)$ has full support, which will always be the case since the conditional prior in Eq. (\ref{eq:asm4_prior}) has full support. Finally, 
 Theorem A.1 in 
\citep{DBLP:journals/corr/abs-2103-02667} also indicates that iii)
 the optimal predictor will
 be invariant across $\mathcal{E}_{all}$. Key to these results is that $\text{\normalfont Pa}(\boldsymbol{Y}^e)$ are available
 when solving (\ref{eq:phase3_phi}). This requires first to identify the latent variables 
 $\boldsymbol{X}$ from $\boldsymbol{O}$, $\boldsymbol{Y}$ and $\boldsymbol{E}$ and second, to discover the direct causes of $\boldsymbol{Y}$. The hypotheses of Theorems \ref{thm:1} and \ref{thm:2} and Assumption \ref{eq:assumption1} provide this guarantee. We therefore have the following result whose proof is in Appendix \ref{appendix:proof}.

\begin{ppn}\label{prop:2}
Under Assumption \ref{eq:assumption1} and the assumptions of Theorems \ref{thm:1} and \ref{thm:2}, the predictor learned by iCaRL across $\mathcal{E}_{tr}$ in the limit of infinite data has optimal OOD generalization across $\mathcal{E}_{all}$.
\end{ppn}
 

\section{Experiments}

We compare our approach with a variety of methods on both synthetic and real-world datasets. In all comparisons, unless stated otherwise, we average performance over ten runs. The supplement contains all the details of the experiments, e.g., datasets (Appendix \ref{appendix:datasets}), implementation details (Appendix \ref{appendix:implementation}), hyperparameters and architectures (Appendix \ref{appendix:architectures}), etc.


\begin{table}
\caption{Regression on synthetic data: Comparison of methods in terms of MSE (mean $\pm$ std deviation).} 
\label{tab:toy_small}
\begin{adjustbox}{width=0.75\columnwidth,center}
\begin{tabular}{clcc}
\multicolumn{1}{c}{\thead{$g(\cdot)$}} &\multicolumn{1}{c}{\bf METHOD}   &\multicolumn{1}{c}{\thead{\bf TRAIN \\ ($\sigma_3=\{0.2, 2\}$)}} &\multicolumn{1}{c}{\thead{\bf TEST \\ ($\sigma_3=100$)}}
\\ \hline \vspace{-2mm}\\
\multirow{4}{*}{Identity} & ERM & $0.00 \pm 0.00$ & $\boldsymbol{0.00 \pm 0.00}$  \\
                          & IRM & $0.00 \pm 0.00$ & $\boldsymbol{0.00 \pm 0.00}$ \\
                          & F-IRM GAME & $0.98 \pm 0.23$ & $1.03 \pm 0.04$ \\
                          & V-IRM GAME & $0.99 \pm 2.74$ & $1.07 \pm 2.26$ \\
                          & {\bf iCaRL (ours)} & $0.01 \pm 0.03$ & $1.00 \pm 0.01$ \\
\hline \vspace{-2mm}\\
\multirow{4}{*}{Linear}   & ERM & $0.00 \pm 0.00$ & $\boldsymbol{0.00 \pm 0.00}$ \\
                          & IRM & $0.00 \pm 0.00$ & $\boldsymbol{0.00 \pm 0.00}$ \\
                          & F-IRM GAME & $0.99 \pm 0.01$ & $1.08 \pm 0.06$ \\
                          & V-IRM GAME & $1.00 \pm 5.98$ & $1.05 \pm 0.04$ \\
                          & {\bf iCaRL (ours)} & $0.01 \pm 0.03$ & $1.01 \pm 0.04$ \\
\hline \vspace{-2mm}\\
\multirow{4}{*}{Nonlinear}& ERM & $0.06 \pm 0.01$ & $220.79 \pm 229.97$ \\
                          & IRM & $0.08 \pm 0.01$ & $149.60 \pm 104.85$ \\
                          & F-IRM GAME & $1.06 \pm 0.09$ & $196.59 \pm 150.71$ \\
                          & V-IRM GAME & $1.00 \pm 0.01$ & $170.46 \pm 125.62$ \\
                          & {\bf iCaRL (ours)} & $ 0.29 \pm 0.04$ & $\boldsymbol{28.16 \pm 2.54}$ \\ 
\end{tabular}
\end{adjustbox} 
\end{table}

\subsection{Synthetic data I} \label{sect:syn_data}

We first conduct a series of experiments on synthetic data generated according to an extension of the SEM in Model \ref{model:toy}. The extension is to map the variables $\boldsymbol{X}:=(X_1, X_2)$ into a 10 dimensional observation $\boldsymbol{O}$ through a linear or nonlinear transformation. Our goal is to predict $Y$ from $\boldsymbol{O}$, where $\boldsymbol{O}=g(\boldsymbol{X})$. We consider three transformations:
\begin{itemize}
\item[(a)] \textit{Identity}: $g(\cdot)$ is the identity matrix $\boldsymbol{I} \in \mathbb{R}^{2 \times 2}$, i.e., $\boldsymbol{O}=g(\boldsymbol{X})=\boldsymbol{X}$. 
\item[(b)] \textit{Linear}: $g(\cdot)$ is a random matrix $\boldsymbol{S} \in \mathbb{R}^{2 \times 10}$, i.e., $\boldsymbol{O}=g(\boldsymbol{X})=\boldsymbol{X}\cdot\boldsymbol{S}$.
\item[(c)] \textit{Nonlinear}: $g(\cdot)$ is given by a neural network with 2-dimensional input and 10-dimensional output, whose parameters are randomly set in advance.
\end{itemize}
Since this is a regression task, we use the mean squared error (MSE) as a metric of performance. Note that, in this problem, there is only one causal latent variable $X_1$, meaning that the conditional prior in Eq. (\ref{eq:asm4_prior}) will not exhibit dependencies. Because of this, in this case we do not include a $\boldsymbol{T}_{NN}(\boldsymbol{X})$ term in our NF-iVAE prior. In the following section we do consider more complicated settings with many potential causal latent variables and, in that case, we do include  $\boldsymbol{T}_{NN}(\boldsymbol{X})$ in the NF-iVAE prior.

We consider a simple scenario in which we fix $\sigma_1=1$ and $\sigma_2=0$ for all environments and only allow $\sigma_3$ to vary across environments. In this case, $\sigma_3$ controls the spurious correlations between $X_2$ and $Y$. Each experiment draws 1000 samples from each of the three environments $\sigma_3=\{0.2, 2, 100\}$, where the first two are for training and the third for testing. We compare with the following baselines:\footnote{We also tried ICP, but ICP was unable to find any parent of $Y$ even in the identity case.} ERM, and two variants of IRMG: F-IRM Game (with $\Phi$ fixed to the identity) and V-IRM Game (with variable $\Phi$).

\begin{figure}[t]
    \centering
    \includegraphics[width=\textwidth]{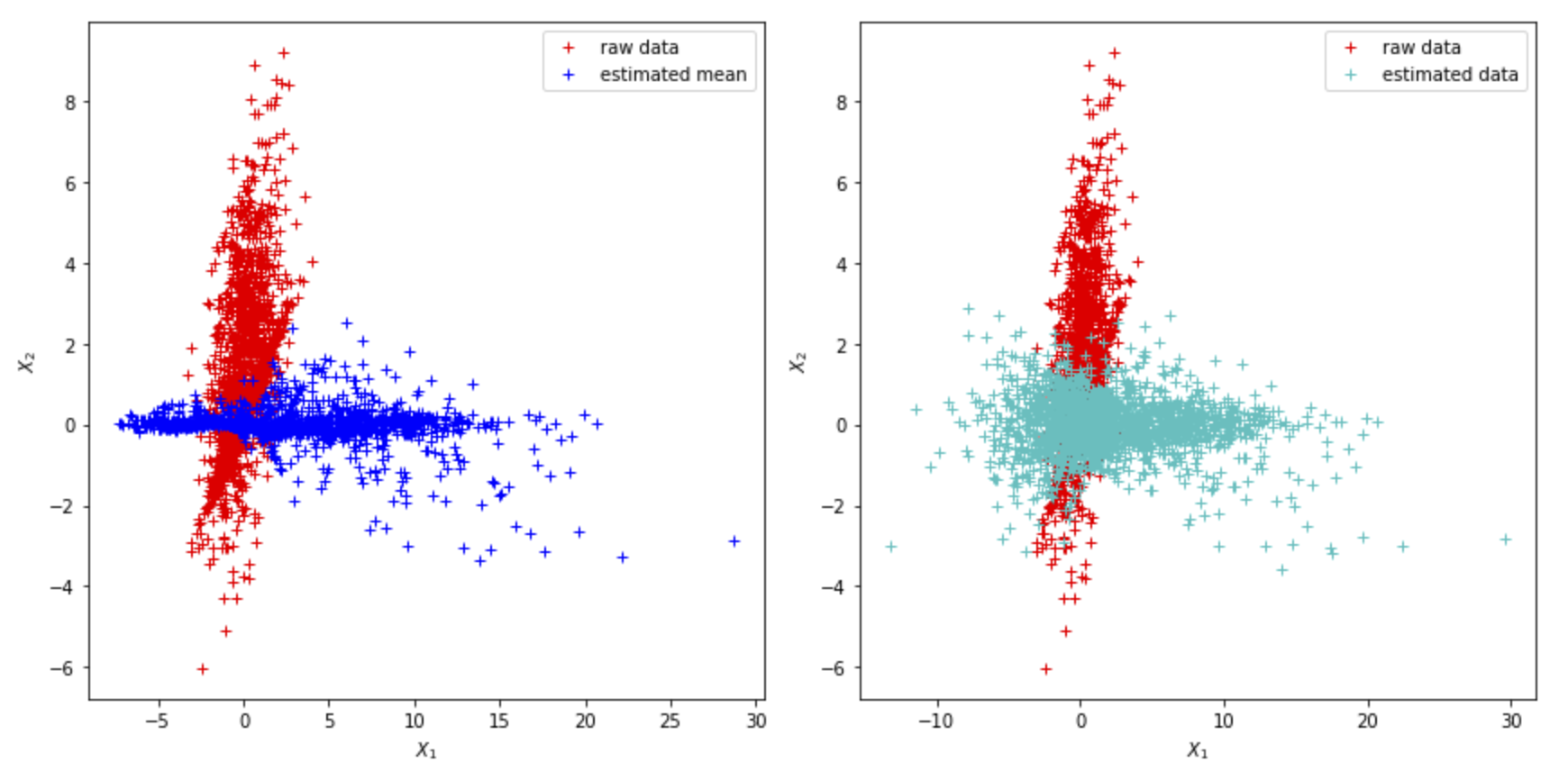}
    \caption{Left: Comparison of the raw data $\boldsymbol{X}$ and the mean of $\hat{\boldsymbol{X}}$ inferred through the learned inference model in the \textit{Nonlinear} case. Right: Comparison of the raw data $\boldsymbol{X}$ and the sampled points $\hat{\boldsymbol{X}}$ using the reparameterization trick in the \textit{Nonlinear} case. The comparisons clearly show that the inferred $\hat{\boldsymbol{X}}$ is equal to $\boldsymbol{X}$ up to a permutation and pointwise transformation.} 
    \label{fig:verify_phase1}
\end{figure}

As shown in Table \ref{tab:toy_small}, in the cases of \textit{Identity} and \textit{Linear}, our approach is better than IRMG but only comparable with ERM and IRM. This might be because the identifiability result up to a simple nonlinear transformation renders the problem more difficult by converting the original identity or linear problem into a nonlinear problem. In the \textit{Nonlinear} case, the gains of iCaRL are very clear. 

We perform additional experiments to analyze the key to our approach, i.e., the importance of NF-iVAE in Phase 1. Theorem \ref{thm:3} tells us that we can leverage NF-iVAE to learn the true latent variables up to a permutation and componentwise transformation. We empirically verify this point by comparing the raw data $\boldsymbol{X}$ with the corresponding $\hat{\boldsymbol{X}}$ inferred through the learned inference model in NF-iVAE. Fig. \ref{fig:verify_phase1} clearly shows that the inferred $\hat{\boldsymbol{X}}$ is equal to $\boldsymbol{X}$ up to a permutation and componentwise transformation.

We also conduct an experiment in which we replace NF-iVAE with the original VAE in Phase 1. As shown in Table \ref{tab:vae}, the performance of iCaRL based on NF-iVAE significantly outperforms the one based on VAE. It is worth noting that when VAE is instead used in Phase 1, it usually occurs in Phase 2 that either all the dimensions or no dimension of $\hat{\boldsymbol{X}}$ are identified as the parents of $\boldsymbol{Y}$. This is because all components of $\hat{\boldsymbol{X}}$ are mixed together and will influence one another even when conditioning on $\boldsymbol{Y}$ and $E$.


\begin{table}[thb]
\caption{Results on synthetic data: Comparison of NF-iVAE and VAE used in Phase 1 in terms of MSE (mean $\pm$ std deviation).}
\label{tab:vae}
\begin{adjustbox}{width=0.75\columnwidth,center}
\begin{tabular}{clcc}
\multicolumn{1}{c}{\thead{$g(\cdot)$}} &\multicolumn{1}{c}{\bf METHOD}   &\multicolumn{1}{c}{\thead{\bf TRAIN \\ ($\sigma_3=\{0.2, 2\}$)}} &\multicolumn{1}{c}{\thead{\bf TEST \\ ($\sigma_3=100$)}}
\\ \hline \\
\multirow{2}{*}{Nonlinear}& iCaRL-VAE & $0.26 \pm 0.09$ & $1174.96 \pm 1385.81$ \\
                          & {\bf iCaRL} & $\boldsymbol{0.29 \pm 0.04}$ & $\boldsymbol{28.16 \pm 2.54}$ \\
\end{tabular}
\end{adjustbox}
\end{table}


\begin{figure}[t]
\centering
\begin{subfigure}{.4\textwidth}
\centering
\begin{tikzpicture}[
	> = latex, 
    auto,
    observed/.style={circle, draw=black, fill=black!15, thick, inner sep=0pt, minimum size=6mm},
    unobserved/.style={circle, draw=black, thick, inner sep=0pt, minimum size=6mm},
    surrogate/.style={rectangle, draw=black, fill=black!15, thick, minimum size=5mm},
]        
		
		\node[unobserved] (X1) {\tiny $X_1$};
		\node[unobserved] (X2) [right=0.6cm of X1] {\tiny $X_2$};
        \node[observed] (Y) [right=0.6cm of X2] {\tiny $Y$};
        \node[surrogate] (E) [above=0.6cm of Y] {\tiny $E$};
        \node[observed] (X) [below=0.6cm of Y] {\tiny $\boldsymbol{O}$};
       
        \path[->, very thick] (X1) edge [bend right=45] (Y);
        \path[->, very thick] (X2) edge (Y);
        \path[->, very thick] (E) edge (X1);
        \path[->, very thick] (E) edge (X2);
		\path[->, very thick] (X1) edge (X);
		\path[->, very thick] (X2) edge (X);
        
\end{tikzpicture}
\caption{} 
\label{fig:two_causes_g}
\end{subfigure}%
\begin{subfigure}{.4\textwidth}
\centering
\begin{align}
    E &\sim \mathcal{U}\left\{0.2, 2, 3, 5\right\} \nonumber\\
    X_1 &\sim \mathcal{N}(X_1 | E, 1) \nonumber\\
    X_2 &\sim \mathcal{N}(X_2 | 2E, 2) \nonumber\\
    Y &\sim \mathcal{N}(Y | X_1 + X_2, 1) \nonumber\\
    \boldsymbol{O} &= g(X_1, X_2) \nonumber
\end{align}
\caption{} 
\label{fig:two_causes_dg}
\end{subfigure}
\caption{(a) Causal structure with $Y$ having two causes. (b) Data generating process corresponding to (a), where $\mathcal{U}\{\cdot\}$ denotes the discrete uniform distribution, $\mathcal{N}(\cdot)$ the Gaussian distribution, and $g(\cdot)$ is given by a neural network with 2-dimensional input and 10-dimensional output, whose parameters are randomly set in advance.} \label{fig:two_causes}
\end{figure}
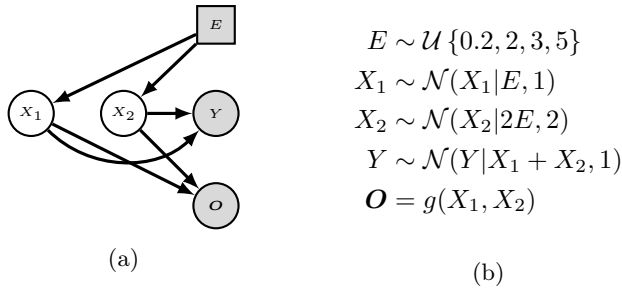

\begin{table}[t]
\centering
\caption{Comparisons of assumptions on the prior leading to identifiability in different algorithms.} 
\label{tab:ivae_comparison}\vspace{3mm}
\begin{tabular}{l|c}
{\bf METHOD} & {\bf Assumption on the Prior for Identifiability}   
\\ \hline \\
    VAE   & {\small Non-identifiability with $p_{\boldsymbol{T}, \boldsymbol{\lambda}}(\boldsymbol{X})=\prod\nolimits_i p(X_i) \stackrel{e.g.}{=} \boldsymbol{\mathcal{N}}(\boldsymbol{0}, \boldsymbol{I})$}   \\
    iVAE        & {\small $p_{\boldsymbol{T}, \boldsymbol{\lambda}}(\boldsymbol{X}|\boldsymbol{Y},\boldsymbol{E}) = \prod\nolimits_i {\mathcal{Q}_i(X_i)}/{\mathcal{Z}_i(\boldsymbol{Y},\boldsymbol{E})} \exp[\sum\nolimits_{j=1}^k T_{i,j}(X_i)\lambda_{i,j}(\boldsymbol{Y},\boldsymbol{E})]$}  \\
    ICE-BeeM  & {\small $p_{\boldsymbol{T}, \boldsymbol{\lambda}}(\boldsymbol{X}|\boldsymbol{Y},\boldsymbol{E}) =
    {\mathcal{Q}(\boldsymbol{X})}/{\mathcal{Z}(\boldsymbol{Y}, \boldsymbol{E})}\prod\nolimits_i  \exp[\sum\nolimits_{j=1}^k T_{i,j}(X_i)\lambda_{i,j}(\boldsymbol{Y},\boldsymbol{E})]$}  \\
    NF-iVAE  & {\small $p_{\boldsymbol{T}, \boldsymbol{\lambda}}(\boldsymbol{X}|\boldsymbol{Y}, \boldsymbol{E}) = {\mathcal{Q}(\boldsymbol{X})}/{\mathcal{Z}(\boldsymbol{Y}, \boldsymbol{E})} \exp\big[ \boldsymbol{T}(\boldsymbol{X})^\text{T} \boldsymbol{\lambda}(\boldsymbol{Y}, \boldsymbol{E}) ]$}  \\
\hline                    
\end{tabular} 

\end{table}
\begin{figure}[t]
\centering
\begin{subfigure}{.3\textwidth}
\centering
\includegraphics[width=\textwidth]{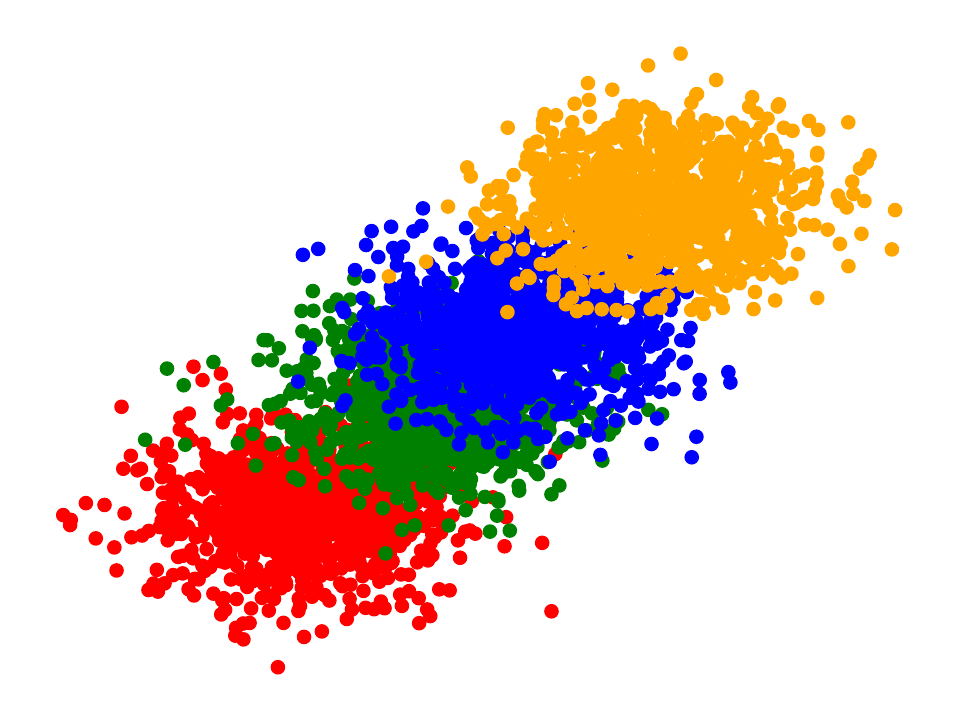}
\caption{Original}
\end{subfigure}%
\begin{subfigure}{.3\textwidth}
\centering
\includegraphics[width=\textwidth]{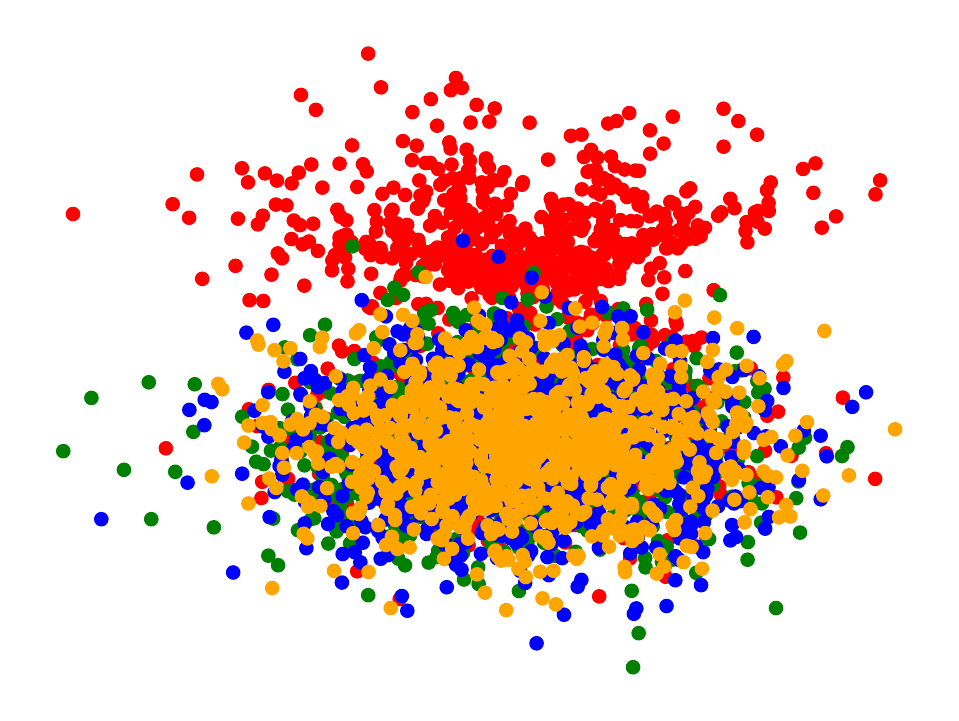}
\caption{VAE}
\end{subfigure}%
\hspace{2mm}
\begin{subfigure}{.3\textwidth}
\centering
\includegraphics[width=\textwidth]{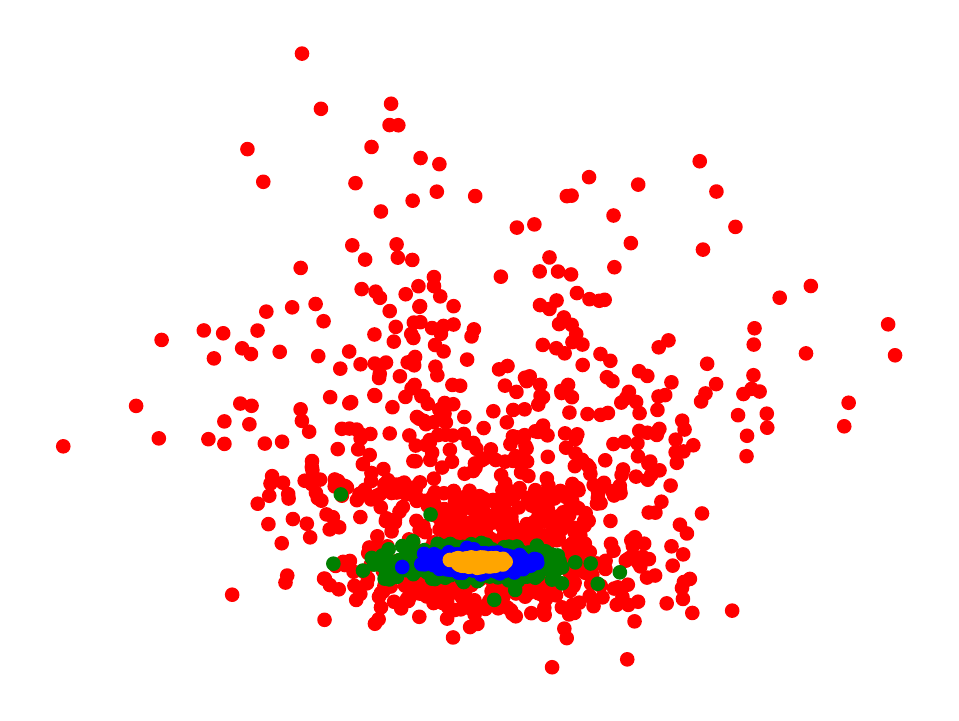}
\caption{iVAE}
\end{subfigure} \\
\begin{subfigure}{.3\textwidth}
\centering
\includegraphics[width=\textwidth]{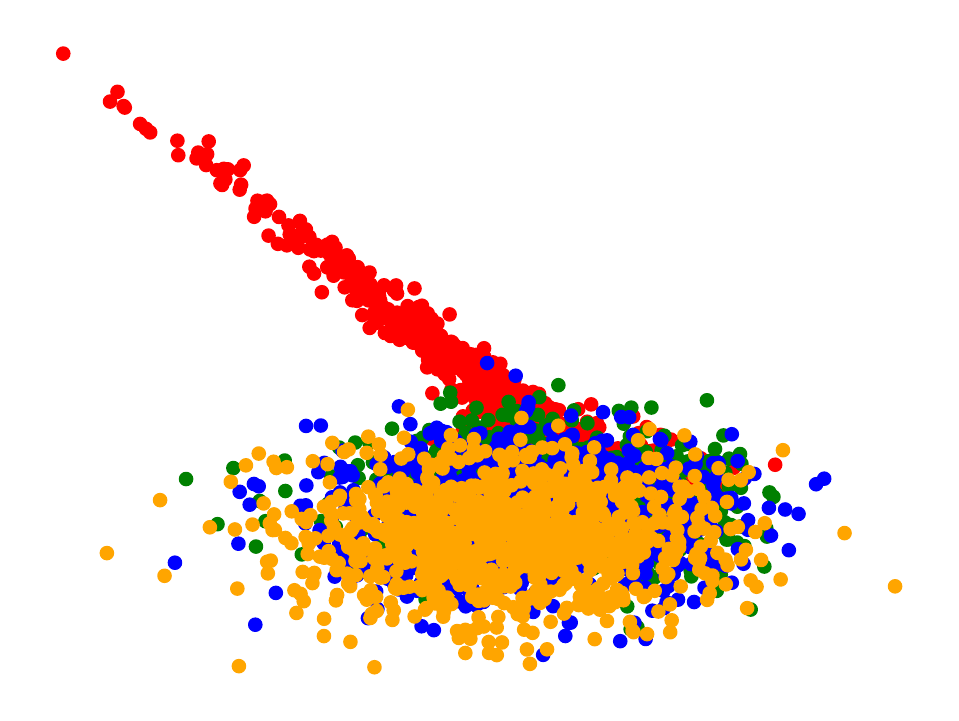}
\caption{ICE-BeeM}
\end{subfigure}%
\hspace{2mm}
\begin{subfigure}{.3\textwidth}
\centering
\includegraphics[width=\textwidth]{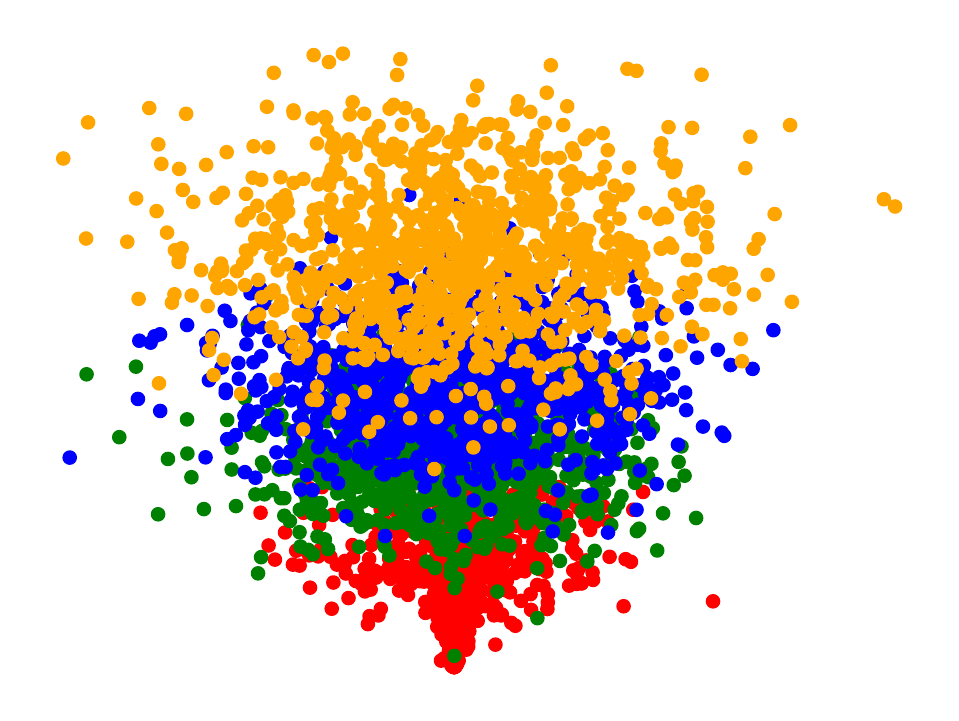}
\caption{NF-iVAE}
\end{subfigure}%
\begin{subfigure}{.3\textwidth}
\centering
\includegraphics[width=\textwidth]{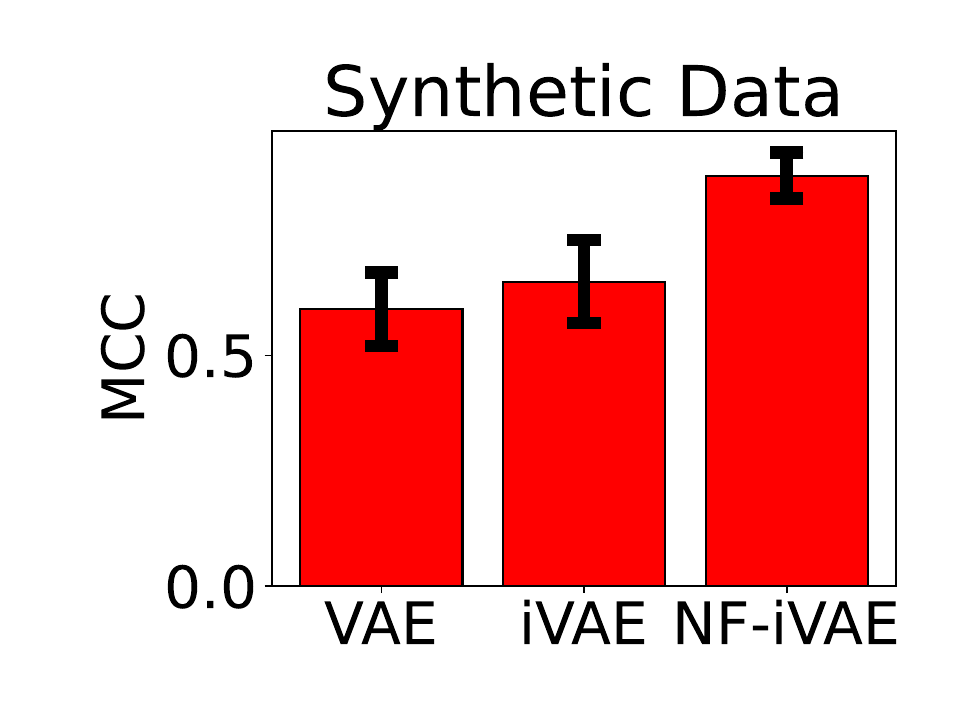}
\caption{MCC on toy data}
\label{fig:4f}
\end{subfigure}
\caption{Visualization of the samples (i.e., $\hat{\boldsymbol{X}}=(\hat{X}_1, \hat{X}_2)$) in latent space recovered through different algorithms: (a) Samples from the true distribution; (b-d) Samples from the posterior inferred using VAE, iVAE and ICE-BeeM, respectively. Apparently, our method (e) can recover the original data up to a permutation and a simple componentwise transformation. (f) Mean correlation coefficient (MCC) scores for VAE, iVAE, and NF-iVAE on synthetic data.} 
\label{fig:idf_visualization}
\end{figure}

\subsection{Synthetic data II}

To further verify the identifiability of NF-iVAE, we conduct a series of experiments on more complicated synthetic data generated according to the data generating process (Fig. \ref{fig:two_causes_dg}) corresponding to the causal graph shown in Fig. \ref{fig:two_causes_g}. The reason we choose this setting is that it is the simplest case satisfying our requirements: a) For ease of visualization, the latent space had better be 2-dimensional; b) To introduce the non-factorized prior given $\boldsymbol{Y}$ and $\boldsymbol{E}$ (i.e., $X_i \notindep X_j| \boldsymbol{Y}, \boldsymbol{E}$), $\boldsymbol{Y}$ has at least two causes.

We draw 1000 samples from each of the four environments $E=\{0.2, 2, 3, 5\}$, and thus the whole synthetic dataset consists of 4000 samples. The task is to recover the true latent variable $\boldsymbol{X}=(X_1, X_2)$ using the samples of $\boldsymbol{O}$, $\boldsymbol{E}$, and $\boldsymbol{Y}$. We compare with the following baselines: VAE \citep{kingma2013auto} (without identifiability guarantees), iVAE \citep{khemakhem2020variational} (with a conditionally factorized prior for identifiability), and ICE-BeeM \citep{khemakhem2020ice}. It is worth noting that in ICE-BeeM the primary assumption leading to identifiability is similar to that in iVAE, where the base measure $\mathcal{Q}(\boldsymbol{X})$ could be arbitrary to capture the dependences between latent variables but the exponential term still has to factorize across components (dimensions). All these are summarized in Table \ref{tab:ivae_comparison}. Clearly, from the table we can see that our method has a more general assumption on the prior leading to identifiability. This is also demonstrated empirically in Fig. \ref{fig:idf_visualization}. Our method NF-iVAE can recover the original data $\boldsymbol{X}$ up to a permutation and a simple componentwise transformation, whereas all the other methods fail because they are unable to handle the non-factorized case in which $X_i \notindep X_j| \boldsymbol{Y}, \boldsymbol{E}$. 

We also compute the mean correlation coefficient (MCC) used in \citet{khemakhem2020variational}, which can be obtained by calculating the correlation coefficient between all pairs of true and recovered latent factors and then solving a linear sum assignment problem by assigning each recovered latent factor to the true latent factor with which it best correlates. By definition, higher MCC scores indicate stronger identifiability. From Fig. \ref{fig:4f}, we can see that the MCC score for NF-iVAE is significantly greater than those of VAE and iVAE, indicating much stronger identifiability.


\subsection{Colored MNIST, Colored Fashion MNIST, and VLCS}

In this section, we first report experiments on two datasets used in IRM and IRMG: Colored MNIST (CMNIST) and Colored Fashion MNIST (CFMNIST). We follow the setting of \cite{ahuja2020invariant} to create these two datasets. The task is to predict a binary label assigned to each image which is originally grayscale but artificially colored in a way that the color is correlated strongly but spuriously with the class label. We add noise to the preliminary binary label\footnote{In Colored MNIST, the preliminary label is defined as $Y = 0$ if the digit is between 0-4 and $Y = 1$ if the digit is between 5-9. In Colored Fashion MNIST, the preliminary label is defined as $Y = 0$ if the image is from categories: ``t-shirt'', ``trouser'', ``pullover'', ``dress'', ``coat'', ``shirt'' and $Y = 1$ if the image is from categories: ``sandal'', ``sneaker'', ``ankle boots''. Please see more details in Appendix \ref{appendix:datasets}.} (i.e., $Y = \{0,1\}$) by flipping it with 25 percent probability to construct the final labels. We sample the color id by flipping the final labels with probability $p_e$, where $p_e$ is $0.2$ in the first environment, $0.1$ in the second environment, and $0.9$ in the third environment. The third environment is the testing environment. For all the experiments on these two datasets, we set the number of the latent variables to $n=10$. 

\begin{figure}[t] 
\centering
\begin{subfigure}{.4\textwidth}
\centering
\includegraphics[width=\textwidth]{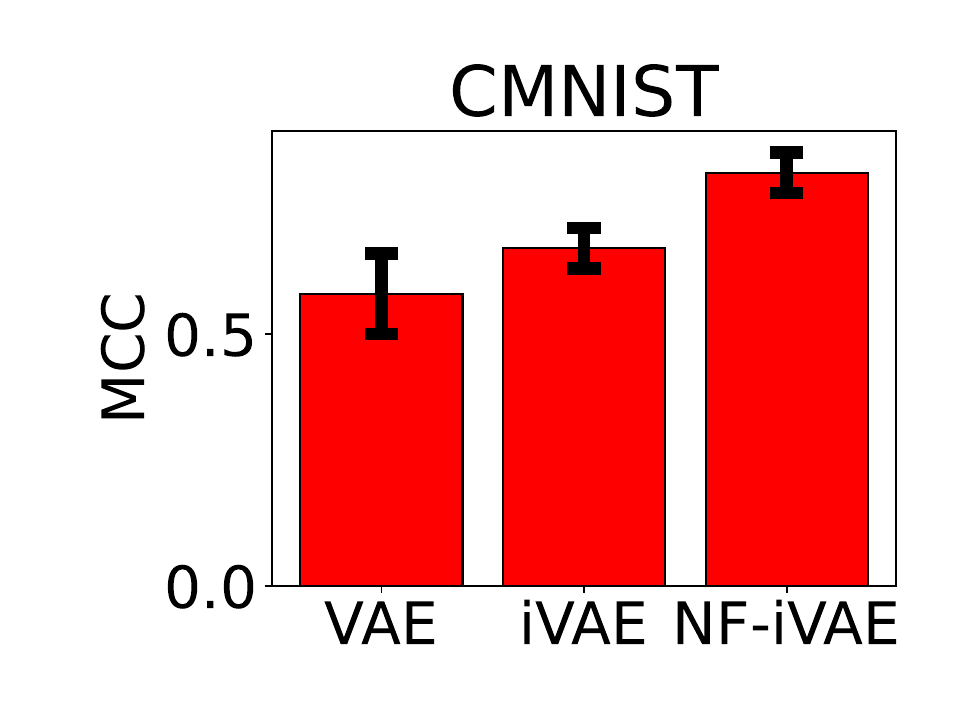}
\caption{MCC on CMNIST} 
\label{fig:3g}
\end{subfigure} \hspace{5mm}%
\begin{subfigure}{.4\textwidth}
\centering
\includegraphics[width=0.92\textwidth]{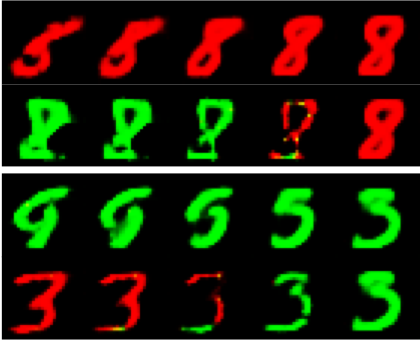}
\caption{Intervening CMNIST} 
\label{fig:3h}
\end{subfigure} 
\caption{(a) MCC scores for VAE, iVAE, ICE-BeeM, and NF-iVAE on CMNIST. (b) The effects on the CMNIST images of digit $8$ (top two rows) and digit $3$ (bottom two rows) when intervening on a causal factor $X_{i \in I_p}$ and on a non-causal factor (effect) $X_{j \in I_c}$, respectively.} \label{fig:main_two_causes}
\end{figure}

\begin{table}[t] 
\centering
\caption{Colored Fashion MNIST: Comparison of methods in terms of accuracy (\%) (mean $\pm$ std deviation).} 
\label{tab:fashion} 
\LARGE
\begin{adjustbox}{width=0.8\columnwidth,center}
\begin{tabular}{lcc}
\multicolumn{1}{c}{\bf ALGORITHM}  &\multicolumn{1}{c}{\bf TRAIN ACCURACY} &\multicolumn{1}{c}{\bf TEST ACCURACY}
\\ \hline \\
ERM                 & $83.17 \pm 1.01$ & $22.46 \pm 0.68$ \\
ERM 1               & $81.33 \pm 1.35$ & $33.34 \pm 8.85$\\
ERM 2               & $84.39 \pm 1.89$ & $13.16 \pm 0.82$ \\
ROBUST MIN MAX      & $82.81 \pm 0.11$ & $29.22 \pm 8.56$ \\
F-IRM GAME          & $62.31 \pm 2.35$ & $69.25 \pm 5.82$ \\
V-IRM GAME          & $68.96 \pm 0.95$ & $70.19 \pm 1.47$ \\
IRM                 & $75.01 \pm 0.25$ & $55.25 \pm 12.42$ \\
{\bf iCaRL (ours)}          & $\boldsymbol{74.96 \pm 0.37}$ & $\boldsymbol{73.61 \pm 0.63}$ \\
ERM GRAYSCALE       & $74.79 \pm 0.37$ & $74.67 \pm 0.48$ \\
OPTIMAL             & $75$ & $75$ \\ \hline 
\end{tabular}
\end{adjustbox}
\end{table}

\begin{table}[t] 
\centering 
\caption{Colored MNIST: Comparison of methods in terms of accuracy (\%) (mean $\pm$ std deviation).}\vspace{-2mm}
\label{tab:mnist}
\LARGE
\begin{adjustbox}{width=0.8\columnwidth,center}
\begin{tabular}{lcc}
\multicolumn{1}{c}{\bf ALGORITHM}  &\multicolumn{1}{c}{\bf TRAIN ACCURACY} &\multicolumn{1}{c}{\bf TEST ACCURACY}
\\ \hline \\
ERM                 & $84.88 \pm 0.16$ & $10.45 \pm 0.66$ \\
ERM 1               & $84.84 \pm 0.21$ & $10.86 \pm 0.52$ \\
ERM 2               & $84.95 \pm 0.20$ & $10.05 \pm 0.23$ \\
ROBUST MIN MAX      & $84.25 \pm 0.43$ & $15.24 \pm 2.45$ \\
F-IRM GAME          & $63.37 \pm 1.14$ & $59.91 \pm 2.69$ \\
V-IRM GAME          & $63.97 \pm 1.03$ & $49.06 \pm 3.43$ \\
IRM                 & $59.27 \pm 4.39$ & $62.75 \pm 9.59$ \\
{\bf iCaRL (ours)}          & $\boldsymbol{70.56 \pm 0.81}$ & $\boldsymbol{68.75 \pm 1.45}$ \\
ERM GRAYSCALE       & $71.81 \pm 0.47$ & $71.36 \pm 0.65$ \\
OPTIMAL             & $75$ & $75$ \\ \hline 
\end{tabular}
\end{adjustbox}
\end{table}

\begin{table}[t] 
\centering 
\caption{VLCS: Comparison of methods in terms of accuracy (\%) (mean $\pm$ std deviation).}\vspace{-2mm}
\label{tab:vlcs}
\LARGE
\begin{adjustbox}{width=0.65\columnwidth,center}
\begin{tabular}{lc}
\multicolumn{1}{c}{\bf ALGORITHM}  &\multicolumn{1}{c}{\bf TEST ACCURACY} 
\\ \hline \\
ERM               & $77.4 \pm 0.3$ \\
IRM               & $78.1 \pm 0.0$ \\
DRO \citep{sagawa2019distributionally}& $77.2 \pm 0.6$ \\
Mixup \citep{yan2020improve} & $77.7 \pm 0.4$ \\
CORAL \citep{sun2016deep}   & $77.7 \pm 0.5$ \\
MMD \citep{li2018domain}    & $76.7 \pm 0.9$ \\
DANN \citep{ganin2016domain} & $78.7 \pm 0.3$ \\
C-DANN \citep{li2018deep}   & $78.2 \pm 0.4$ \\
LaCIM \citep{sun2020latent} & $78.4 \pm 0.5$ \\
{\bf iCaRL (ours)}& $\boldsymbol{81.8 \pm 0.6}$ \\ \hline 
\end{tabular}
\end{adjustbox}
\end{table}

Likewise, we investigate the identifiability of NF-iVAE on CMNIST by computing the MCC score between samples of the true latent variable and of the recovered latent variable. Since the true latent variable on CMNIST is inaccessible to us, we follow \citet{khemakhem2020ice} and compute an average MCC score between samples of latent variables recovered by different models trained with different random initialization. As shown in Fig. \ref{fig:3g}, it is evident that the MCC score for NF-iVAE greatly outperforms the others, showing that the latent variables recovered by NF-iVAE have much better identifiability.

Furthermore, we demonstrate the ability of iCaRL to discover the causal latent variables (Phase 2) by visualizing the generated images through performing intervention upon a causal latent variable and a non-causal latent variable, respectively. Fig. \ref{fig:3h} shows how intervening upon each of them affects the image. Obviously, intervening on a causal latent variable affects the shape of the digit but not its color (top plots), whilst intervening on a non-causal latent variable, which is an effect in Fig. \ref{fig:3h}, affects the color of the digit only (bottom plots). This visually verifies the results of iCaRL in Phase 2. 

In terms of the OOD generalization performance, we compare iCaRL with 1) IRM, 2) two variants of IRMG: F-IRM Game (with $\Phi$ fixed to the identity) and V-IRM Game (with a variable $\Phi$), 3) three variants of ERM: ERM (on entire training data), ERM $e$ (on each environment $e$), and ERM GRAYSCALE (on data with no spurious correlations), and 4) ROBUST MIN MAX (minimizing the maximum loss across the multiple environments). Table \ref{tab:fashion} shows that iCaRL outperforms all other baselines on CFMNIST. It is worth emphasising that the train and test accuracies of iCaRL closely approach the ones of ERM GRAYSCALE and OPTIMAL, implying that iCaRL approximately learns the true invariant causal representations with almost no correlation with the spurious color feature. We can draw similar conclusions from the results on CMNIST in Table \ref{tab:mnist}. However, this dataset seems more difficult because even ERM GRAYSCALE, where the spurious correlation with color is removed, falls well short of the optimum. 

We also report the results on one of the widely used realistic datasets for OOD generalization: VLCS \citep{fang2013unbiased}. This dataset consists of $10,729$ photographic images of dimension $(3, 224, 224)$ and $5$ classes from four domains: Caltech101, LabelMe, SUN09, and VOC2007. We used the exact experimental setting that is described in \citet{gulrajani2020search}. We provide results averaged over all possible train and test environment combination for one of the commonly used hyper-parameter tuning procedure: train domain validation. As shown in Table \ref{tab:vlcs}, iCaRL achieves state-of-the-art performance when compared to those most popular domain generalization alternatives.


\section{Related Work}

\paragraph{Invariant Prediction} Invariant Causal Prediction (ICP), aims to find the \textit{causal feature set} (i.e., all direct causes of a target variable of interest) \citep{peters2015causal} by exploiting the invariance property in causality which has been discussed under the term ``autonomy'', ``modularity'', and ``stability'' \citep{haavelmo1944probability,aldrich1989autonomy,hoover1990logic,pearl2009causality,dawid2010identifying,scholkopf2012causal}. This invariance property assumed in ICP and its nonlinear extension \citep{heinze2018invariant} is limited, because no intervention is allowed on the target variable $\boldsymbol{Y}$. Besides, ICP methods implicitly assume that the variables of interest $\boldsymbol{X}$ are given. The works \cite{magliacane2018domain} and \cite{subbaswamy2019universal} attempt to find invariant predictors that are maximally predictive using conditional independence tests and other graph-theoretic tools, both of which also assume that the $\boldsymbol{X}$ are given and further assume that additional information about the structure over $\boldsymbol{X}$ is known. \citet{mitrovic2020representation} analyze data augmentations in self-supervised learning from the perspective of invariant causal mechanisms. \citet{arjovsky2019invariant} reformulate this invariance as an optimization-based problem, allowing us to learn an invariant data representation from $\boldsymbol{O}$ constrained to be a linear transformation of $\boldsymbol{X}$. The risks of this approach have been discussed in
\cite{rosenfeld2020risks,kamath2021does,nagarajan2020understanding} and its sample complexity is analyzed in \cite{ahuja2020empirical}. 

\paragraph{Domain Generalization} The goal of domain generalization (DG) \citep{DBLP:conf/icml/MuandetBS13} is OOD generalization: learning a predictor that performs well at unseen test domains. Unlike domain adaptation \citep{pan2009survey,ben2007analysis,ben2010theory,crammer2008learning,patel2015visual,zhao2019learning,wilson2020survey,zhang2015multi}, DG assumes that the test domain data are not available during training. One thread of DG is to explore techniques from kernel methods \citep{DBLP:conf/icml/MuandetBS13,niu2015multi,erfani2016robust,li2017domain}.  \citet{DBLP:conf/icml/MuandetBS13} propose a kernel-based optimization algorithm that learns an invariant transformation by minimizing the discrepancy among domains and preventing the loss of relationship between input and output features. Another line of DG work is using end-to-end methods from deep learning: (a) reducing the differences of representations across domains through adversarial or similar techniques \citep{DBLP:conf/iccv/GhifaryKZB15,wang2017select,DBLP:conf/iccv/MotiianPAD17,li2018domain,li2018deep}; (b) projecting out superficial domain-specific statistics to reduce sensitivity to the domain \citep{wang2019learning}; (c) fusing representations from an ensemble of models across domains \citep{ding2017deep,mancini2018best}. Meta-learning can also be applied to domain generalization, by dividing source domains into meta-training and meta-test sets, and aiming for a low generalization error on meta-test sets after training on meta-training sets \citep{balaji2018metareg,dou2019domain,li2018learning,li2019episodic,li2019feature}. Recently, an extensive empirical survey of many DG algorithm \citep{gulrajani2020search} suggested that with current models and data augmentation techniques, plain ERM may be competitive with the state-of-the-art. It is worth noting that \citet{sun2020latent} also propose an approach to learning latent causal factors for prediction. However, their assumptions over the underlying causal graph are restricted due to two reasons: 1) they only consider the scenarios when $\boldsymbol{X}$ and $\boldsymbol{Y}$ are generated concurrently, which excludes the cases in which some part of $\boldsymbol{X}$ could also be affected by $\boldsymbol{Y}$ in some manner; 2) They assume that the causal latent factors $\boldsymbol{X}_p$ and the non-causal latent factors $\boldsymbol{X}_c$ are independent when conditioning on $\boldsymbol{E}$, that is, $\boldsymbol{X}_p \indep \boldsymbol{X}_c | \boldsymbol{E}$. In practice, during model learning, they actually further assume that $X_i \indep X_j | \boldsymbol{E}$ for any $i \neq j$ so that VAE could be leveraged to learn the model. In this sense, their approach has the same issue as the one in iVAE, i.e., unable to deal with the non-factorized cases. This point is also verified in Table \ref{tab:vlcs}, where our approach greatly outperforms theirs.

\section{Looking Forward: The Agnostic Hypothesis}

In this paper, we developed a novel framework to learn invariant predictors from a diverse set of training environments. As we have seen, the key to our approach is how to identify all the direct causes of the outcome, both theoretically and practically. Our implementation is predicated on Assumptions \ref{eq:assumption1}\&\ref{eq:assumption2}, that the data representation can be factorized when conditioning on the outcome and the environment, as shown in Fig. \ref{fig:general}. More importantly, when taking a closer look at Fig. \ref{fig:general}, one more fundamental underlying assumption is that there exist a set of hidden causal factors ($\{X_i\}_{i \in I_{p}}$) affecting both input images ($\boldsymbol{O}$) and class labels ($\boldsymbol{Y}$), that is,
\begin{align}
\boldsymbol{O} \leftarrow \{X_i\}_{i \in I_{p}} \rightarrow \boldsymbol{Y}.
\end{align}  
In other words, $\{X_i\}_{i \in I_{p}}$ are the common causes of both $\boldsymbol{O}$ and $\boldsymbol{Y}$. We call this underlying assumption \textbf{\textit{the Agnostic Hypothesis}}, which is originally presented in \citet{lu2020unifyingview,lu2020imageclassification} and inspired by the discussion on the example of \textit{optical character recognition} in \citet{peters2017elements}. As an example, the Agnostic Hypothesis provides a more generally reasonable way to look at image classification, in comparison to two existing popular but mutually exclusive opinions: \textbf{Causal} (i.e., labels are viewed as an effect of images) and \textbf{Anticausal} (i.e., images are viewed as an effect of labels). In fact, the Agnostic Hypothesis is more highly beneficial to identifying hidden causal factors $X_{p_r}$. Before diving into the reason behind it, for completeness, we first use the previous digit recognition example to briefly summarise the two possible opinions. 

\subsection{Opinion 1: Anticausal}

\begin{figure}[t]
\centering
\includegraphics[width=0.77\textwidth]{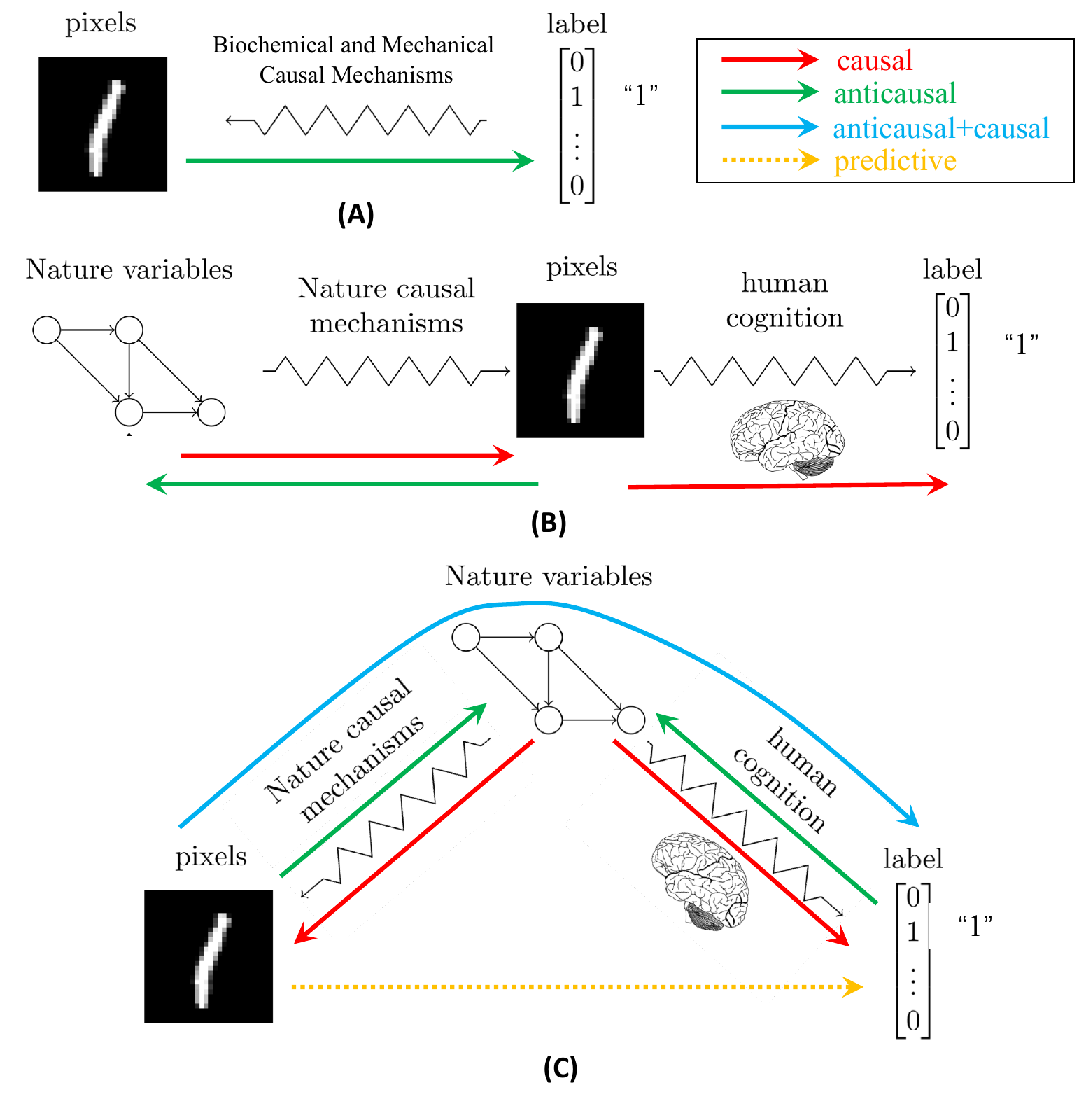}
\caption{Three different opinions on digit recognition. (A) Digit recognition is an anticausal problem, i.e., predicting cause from effect, where images are thought of as the effect of labels. (B) Digit recognition is a causal problem, i.e., predicting effect form cause, where images are thought of as the cause of labels. (C) The Agnostic Hypothesis: there exist a third party of hidden variables (Nature Variables) affecting both images and labels. Note that, for a straightforward comparison, we accordingly modified the figure presented in \cite{arjovsky2019invariant} for each opinion.}
\label{fig:opinions}
\end{figure}

One dominant viewpoint is that digit recognition is an anticausal problem, i.e., predicting cause from effect, where the observed digit image is viewed as an effect of its label which is encoded as an high level concept of interest in human brain \citep{scholkopf2012causal,peters2017elements}. In this case, the process of generating a digit image can be described as follows: the writer first has a high level concept of interest in his mind (i.e., which digit to draw?), and then writes the digit through a complex sequence of biochemical reactions and sensorimotor movements. As such, when trying to predict its labels from an image, we are actually inverting the generating process, with the aim to predict cause (label) from effect (image), as shown in Fig. \ref{fig:opinions}\textcolor{red}{A}. A supporting evidence is that in semi-supervised learning (i.e., an approach to machine learning that integrates a small amount of labeled data with a large amount of unlabeled data during training), an additional set of images usually contribute to the improvement of classification performance. Technically, given training points from $P(X, Y)$ and an additional set of inputs sampled from $P(X)$ where $X$ are images and $Y$ labels, our goal is to estimate $P(Y|X)$. In light of $P(X, Y)=P(Y|X)P(X)$, let us consider two scenarios: (a) If $X \rightarrow Y$, by independence of the mechanism, $P(X)$ is independent of $P(Y|X)$. In other words, $P(X)$ contains no information about $P(Y|X)$, and therefore the additional sample from $P(X)$ will not generally give a better estimate of $P(Y|X)$. (b) If $Y \rightarrow X$, then $P(X)$ is no longer independent of $P(Y|X)$, that is, $P(X)$ provides some information about $P(Y|X)$. In this case, estimating $P(Y|X)$ will definitely benefit from the addition data sampled from $P(X)$. A large number of semi-supervised image classification experiments favor the latter scenario in which $Y$ is a cause and $X$ its effect \citep{scholkopf2012causal}. 

\subsection{Opinion 2: Causal}

An opposite viewpoint was proposed in the concluding dialogue of \cite{arjovsky2019invariant}. They claimed that digit recognition is a causal problem, i.e., predicting effect from cause, which attempts to predict human annotations $Y$ from images $X$ in order to imitate the cognitive process (i.e., humans produce labels by following a causal and cognitive process after observing images), as shown on the RHS of Fig. \ref{fig:opinions}\textcolor{red}{B}. From this perspective, in supervised learning problems about predicting annotations, $P(Y|X)$ should be stable across environments or domains. Hence, the ubiquitous Empirical Risk Minimization principle \citep{vapnik1992principles} should work quite well in this setting and we should not worry too much about it. On the LHS of Fig. \ref{fig:opinions}\textcolor{red}{B}, we can interpret the process from two directions. In the causal direction, Nature Variables (e.g., colour, light, angle, etc.) produce images through Nature causal mechanisms. In the anticausal direction, we attempt to disentangle the underlying causal factors (i.e., Nature Variables) of variation behind images. In the anticausal process, inference might be a more accurate term than disentanglement, because Nature Variables could be dependent on one another in some cases (e.g., the concept of predicates proposed in \cite{vapnik2019complete} can be viewed in some form of dependencies between Nature Variables, not Nature Variables alone, which will be discussed later). 

\subsection{Opinion 3: The Agnostic Hypothesis}

Although it seems that both opinions above make sense to some extent, viewing either images or labels as the cause is not accurate. Technically, in the anticausal view, the learned mapping from images to labels is unstable because it is affected by the image distribution. In the causal view, viewing images as the cause has a conflict with various practical observations that the learned mapping from images to labels is not quite stable across domains and also can be easily fooled or attacked \citep{goodfellow2014explaining} in a large number of real world applications.

According to the analysis above, we propose to look at the digit recognition problem from the agnostic viewpoint (i.e., absence of evidence is not evidence of absence). Precisely, we would like to believe in \textbf{\textit{the Agnostic Hypothesis}} that there must be a third party of hidden causal factors ($Z$), namely \textit{Nature Variables} (e.g., $X_{p_r}$ in our case), affecting both images ($X$) and labels ($Y$). That is, both images and labels are effects of Nature Variables, as shown in Fig. \ref{fig:opinions}\textcolor{red}{C}. We can see that the Agnostic Hypothesis is so general as to include both causal and anticausal cases discussed above. Specifically, if the error in labelling is small (i.e., the error between $Z$ and $Y$ is small), then we can use the labels as a proxy for Nature Variables, in which case the Agnostic Hypothesis is reduced to the anticausal case. Similarly, when the error between $Z$ and $X$ is small, the Agnostic Hypothesis is reduced to the causal case. In general, however, these two extreme cases do not hold true, because even those abstract labels are incapable of accurately capturing the true meaning of Nature Variables due to the unsayable property in language \citep{agamben1999potentialities,wittgenstein2009philosophical}. 

Historically, the Agnostic Hypothesis can be traced at least to the theories below.
\begin{itemize}
\item \textbf{Theory of Forms. } Plato's Theory of Forms \citep{edelstein1966plato} asserts that there are two realms: the physical realm, which is a changing and imperfect world we see and interact with on a daily basis, and the realm of Forms, which exists beyond the physical realm and which is stable and perfect. This theory asserts that the physical realm is only a projection of the realm of Forms. Plato further claimed that each being has five things: the name, the definition, the image, the knowledge, and the Form, the first four of which can be viewed as the projections of the fifth. In this sense, the Agnostic Hypothesis is explicitly consistent with Plato's theory.
\item \textbf{Manipulability Theory. } It provides a paradigmatic assertion in causal relationships that manipulation of a cause will result in the manipulation of an effect. In other words, causation implies that by varying one factor, we can make another vary \citep{cook1979experimental}. In digit recognition, it makes sense that image causes label because changes in image will apparently lead to changes in label. Nevertheless, the converse that label causes image is also reasonable in that changes in label will definitely result in changes in images. Since it is clear that there is no causal loop, be it temporal or not, between images and labels, there must exist something hidden behind. Otherwise, it will violate the manipulability theory. Hence, the theory indirectly supports the Agnostic Hypothesis. 
\item \textbf{Principle of Common Cause. }  The relation between association and causation was formalized in \cite{reichenbach1991direction}: If two random variables $X$ and $Y$ are statistically dependent, then one of the following causal explanations must hold: a) $X$ causes $Y$; b) $Y$ causes $X$; c) there exists a random variable $Z$ that is the common cause of both $X$ and $Y$. This principle is directly in favor of the Agnostic Hypothesis. Because the previous analysis shows that there is no convincing evidence in full support of either image causes label or label causes image, there must be the third case in which there exists a hidden variable affecting both image and label.
\end{itemize}

Now let us investigate whether or not the Agnostic Hypothesis can give a better explanation of image classification. 

Firstly, we can think of image and label as two different representation spaces projected from Nature Variables via Nature causal mechanism and human cognitive mechanism, respectively. In other words, image is the way Nature interprets Nature Variables whilst label is the way humans interpret them. Hence, both are the effects of Nature Variables, but on different levels (i.e., label is more abstract than image). 

Secondly, in light of the fork junction (i.e., $X \leftarrow Z \rightarrow Y$), we can claim two things: (1) Given Nature Variables, image is independent of label; (2) If Nature Variable is hidden (i.e., not given), image is NOT independent of label. Both indeed make sense. The former states that image can be perfectly explained by Nature Variables without the help of label, vice versa (see the red arrows in Fig. \ref{fig:opinions}\textcolor{red}{C}). The latter explains why it is meaningful to predict label from image, and actually the reason holds in most supervised learning applications (see the yellow dashed arrow in Fig. \ref{fig:opinions}\textcolor{red}{C}). 

Thirdly, almost all the existing approaches to digit recognition, or more general image classification, consist of two parts: a feature extractor and a classifier whether explicitly or implicitly. For example, traditional approaches are usually explicitly made up of a handcrafted feature extractor and a well-chosen classifier. By contrast, deep learning approaches appear more implicit, where deep neural networks can always be split into two components: the first of which can be viewed as a feature extractor and the second as a classifier. In effect, this setting can be well explained by the blue arrow in Fig. \ref{fig:opinions}\textcolor{red}{C} that is composed of an anticausal process from image to Nature Variables and a causal process from Nature Variables to label. We can think of the anticausal part as a feature extractor and the causal part as a classifier. 

\textbf{Inspired by the points above, we argue that Nature Variables are the key to the OOD generalization in invariant risk minimization}. Here we conceptually explain why this argument universally makes sense. Without loss of generality, we continue taking digit/image classification for example. Ideally, if we have a perfect feature extractor, then we can infer Nature Variables from images, which will, to the largest extent, reduce the negative influence from irrelevant noisy information on the input. In other words, the perfect feature extractor can defeat the attack on the input so as to guarantee the OOD generalization of the systems in terms of the input information. Furthermore, because Nature Variables are causal parents of labels, the learned classifier based on Nature Variables should be invariant across environments or domains as discussed in \citet{arjovsky2019invariant} and \citet{peters2016causal}, which guarantees the OOD generalization of the systems in terms of the output information. More importantly, once Nature Variables are obtained, we would know how they influence one another and how they affect both images and labels. It would thus render the systems' behaviours more interpretable.  

\begin{figure}[t]
\centering
\includegraphics[width=0.6\linewidth]{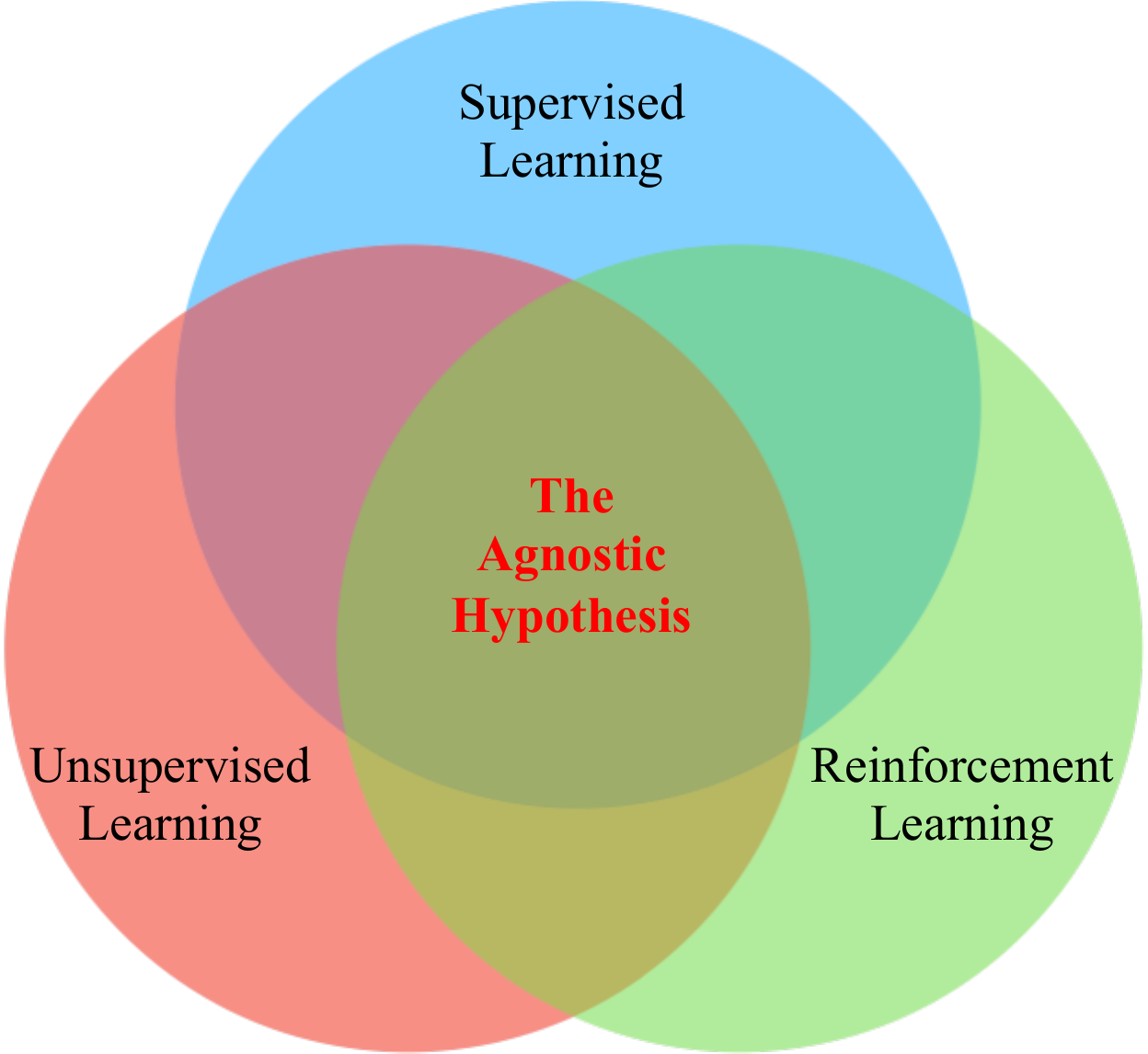}
\caption{The Agnostic Hypothesis provides a unifying view of machine learning.}
\label{fig:unifyingview}
\end{figure}

\subsection{A Unifying View of Machine Learning}

All the explanations above are not limited to digit recognition and also apply to general machine learning problems. In fact, the Agnostic Hypothesis provides a unifying view of machine learning as shown in Fig. \ref{fig:unifyingview}, which paves the way for inspiring both new algorithm designs and a new theory of machine learning. In this section, we first re-interpret machine learning, which are widely categorised as supervised, unsupervised, or reinforcement learning, under the Agnostic Hypothesis. Then, we discuss the implications of such a unifying view and its practical consequences, which are substantiated by the experimental results found in the literature. 

\subsubsection{Supervised Learning} When the training data contains examples of what the correct output should be for given inputs, as aforementioned, under the Agnostic Hypothesis the inputs and the correct outputs can be thought of as two different representation spaces projected from Nature Variables via two different mechanisms. In other words, inputs and outputs are two distinct interpretations or views of Nature Variables.

\subsubsection{Unsupervised Learning} When the training data just contains inputs without any output information, there are several scenarios. If the input data come only from one domain, under the Agnostic Hypothesis they are interpreted as just one projection or view of Nature Variables. If they are from multiple transforms or variants of the same domain \citep{chen2020simple}, they are naturally viewed as multiple projections or views of Nature Variables.

\subsubsection{Reinforcement Learning} A reinforcement learning agent is learning what to do, i.e., how to map situations to actions, so as to maximise a numerical reward signal \citep{sutton2018reinforcement}. Rather than being directly told which action to take, the agent must learn from sequences of observed information (i.e., states/observations in a fully/partially observable environment, actions, and rewards). Under the Agnostic Hypothesis, each component of the observed information reflects one aspect of the environment and can be thus viewed as one projection or view of Nature Variables \footnote{Although uncovering structure in an agent's experience itself, whose perfect representation should be undoubtedly in some form of Nature Variables, does not address the reinforcement learning problem of maximising a reward signal, it plays a vital role in real world reinforcement learning applications with strong demand for the OOD generalization \citep{garcia2015comprehensive,gros2020safe}.}.

\subsubsection{Implications of a Unifying View} The reason why we should look at machine learning from the unifying view of the Agnostic Hypothesis is that compared to the traditional view, it offers a promising way to explore the requirements of machine learning for the OOD generalization in the real world applications. As stated in \cite{russell2002artificial}, the representation of the learned information plays a pivotal role in determining how the learning algorithm must work. Hence, from the unifying view, all the required properties of machine learning are rooted in how accurately we can estimate Nature Variables from data, because Nature Variables are, without doubt, the optimal representation of the data.

Now the question comes to whether or not it is possible to identify Nature Variables from data. Considering that under the Agnostic Hypothesis the data in reality are viewed as projections or views of Nature Variables, one naturally has an intuition that the more diverse views of Nature Variables we have, the more accurate estimate of Nature Variables we can attain. For convenience, we call this intuition \textbf{\textit{Multiview Universal Approximation}} (\textbf{MUA}). Not surprisingly, some theoretical works have demonstrated MUA to some extent by showing that under some assumptions multiple views will lead to identifiability of Nature Variables up to some affine ambiguity \citep{gresele2019incomplete}. Actually, under some stronger assumptions only two views are even capable of identifying Nature Variables up to some unavoidable indeterminacy \citep{hyvarinen2019nonlinear,gresele2019incomplete}. Although these initial works are predicated on strict assumptions on Nature Variables (e.g., Nature Variables are assumed to be independent or conditionally independent of each other, which, as aforementioned, is unnecessary in reality, etc.), they are a good starting point on the road towards the general theory of MUA. 

From this point of view, the unifying view based on the Agnostic Hypothesis has at least four practical implications. 

a) It can help identify the issues in current machine learning algorithms. For example, in the research of adversarial attacks \citep{goodfellow2014explaining}, the reason why it is far too easy to fool convolutional networks with an imperceivable but carefully constructed noise in the input is that the feature extractor part of the networks cannot accurately infer the Nature Variables in the anticausal direction. Hence, the learned predictive link between image and label is so unstable that a small disturbance on the input image will lead to wrong Nature Variables misguiding the classifier part. This issue widely exists in supervised learning, because at the testing time only one view is used to infer Nature Variables. This issue could be mitigated if the input data involve multiple views in some scenarios, such as in time series prediction problems that take as input multiple time step data and each time step can be viewed as one view of Nature Variables.

b) It can help understand why algorithms work. For instance, it is reasonable that multitask learning \citep{caruana1997multitask} has been used successfully across all applications of machine learning, because multitask data provide multiple views of their shared features (Nature Variables), making inferring them more accurate as suggested in MUA. Another example is that in reinforcement learning, one widely leveraged multiview data to discover the invariant part of states \citep{lu2018deconfounding, zhang2020invariant}. 

c) It can help inspire a new algorithm design. It is worth noting that an unsupervised learning approach, proposed in a very recent work \citep{chen2020simple}, leverages data augmentation to considerably outperform previous methods for self-supervised and semi-supervised learning and even to be on a par with supervised learning methods on ImageNet. It indeed makes sense, because data augmentation created multiple views of latent features (Nature Variables), leading to identifying Nature Variables more accurately as stated in MUA. 

d) It can help inspire a new theory of machine learning. As mentioned previously, current machine learning theories neither satisfy the demands for the OOD generalization in real world applications nor answer the key question of how to discover Nature Variables from data \citep{vapnik2019complete}. The unifying view provides a promising way to address both issues by developing a general theory of MUA with more relaxed assumptions, because under the Agnostic Hypothesis MUA is a natural and feasible way to identify Nature Variables as mentioned previously. Another possible thread of discovering them is through intervention \citep{pearl2009causality} if it is allowed to interact with environments. 

\subsection{Broader Discussion}

The Agnostic Hypothesis can be viewed as a kind of description of the relationship between invariants and variants, where invariants are reflected by Nature Variables and variants by their corresponding projections or views. In this sense, the Agnostic Hypothesis has a long history in philosophy and science. It can be dated back to the time of Plato, whose Theory of Forms described the relationship as aforementioned. Hegel argued that there exists a higher level of cognition commonly taken as capable of having purportedly eternal contents (i.e., invariants) which come from the changing contents (i.e., variants) based in everyday perceptual experience. Furthermore, Wigner summarised physics as a process of discovering the laws of inanimate nature, that is, recognising invariance from the world of baffling complexity around us \citep{wigner1990unreasonable}. Recently, Vapnik was motivated to propose a statistical theory of learning based on statistical invariants constructed using training data and given predicates \citep{vapnik2019complete}. Roughly speaking, a predicate is a function revealing some invariant property of the world of interest, like the Form in Plato's theory.

Although a number of philosophers agree that there exist such invariants beyond the variants, there is no consensus on whether or not they are apprehensible. For example, Kant thought that there is some unknowable invariant, called thing-in-itself, outside of all possible human experience \citep{kant1998critique}. However, Schopenhauer believed that the supreme invariant principle of the universe is likewise apprehensible through introspection, and that we can understand the world as various manifestations of this general principle \citep{schopenhauer2012world}. Despite the controversy in philosophy, Vapnik still provided two examples in art to show that it might be possible to comprehend those invariants to some extent \citep{vapnik2020youtube}. One is that Bach's music is full of repeated patterns. The other is that Vladimir Propp, a Soviet formalist scholar, analyzed the basic plot components of Russian folk tales to identify 31 simplest irreducible narrative elements which are so general that they also apply to many other stories and movies. Both seemingly demonstrate that there exist such invariants at least in some form. It is thus natural to ask how we can identify them from data, which is so important that Vapnik thought that the essence of intelligence is the discovery of good predicates \citep{vapnik2020youtube}. Schmidhuber once expressed a similar opinion that ``all the history of science is the history of compression progress'', where obviously the optimal compression should be in the form of Nature Variables \citep{schmidhuber2020youtube}. Bengio proposed a "consciousness prior" for learning representations of high-level concepts \citep{bengio2017consciousness}. We hope that the Agnostic Hypothesis can provide inspiration to explore the general theory of MUA for identifying Nature Variables both theoretically and practically, which is key to enabling OOD generalization guarantees in machine learning.

\section*{Acknowledgements}

We are thankful to Wenlin Chen for contributing to Step III in the proof of Theorem \ref{thm:4} and for generalizing the proof in Theorem \ref{thm:5} to the case in which
each $\boldsymbol{T}_i(X_i)$ in $\boldsymbol{T}_f(\boldsymbol{X})$
contains arbitrary sufficient statistics instead of just $X_i$ and $X_i^2$. We also thank Ilyes Khemakhem and Aapo Hyv{\"a}rinen for his helpful discussions and comments, and an anonymous reviewers of an earlier version of this paper for pointing out an issue with our use of the iVAE identifiability result.

\bibliography{reference}
\bibliographystyle{apalike}


\newpage

\appendix


\section{Variational Autoencoders} \label{appendix:vae}

We briefly describe the framework of variational autoencoders (VAEs), which allows us to efficiently learn deep latent-variable models and their corresponding inference models \citep{kingma2013auto,rezende2014stochastic}. Consider a simple latent variable model where $\boldsymbol{O} \in \mathbb{R}^d$ stands for an observed variable and $\boldsymbol{X} \in \mathbb{R}^n$ for a latent variable. A VAE method learns a full generative model $p_{\boldsymbol{\theta}}(\boldsymbol{O}, \boldsymbol{X})=p_{\boldsymbol{\theta}}(\boldsymbol{O}|\boldsymbol{X})p_{\boldsymbol{\theta}}(\boldsymbol{X})$ and an inference model $q_{\boldsymbol{\phi}}(\boldsymbol{X}|\boldsymbol{O})$, typically a factorized Gaussian distribution whose mean and variance parameters are given by the output of a neural network with input $\boldsymbol{O}$. This inference model approximates the posterior $p_{\boldsymbol{\theta}}(\boldsymbol{X}|\boldsymbol{O})$, where $\boldsymbol{\theta}$ is a vector of parameters of the generative model, $\boldsymbol{\phi}$ a vector of parameters of the inference model, and $p_{\boldsymbol{\theta}}(\boldsymbol{X})$ is a prior distribution over the latent variables. Instead of maximizing the data log-likelihood, we maximize its lower bound $\mathcal{L}_{\text{VAE}}(\boldsymbol{\theta}, \boldsymbol{\phi})$:
\begin{align}
\log p_{\boldsymbol{\theta}}(\boldsymbol{O}) \geq \mathcal{L}_{\text{VAE}}(\boldsymbol{\theta}, \boldsymbol{\phi}) := \mathbb{E}_{q_{\boldsymbol{\phi}}(\boldsymbol{X}|\boldsymbol{O})}\left[\log p_{\boldsymbol{\theta}}(\boldsymbol{O} | \boldsymbol{X})\right] - \text{KL}\left(q_{\boldsymbol{\phi}}(\boldsymbol{X}|\boldsymbol{O})||p_{\boldsymbol{\theta}}(\boldsymbol{X})\right), \nonumber
\end{align}
where we have used Jensen's inequality, and $\text{KL}(\cdot||\cdot)$ denotes the Kullback-Leibler divergence between two distributions.


\section{Derivation} \label{appendix:derivation}

\subsection{iVAE}

In Section \ref{sect:ivae}, the evidence lower bound of iVAE is defined by
\begin{align}
&\mathcal{L}_{\text{iVAE}}(\boldsymbol{\theta}, \boldsymbol{\phi}) \nonumber \\
:= & \mathbb{E}_{p_D}\left[\mathbb{E}_{q_{\boldsymbol{\phi}}(\boldsymbol{X}|\boldsymbol{O}, \boldsymbol{U})}\left[\log p_{\boldsymbol{f}}(\boldsymbol{O} | \boldsymbol{X})\right] - \text{KL}\left(q_{\boldsymbol{\phi}}(\boldsymbol{X}|\boldsymbol{O},\boldsymbol{U})||p_{\boldsymbol{T}, \boldsymbol{\lambda}}(\boldsymbol{X}|\boldsymbol{U})\right)\right] \nonumber\\
= & \mathbb{E}_{p_D}\left[\mathbb{E}_{q_{\boldsymbol{\phi}}(\boldsymbol{X}|\boldsymbol{O}, \boldsymbol{U})}\left[\log p_{\boldsymbol{f}}(\boldsymbol{O}|\boldsymbol{X}) + \log p_{\boldsymbol{T}, \boldsymbol{\lambda}}(\boldsymbol{X}|\boldsymbol{U}) -\log q_{\boldsymbol{\phi}}(\boldsymbol{X}|\boldsymbol{O}, \boldsymbol{U})\right]\right]. \nonumber
\end{align}


\subsection{Score Matching} \label{appendix:dev_sm}

In Section \ref{sect:phase1}, we follow \citet{hyvarinen2005estimation} and use a simple trick of partial integration to simplify the evaluation of the score matching objective $\mathcal{L}^{\text{SM}}_{\text{phase1}}$ in Eq. (10) of the main text:
{\small
\begin{align}  
& \mathcal{L}^{\text{SM}}_{\text{phase1}}(\boldsymbol{T}, \boldsymbol{\lambda}) \nonumber \\
:= & \mathbb{E}_{p_D}\left[\mathbb{E}_{q_{\boldsymbol{\phi}}(\boldsymbol{X}|\boldsymbol{O}, \boldsymbol{Y}, \boldsymbol{E})}\left[||\nabla_{\boldsymbol{X}} \log q_{\boldsymbol{\phi}}(\boldsymbol{X}|\boldsymbol{O}, \boldsymbol{Y}, \boldsymbol{E}) - \nabla_{\boldsymbol{X}} \log p_{\boldsymbol{T}, \boldsymbol{\lambda}}(\boldsymbol{X}|\boldsymbol{Y}, \boldsymbol{E})||^2
\right]\right] \nonumber \\
= & \mathbb{E}_{p_D}\left[\mathbb{E}_{q_{\boldsymbol{\phi}}(\boldsymbol{X}|\boldsymbol{O}, \boldsymbol{Y}, \boldsymbol{E})}\left[
\sum_{j=1}^n \left[
\frac{\partial^2 p_{\boldsymbol{T}, \boldsymbol{\lambda}}(\boldsymbol{X}| \boldsymbol{Y}, \boldsymbol{E})}{\partial X_j^2} 
+ \frac{1}{2}\left(
\frac{\partial p_{\boldsymbol{T}, \boldsymbol{\lambda}}(\boldsymbol{X}| \boldsymbol{Y}, \boldsymbol{E})}{\partial X_j}\right)^2
\right]
\right]\right] + const. \nonumber
\end{align}}
where the last equality is due to the Theorem 1 in \citet{hyvarinen2005estimation}.

\section{Definitions for SEM and IRM} \label{appendix:dfn}

For convenience, we restate some definitions here and refer to the original papers \citep{arjovsky2019invariant,peters2017elements} for more details.

\begin{dfn}
A structural equation model (SEM) $\mathcal{C}:=(\mathcal{S}, N)$ governing the random vector $\boldsymbol{X}=(X_1, \ldots, X_d)$ is a set of structural equations:
\begin{align}
\mathcal{S}_i: X_i \leftarrow f_i(\text{\normalfont Pa}(X_i), N_i), \nonumber
\end{align}
where $\text{\normalfont Pa}(X_i) \subseteq \{X_1, \ldots, X_d\} \setminus \{X_i\}$ are called the parents of $X_i$, and the $N_i$ are independent noise random variables. We say that ``$X_i$ causes $X_j$'' if $X_i \in \text{\normalfont Pa}(X_j)$. We call causal graph of $\boldsymbol{X}$ to the graph obtained by drawing i) one node for each $X_i$, and ii) one edge from $X_i$ to $X_j$ if $X_i \in \text{\normalfont Pa}(X_j)$. We assume acyclic causal graphs.
\end{dfn}

\begin{dfn}
Consider a SEM $\mathcal{C}:=(\mathcal{S}, N)$. An intervention $e$ on $\mathcal{C}$ consists of replacing one or several of its structural equations to obtain an intervened SEM $\mathcal{C}^e:=(\mathcal{S}^e, N^e)$, with structural equations:
\begin{align}
\mathcal{S}_i^e: X_i^e \leftarrow f_i^e(\text{\normalfont Pa}^e(X_i^e), N_i^e), \nonumber
\end{align}
The variable $\boldsymbol{X}^e$ is intervened if $\mathcal{S}_i \neq \mathcal{S}_i^e$ or $N_i \neq N_i^e$.
\end{dfn}

\begin{dfn} \label{dfn: invariance}
Consider a structural equation model (SEM) $\mathcal{S}$ governing the random vector $(X_1, \ldots, X_n, \boldsymbol{Y})$, and the learning goal of predicting $\boldsymbol{Y}$ from $\boldsymbol{X}$. Then, the set of all environments $\mathcal{E}_{all}(\mathcal{S})$ indexes all the interventional distributions $P(\boldsymbol{X}^e, \boldsymbol{Y}^e)$ obtainable by valid interventions $e$. An intervention $e \in \mathcal{E}_{all}(\mathcal{S})$ is valid as long as (i) the causal graph remains acyclic, (ii) $\mathbb{E}\left[\boldsymbol{Y}^e|\text{\normalfont Pa}(\boldsymbol{Y})\right]=\mathbb{E}\left[\boldsymbol{Y}|\text{\normalfont Pa}(\boldsymbol{Y})\right]$, and (iii) $\mathbb{V}\left[\boldsymbol{Y}^e|\text{\normalfont Pa}(\boldsymbol{Y})\right]$ remains within a finite range.
\end{dfn}


\section{Definitions and Lemmas for the Exponential Families}

\begin{dfn}(\textbf{Exponential family}) 
A multivariate exponential family is a set of distributions whose probability density function can be written as 
\begin{align}
    p(\boldsymbol{X})
    = \frac{\mathcal{Q}(\boldsymbol{X})}{\mathcal{Z}(\boldsymbol{\theta})}\exp(\left<\boldsymbol{T}(\boldsymbol{X}),\boldsymbol{\theta}\right>),
\end{align}
where $\mathcal{Q}:\mathcal{X}\to\mathbb{R}$ is the base measure, $\mathcal{Z}(\boldsymbol{\theta})$ is the normalizing constant, $\boldsymbol{T}:\mathcal{X}\to\mathbb{R}^k$ is the sufficient statistics, and $\boldsymbol{\theta} \in \mathbb{R}^k$ is the natural parameter. The size $k\geq n$ is the dimension of the sufficient statistics $\boldsymbol{T}$ and depends on the latent space dimension $n$. Note that $k$ is fixed given $n$.
\end{dfn}

\begin{dfn}(\textbf{Strongly exponential distributions}) \label{def:strong_ex}
    A multivariate exponential family distribution
    \begin{align}
        p(\boldsymbol{X})
        = \frac{\mathcal{Q}(\boldsymbol{X})}{\mathcal{Z}(\boldsymbol{\theta})}\exp(\left<\boldsymbol{T}(\boldsymbol{X}),\boldsymbol{\theta}\right>)
\end{align}
    is strongly exponential, if
    \begin{multline}
        (\exists\boldsymbol{\theta}\in\mathbb{R}^k~s.t.~\left<\boldsymbol{T}(\boldsymbol{X}),\boldsymbol{\theta}\right>=const,~\forall \boldsymbol{X}\in\mathcal{X}) \\ \implies (l(\mathcal{X})=0~\text{or}~\boldsymbol{\theta}=\mathbf{0}),~\forall\mathcal{X}\subset\mathbb{R}^n,
    \end{multline}
     where $l$ is the Lebesgue measure. 
\end{dfn}
\noindent The density of a strongly exponential distribution has almost surely the exponential component and can only be reduced to the base measure on a set of measure zero. Note that all common multivariate exponential family distributions (e.g. multivariate Gaussian) are strongly exponential.


\section{Definitions for Identifiability} \label{appendix:identifiability}

\begin{dfn}
Let $\Theta$ be the domain of the parameters $\boldsymbol{\theta}=\{\boldsymbol{f},\boldsymbol{T},\boldsymbol{\lambda}\}$. Let $\sim$ be an equivalence relation on $\Theta$. A deep generative model is said to be $\sim$--identifiable if
\begin{align}
    p_{\boldsymbol{\theta}}(\boldsymbol{O})=p_{\Tilde{\boldsymbol{\theta}}}(\boldsymbol{O}) \implies \boldsymbol{\theta}\sim\Tilde{\boldsymbol{\theta}}.
\end{align}
The elements in the quotient space $\Theta\backslash\sim$ are called the identifiability classes.
\end{dfn}

\begin{dfn} \label{dfn:A_idf}
Let $\sim_A$ be an equivalence relation on $\Theta$ defined by:
\begin{multline}
    (\boldsymbol{f},\boldsymbol{T},\boldsymbol{\lambda})\sim_A(\Tilde{\boldsymbol{f}},\Tilde{\boldsymbol{T}},\Tilde{\boldsymbol{\lambda}}) \\
    \iff \exists A,\boldsymbol{c}~\text{s.t.}~ \boldsymbol{T}(\boldsymbol{f}^{-1}(\boldsymbol{O}))=A\Tilde{\boldsymbol{T}}(\Tilde{\boldsymbol{f}}^{-1}(\boldsymbol{O}))+\boldsymbol{c},\forall \boldsymbol{O}\in\mathcal{O},
\end{multline}
where $A\in\mathbb{R}^{k\times k}$ is an invertible matrix and $\boldsymbol{c}\in\mathbb{R}^k$ is a vector.
\end{dfn}

\begin{dfn} \label{dfn:P_idf}
    Let $\sim_P$ be an equivalence relation on $\Theta$ defined by:
    \begin{multline}
        (\boldsymbol{f},\boldsymbol{T},\boldsymbol{\lambda})\sim_P(\Tilde{\boldsymbol{f}},\Tilde{\boldsymbol{T}},\Tilde{\boldsymbol{\lambda}}) \\
        \iff \exists P,\boldsymbol{c}~\text{s.t.}~ \boldsymbol{T}(\boldsymbol{f}^{-1}(\boldsymbol{O}))=P\Tilde{\boldsymbol{T}}(\Tilde{\boldsymbol{f}}^{-1}(\boldsymbol{O}))+\boldsymbol{c},\forall \boldsymbol{O}\in\mathcal{O},
    \end{multline}
    where $P\in\mathbb{R}^{k\times k}$ is a block permutation matrix and $\boldsymbol{c}\in\mathbb{R}^k$ is a vector.
\end{dfn}

\input{fig2.tex}

\section{Phase 2: Discovering Single Cause} \label{appendix:phase2_single}

In the single-cause case, by following \citet{wang2014discovering}, we leverage the \textit{MB-by-MB} algorithm to first construct a \textit{local} structure around $\boldsymbol{Y}$ and then discover the single parent of $\boldsymbol{Y}$ according to the constructed \textit{local} graph. One obvious advantage of this approach is in efficiency, because there is no need to construct the whole causal graph containing all the latent variables and $\boldsymbol{Y}$.

We could have an even more efficient solution to the single-cause case in some special scenarios where we assume that $X_i \indep X_j | \boldsymbol{Y}, \boldsymbol{E}$ for any $i \neq j$. In fact, this assumption covers more scenarios than the common assumption that $X_i \indep X_j$ for $i \neq j$ in latent variable models, e.g., disentanglement \citep{bengio2013representation,locatello2019challenging}, autoencoders \citep{kingma2013auto,rezende2014stochastic}, ICA \citep{comon1994independent,hyvarinen2000independent}, etc. If $\boldsymbol{Y}$ is caused by at most one of $X_i$ and $X_j$, and no matter whether $X_i$ and $X_j$ are caused by $\boldsymbol{E}$ or not, then  $X_i \indep X_j | \boldsymbol{Y}, \boldsymbol{E}$ holds, but we may well have $X_i \notindep X_j$ (e.g., if $\boldsymbol{Y}$ causes both $X_i$ and $X_j$, or if there is a chain $X_i\rightarrow \boldsymbol{Y} \rightarrow X_j$). If $\boldsymbol{Y}$ causes or is caused by at most one of $X_i$ and $X_j$, and at most one of $X_i$ and $X_j$ is caused by $\boldsymbol{E}$, then both $X_i \indep X_j$ and $X_i \indep X_j | \boldsymbol{Y}, \boldsymbol{E}$ hold. If $\{\boldsymbol{Y}, X_i, X_j\}$ form a collider $X_i \rightarrow \boldsymbol{Y} \leftarrow X_j$, and no matter whether $X_i$ and $X_j$ are caused by $\boldsymbol{E}$ or not, then $X_i \indep X_j | \boldsymbol{E}$ hold, but we may have $X_i \notindep X_j$ (e.g., when both $X_i$ and $X_j$ are caused by $\boldsymbol{E}$).

Under this assumption, we are able to separately look into each $X_i$ given $\boldsymbol{Y}$ and $\boldsymbol{E}$, without considering any other $X_j$. Fig.~\ref{fig:2a} shows that there exist only five possible connections between $X_i$, $\boldsymbol{Y}$, $\boldsymbol{E}$, and $\boldsymbol{O}$. Among them, only the arrow from $X_i$ to $\boldsymbol{O}$ must exist, because $\boldsymbol{O}$ is generated from $\boldsymbol{X}$. The other four arrows might not be present, with the exception that there must be at least one connection between $X_i$ and $\boldsymbol{Y}$ or $\boldsymbol{E}$ (Assumption~\ref{eq:assumption1}a). This leaves ten possible structures, shown in Figs.~\ref{fig:2c}-\ref{fig:2m}. 

Given data $\{\hat{X}_i, \boldsymbol{Y}, \boldsymbol{E}, \boldsymbol{O}\}$ in which the $\hat{X}_i$ are obtained using $q_{\boldsymbol{\phi}}(\boldsymbol{X}|\boldsymbol{O}, \allowbreak \boldsymbol{Y}, \boldsymbol{E})$ (for example, as given by the mean of this distribution), we are able to distinguish all ten structures in
Figs.~\ref{fig:2c}-\ref{fig:2m}
by using causal discovery algorithms \citep{peters2017elements,zhang2017causal,huang2020causal} and
performing conditional independence tests \citep{spirtes2000causation,zhang2012kernel}. This is summarized in Proposition \ref{prop:1} below. Its proof can be found in Appendix \ref{appendix:proof}, which also describes the specific assumptions made. 
In practice, we can assess in parallel whether or not each $X_i$ is a direct cause of $\boldsymbol{Y}$, which accelerates this phase significantly.
\begin{ppn} \label{prop:1}
Under the assumptions stated in Appendix \ref{appendix:proof},
the ten structures shown in Figs.~\ref{fig:2c}-\ref{fig:2m} can be identified  using causal discovery methods consistent in the infinite sample limit.
\end{ppn}

Note that there are only four cases in which $X_i$ is a parent of $\boldsymbol{Y}$ (i.e., Figs.~\ref{fig:2c}, \ref{fig:2f}, \ref{fig:2i}, and \ref{fig:2l}). 
We can identify these by applying rules \textbf{1.2}, \textbf{1.6}, \textbf{2.1}, and \textbf{3.1} from Appendix \ref{appendix:proof4}.


\section{Proofs} \label{appendix:proof}


\subsection{Proof of Theorem \ref{thm:1}}

The proof of this theorem consists of three parts. 

In Part I, we prove that the parameters $\boldsymbol{\theta}$ are $\sim_A$ identifiable (Definition \ref{dfn:A_idf}) by using assumption (i), the first half of assumption (ii), and assumption (iv) of Theorem \ref{thm:1}.

In Part II, based on the result in Part I, we further prove that the parameters $\boldsymbol{\theta}$ are $\sim_P$ identifiable (Definition \ref{dfn:P_idf}) by additionally using Assumption 2, the second half of assumption (ii) and assumption (iii) of Theorem \ref{thm:1}.

In Part III, we combine the results (Theorems \ref{thm:4}\&\ref{thm:5}) in both Part I and Part II into one theorem (Theorem \ref{thm:1}), which completes the proof.

It is worth noting that, compared to the proof in iVAE, the main changes in our proof consist of
\begin{itemize}
    \item Part I, In step III.
        \begin{itemize}
            \item It has been updated to account for vectors of sufficient statistics whose entries can be arbitrary functions of all entries in the random variable vector, while in the previous proof the sufficient statistics contained entries that are functions of individual entries in the random variable vector.
            \item The assumption of ``The sufficient statistics in $\boldsymbol{T}$ are all linearly independent.'' is not required in our proof, but it is in the proof of iVAE.
        \end{itemize}
    \item Part II.
        \begin{itemize}
            \item It has been updated to account for the part of the sufficient statistics which is the output of a deep neural network with ReLU activation functions.
        \end{itemize}
\end{itemize}

\subsubsection{Part I} 

For notational simplicity, in the proof of this part we denote $(\boldsymbol{Y}, \boldsymbol{E})$ by $\boldsymbol{U}$. Hence, our generative model defined according to Eqs. (6-8) in the main text now becomes:
\begin{align} 
p_{\boldsymbol{\theta}}(\boldsymbol{O}, \boldsymbol{X}|\boldsymbol{U}) &= p_{\boldsymbol{f}}(\boldsymbol{O}|\boldsymbol{X})p_{\boldsymbol{T}, \boldsymbol{\lambda}}(\boldsymbol{X}|\boldsymbol{U}), \label{eq:thm4_gen}\\
p_{\boldsymbol{f}}(\boldsymbol{O}|\boldsymbol{X}) &= p_{\boldsymbol{\epsilon}}(\boldsymbol{O}-\boldsymbol{f}(\boldsymbol{X})), \label{eq:thm4_likelihood} \\
p_{\boldsymbol{T}, \boldsymbol{\lambda}}(\boldsymbol{X}|\boldsymbol{U}) &= {\mathcal{Q}(\boldsymbol{X})}/{\mathcal{Z}(\boldsymbol{U})} \exp\big[ \boldsymbol{T}(\boldsymbol{X})^\text{T} \boldsymbol{\lambda}(\boldsymbol{U}) ].\label{eq:thm4_prior}
\end{align} 

\begin{thm} \label{thm:4}
    Suppose that we observe data sampled from a deep generative model defined according to Eqs. (\ref{eq:thm4_gen}-\ref{eq:thm4_prior}) with parameters $(\boldsymbol{f},\boldsymbol{T},\boldsymbol{\lambda})$. Assume that
    \begin{enumerate}[(i)]
    \item The set $\{\boldsymbol{O}\in\mathbb{O}|\varphi_{\boldsymbol{\varepsilon}}(\boldsymbol{O})=0\}$ has measure zero, where $\varphi_{\boldsymbol{\varepsilon}}$ is the characteristic function of the density $p_{\boldsymbol{\varepsilon}}$ defined in Eq. (\ref{eq:thm4_likelihood});
    \item The mixing function $\boldsymbol{f}$ in Eq. (\ref{eq:thm4_likelihood}) is injective;
    \item There exist $k+1$ points $\boldsymbol{U}^0,\boldsymbol{U}^1,\cdots,\boldsymbol{U}^k\in\mathcal{U}$ such that the matrix
    \begin{align}
        L=[\boldsymbol{\lambda}(\boldsymbol{U}^1)-\boldsymbol{\lambda}(\boldsymbol{U}^0),\cdots,\boldsymbol{\lambda}(\boldsymbol{U}^k)-\boldsymbol{\lambda}(\boldsymbol{U}^0)]\in\mathbb{R}^{k\times k}
    \end{align}
    is invertible.
    \end{enumerate}
    Then the parameters $\{\boldsymbol{f},\boldsymbol{T},\boldsymbol{\lambda}\}$ are $\sim_A$ identifiable.
\end{thm}

\begin{proof}
Define $\vol(B)=\sqrt{\det(B^TB)}$ for any full column rank matrix $B$. Suppose that we have two sets of parameters $\boldsymbol{\theta}=(\boldsymbol{f},\boldsymbol{T},\boldsymbol{\lambda})$ and $\Tilde{\boldsymbol{\theta}}=(\Tilde{\boldsymbol{f}},\Tilde{\boldsymbol{T}},\Tilde{\boldsymbol{\lambda}})$ such that $p_{\boldsymbol{\theta}}(\boldsymbol{O}|\boldsymbol{U})=p_{\Tilde{\boldsymbol{\theta}}}(\boldsymbol{O}|\boldsymbol{U}),~\forall (\boldsymbol{O},\boldsymbol{U})\in\mathcal{O}\times\mathcal{U}$. We want to show $\boldsymbol{\theta}\sim_A\Tilde{\boldsymbol{\theta}}$.
    
\paragraph{Step I.} The proof of this step is similar to Step I in the proof of Theorem 1 in \citet{khemakhem2020variational}. We transform the equality of the marginal distributions over observed data into the equality of noise-free distributions. For all pairs $(\boldsymbol{O},\boldsymbol{U})\in\mathcal{O}\times\mathcal{U}$, we have

\makebox[\textwidth][c]{\parbox{1.3\textwidth}{%
{\small
\begin{align}
    p_{\boldsymbol{\theta}}(\boldsymbol{O}|\boldsymbol{U})
    &=p_{\Tilde{\boldsymbol{\theta}}}(\boldsymbol{O}|\boldsymbol{U})\\
    \implies \int_{\mathcal{X}} p_{\boldsymbol{f}}(\boldsymbol{O}|\boldsymbol{X})p_{\boldsymbol{T},\boldsymbol{\lambda}}(\boldsymbol{X}|\boldsymbol{U}) d\boldsymbol{X}
    &= \int_{\mathcal{X}} p_{\Tilde{\boldsymbol{f}}}(\boldsymbol{O}|\boldsymbol{X})p_{\Tilde{\boldsymbol{T}},\Tilde{\boldsymbol{\lambda}}}(\boldsymbol{X}|\boldsymbol{U}) d\boldsymbol{X}\\
    \implies \int_{\mathcal{X}} p_{\boldsymbol{\varepsilon}}(\boldsymbol{O}-\boldsymbol{f}(\boldsymbol{X}))p_{\boldsymbol{T},\boldsymbol{\lambda}}(\boldsymbol{X}|\boldsymbol{U}) d\boldsymbol{X}
    &= \int_{\mathcal{X}} p_{\boldsymbol{\varepsilon}}(\boldsymbol{O}-\Tilde{\boldsymbol{f}}(\boldsymbol{X}))p_{\Tilde{\boldsymbol{T}},\Tilde{\boldsymbol{\lambda}}}(\boldsymbol{X}|\boldsymbol{U}) d\boldsymbol{X}\\
    \implies \int_{\mathcal{O}} p_{\boldsymbol{\varepsilon}}(\boldsymbol{O}-\Bar{\boldsymbol{O}})p_{\boldsymbol{T},\boldsymbol{\lambda}}(\boldsymbol{f}^{-1}(\Bar{\boldsymbol{O}})|\boldsymbol{U})\vol(J_{\boldsymbol{f}^{-1}}(\Bar{\boldsymbol{O}})) d\Bar{\boldsymbol{O}}
    &= \int_{\mathcal{O}} p_{\boldsymbol{\varepsilon}}(\boldsymbol{O}-\Bar{\boldsymbol{O}})p_{\Tilde{\boldsymbol{T}},\Tilde{\boldsymbol{\lambda}}}(\Tilde{\boldsymbol{f}}^{-1}(\Bar{\boldsymbol{O}})|\boldsymbol{U})\vol(J_{\Tilde{\boldsymbol{f}}^{-1}}(\Bar{\boldsymbol{O}})) d\Bar{\boldsymbol{O}} \label{eq:appx_14}\\
    \implies
    \int_{\mathbb{R}^d} p_{\boldsymbol{\varepsilon}}(\boldsymbol{O}-\Bar{\boldsymbol{O}})\Tilde{p}_{\boldsymbol{f}, \boldsymbol{T}, \boldsymbol{\lambda}, \boldsymbol{U}}(\Bar{\boldsymbol{O}}) d\Bar{\boldsymbol{O}}
    &= \int_{\mathbb{R}^d} p_{\boldsymbol{\varepsilon}}(\boldsymbol{O}-\Bar{\boldsymbol{O}})\Tilde{p}_{\Tilde{\boldsymbol{f}}, \Tilde{\boldsymbol{T}}, \Tilde{\boldsymbol{\lambda}}, \Tilde{\boldsymbol{U}}}(\Bar{\boldsymbol{O}}) d\Bar{\boldsymbol{O}}\label{eq:appx_15}\\
    \implies (\Tilde{p}_{\boldsymbol{f}, \boldsymbol{T}, \boldsymbol{\lambda}, \boldsymbol{U}}*~p_{\boldsymbol{\varepsilon}})(\boldsymbol{O})&=(\Tilde{p}_{\Tilde{\boldsymbol{f}}, \Tilde{\boldsymbol{T}}, \Tilde{\boldsymbol{\lambda}}, \Tilde{\boldsymbol{U}}}*~p_{\boldsymbol{\varepsilon}})(\boldsymbol{O})\label{eq:appx_16}\\
    \implies F[\Tilde{p}_{\boldsymbol{f}, \boldsymbol{T}, \boldsymbol{\lambda}, \boldsymbol{U}}](\boldsymbol{\omega})\varphi_{\boldsymbol{\varepsilon}}(\boldsymbol{\omega})&=F[\Tilde{p}_{\Tilde{\boldsymbol{f}}, \Tilde{\boldsymbol{T}}, \Tilde{\boldsymbol{\lambda}}, \Tilde{\boldsymbol{U}}}](\boldsymbol{\omega})\varphi_{\boldsymbol{\varepsilon}}(\boldsymbol{\omega})\label{eq:appx_17}\\
    \implies F[\Tilde{p}_{\boldsymbol{f}, \boldsymbol{T}, \boldsymbol{\lambda}, \boldsymbol{U}}](\boldsymbol{\omega})&=F[\Tilde{p}_{\Tilde{\boldsymbol{f}}, \Tilde{\boldsymbol{T}}, \Tilde{\boldsymbol{\lambda}}, \Tilde{\boldsymbol{U}}}](\boldsymbol{\omega})\label{eq:appx_18}\\
    \implies \Tilde{p}_{\boldsymbol{f}, \boldsymbol{T}, \boldsymbol{\lambda}, \boldsymbol{U}}(\boldsymbol{O})&=\Tilde{p}_{\Tilde{\boldsymbol{f}}, \Tilde{\boldsymbol{T}}, \Tilde{\boldsymbol{\lambda}}, \Tilde{\boldsymbol{U}}}(\boldsymbol{O})\label{eq:appx_19}
\end{align}}
}}
where 
\begin{itemize}
    \item in Eq. (\ref{eq:appx_14}), $J$ denotes the Jacobian, and we made the change of variable $\Bar{\boldsymbol{O}}=\boldsymbol{f}(\boldsymbol{X})$ on the left hand side, and $\Bar{\boldsymbol{O}}=\Tilde{\boldsymbol{f}}(\boldsymbol{X})$ on the right hand side.
    \item in Eq. (\ref{eq:appx_15}), we introduced
    \begin{align}
        \Tilde{p}_{\boldsymbol{f}, \boldsymbol{T}, \boldsymbol{\lambda}, \boldsymbol{U}}(\boldsymbol{O})\triangleq p_{\boldsymbol{T},\boldsymbol{\lambda}}(\boldsymbol{f}^{-1}(\boldsymbol{O})|\boldsymbol{U})\vol(J_{\boldsymbol{f}^{-1}}(\boldsymbol{O}))\mathbb{I}_{\mathcal{O}}(\boldsymbol{O}) \label{eq:appx_21}
    \end{align}
    on the left hand side, and similarly on the right hand side.
    \item in Eq. (\ref{eq:appx_16}), we used $*$ for the convolution operator.
    \item in Eq. (\ref{eq:appx_17}), we used $F[.]$ to designate the Fourier transform, and where $\varphi_{\boldsymbol{\varepsilon}}=F[p_{\boldsymbol{\varepsilon}}]$ (by definition of the characteristic function).
    \item in Eq. (\ref{eq:appx_18}), we dropped $\varphi_{\boldsymbol{\varepsilon}}(\boldsymbol{\omega})$ from both sides as it is non-zero almost everywhere (by assumption (i)).
\end{itemize}
    
\paragraph{Step II.} In this step, we remove all terms that are either a function of $\boldsymbol{O}$ or $\boldsymbol{U}$. First, by replacing both sides of Eq. (\ref{eq:appx_19}) by their corresponding expressions from Eq. (\ref{eq:appx_21}), we have
\begin{align}
p_{\boldsymbol{T},\boldsymbol{\lambda}}(\boldsymbol{f}^{-1}(\boldsymbol{O})|\boldsymbol{U})\vol(J_{\boldsymbol{f}^{-1}}(\boldsymbol{O}))&=p_{\Tilde{\boldsymbol{T}},\Tilde{\boldsymbol{\lambda}}}(\Tilde{\boldsymbol{f}}^{-1}(\boldsymbol{O})|\boldsymbol{U})\vol(J_{\Tilde{\boldsymbol{f}}^{-1}}(\boldsymbol{O})).\label{eq:Thm1stepI}
\end{align}

Then, by taking logarithm on both sides of Eq. (\ref{eq:Thm1stepI}) and replacing $p_{\boldsymbol{T}, \boldsymbol{\lambda}}$ by its expression from Eq. (\ref{eq:thm4_prior}), we obtain
\begin{multline}
    \log\vol(J_{\boldsymbol{f}^{-1}}(\boldsymbol{O}))+\log Q(\boldsymbol{f}^{-1}(\boldsymbol{O}))-\log Z(\boldsymbol{U})+\left<\boldsymbol{T}(\boldsymbol{f}^{-1}(\boldsymbol{O})),\boldsymbol{\lambda}(\boldsymbol{U})\right>\\
    =\log \vol(J_{\Tilde{\boldsymbol{f}}^{-1}}(\boldsymbol{O})) + \log \Tilde{Q}(\Tilde{\boldsymbol{f}}^{-1}(\boldsymbol{O}))-\log\Tilde{Z}(\boldsymbol{U})+\left<\Tilde{\boldsymbol{T}}(\Tilde{\boldsymbol{f}}^{-1}(\boldsymbol{O})),\Tilde{\boldsymbol{\lambda}}(\boldsymbol{U})\right>.
\end{multline}
Let $\boldsymbol{U}^0,\boldsymbol{U}^1,\cdots,\boldsymbol{U}^k\in\mathcal{U}$ be the $k+1$ points defined in assumption (iv). We evaluate the above equation at these points to obtain $k+1$ equations, and subtract the first equation from the remaining $k$ equations to obtain:
\begin{multline}
    \left<\boldsymbol{T}(\boldsymbol{f}^{-1}(\boldsymbol{O})),\boldsymbol{\lambda}(\boldsymbol{U}^l)-\boldsymbol{\lambda}(\boldsymbol{U}^0)\right>+\log\frac{Z(\boldsymbol{U}^0)}{Z(\boldsymbol{U}^l)} \\
    = \left<\Tilde{\boldsymbol{T}}(\Tilde{\boldsymbol{f}}^{-1}(\boldsymbol{O})),\Tilde{\boldsymbol{\lambda}}(\boldsymbol{U}^l)-\Tilde{\boldsymbol{\lambda}}(\boldsymbol{U}^0)\right>+\log\frac{\Tilde{Z}(\boldsymbol{U}^0)}{\Tilde{Z}(\boldsymbol{U}^l)},\quad l=1,\cdots,k.
\end{multline}
Let $L$ be defined as in assumption (iv) and $\Tilde{L}$ defined similarly for $\Tilde{\boldsymbol{\lambda}}$. Note that $L$ is invertible by assumption, but $\Tilde{L}$ is not necessarily invertible. Letting $\boldsymbol{b}\in\mathbb{R}^k$ in which $b_l=\log\frac{\Tilde{Z}(\boldsymbol{U}^0)Z(\boldsymbol{U}^l)}{\Tilde{Z}(\boldsymbol{U}^l)Z(\boldsymbol{U}^0)}$, we have
\begin{align}
    L^T\boldsymbol{T}(\boldsymbol{f}^{-1}(\boldsymbol{O}))=\Tilde{L}^T\Tilde{\boldsymbol{T}}(\Tilde{\boldsymbol{f}}^{-1}(\boldsymbol{O}))+\boldsymbol{b}.
\end{align}
Left multiplying both sides of the above equation by $L^{-T}$ gives
\begin{align}
    \boldsymbol{T}(\boldsymbol{f}^{-1}(\boldsymbol{O}))=A\Tilde{\boldsymbol{T}}(\Tilde{\boldsymbol{f}}^{-1}(\boldsymbol{O}))+\boldsymbol{c},\label{eq:Thm1stepII}
\end{align}
where $A=L^{-T}\Tilde{L}\in\mathbb{R}^{k\times k}$ and $\boldsymbol{c}=L^{-T}\boldsymbol{b}\in\mathbb{R}^{k}$.

\paragraph{Step III.} To complete the proof, we need to show that $A$ is invertible. Let $\boldsymbol{X}_l\in\mathcal{X},~\boldsymbol{O}_l=\boldsymbol{f}(\boldsymbol{X}_l),~l=0,\cdots,k$. We evaluate Eq. (\ref{eq:Thm1stepII}) at these $k+1$ points to obtain $k+1$ equations and subtract the first equation from the remaining $k$ equations to obtain
\begin{multline}
    \underbrace{[\boldsymbol{T}(\boldsymbol{X}_1)-\boldsymbol{T}(\boldsymbol{X}_0),\cdots,\boldsymbol{T}(\boldsymbol{X}_k)-\boldsymbol{T}(\boldsymbol{X}_0)]}_{\triangleq R\in\mathbb{R}^{k\times k}} \\
    =A\underbrace{[\Tilde{\boldsymbol{T}}(\Tilde{\boldsymbol{f}}^{-1}(\boldsymbol{O}_1))-\Tilde{\boldsymbol{T}}(\Tilde{\boldsymbol{f}}^{-1}(\boldsymbol{O}_0)),\cdots,\Tilde{\boldsymbol{T}}(\Tilde{\boldsymbol{f}}^{-1}(\boldsymbol{O}_k))-\Tilde{\boldsymbol{T}}(\Tilde{\boldsymbol{f}}^{-1}(\boldsymbol{O}_0))]}_{\triangleq\Tilde{R}\in\mathbb{R}^{k\times k}}.
\end{multline}
We need to show that for a given $\boldsymbol{X}_0\in\mathcal{X}$, there exist $k$ points $\boldsymbol{X}_1,\cdots,\boldsymbol{X}_k\in\mathcal{X}$ such that the columns of $R$ are linearly independent. Suppose, for contradiction, that the columns of $R$ would never be linearly independent for any choice of $\boldsymbol{X}_1,\cdots,\boldsymbol{X}_k\in\mathcal{X}$. Then the function $\mathbf{g}(\boldsymbol{X})\triangleq \boldsymbol{T}(\boldsymbol{X})-\boldsymbol{T}(\boldsymbol{X}_0)$ would live in a $k-1$ or lower dimensional subspace, and thus we could find a non-zero vector $\boldsymbol{\lambda}\in\mathbb{R}^k$ orthogonal to that subspace. This would imply that $\left<\boldsymbol{T}(\boldsymbol{X})-\boldsymbol{T}(\boldsymbol{X}_0),\boldsymbol{\lambda}\right>=0$ and thus $\left<\boldsymbol{T}(\boldsymbol{X}),\boldsymbol{\lambda}\right>=\left<\boldsymbol{T}(\boldsymbol{X}_0),\boldsymbol{\lambda}\right>=const,~ \forall \boldsymbol{X}\in\mathcal{X}$, which contradicts the assumption that the prior is strongly exponential (Definition \ref{def:strong_ex}). Therefore, we have shown that there exist $k+1$ points $\boldsymbol{X}_0,\boldsymbol{X}_1,\cdots,\boldsymbol{X}_k\in\mathcal{X}$ such that $R$ is invertible. Since $R=A\Tilde{R}$ and $A$ is not a function of $\boldsymbol{X}$, $A$ must be invertible. This completes the proof.
\end{proof}

\subsubsection{Part II}

\begin{thm} \label{thm:5}
    Suppose that all assumptions in Theorem \ref{thm:4} hold. Let the sufficient statistics $\boldsymbol{T}(\boldsymbol{X})=[ \boldsymbol{T}_{f}(\boldsymbol{X})^\text{T}, \boldsymbol{T}_{NN}(\boldsymbol{X})^\text{T}]^\text{T}$ given by the concatenation of a) the sufficient statistics $\boldsymbol{T}_{f}(\boldsymbol{X})=[\boldsymbol{T}_{1}(X_1)^T,\cdots,\boldsymbol{T}_{n}(X_n)^T]^T$ of a factorized exponential family, where all the $\boldsymbol{T}_{i}(X_i) $ have dimension larger or equal to 2, and b) the output $\boldsymbol{T}_{NN}(\boldsymbol{X})$ of a neural network with ReLU activations. (note that a neural network with ReLU activation has universal approximation power and should be able to capture any dependencies of interest). Let $k'$ be the dimension of $\boldsymbol{T}_{f}$ and thus that $k'\geq 2n$. Assume that
    \begin{enumerate}[(i)]
        \item the sufficient statistics $\boldsymbol{T}_{f}$ have all second-order own derivatives;
        \item the mixing function $\boldsymbol{f}$ has all second-order cross derivatives.
    \end{enumerate}
    Then the parameters $\{\boldsymbol{f},\boldsymbol{T},\boldsymbol{\lambda}\}$ are $\sim_P$ identifiable.
\end{thm}

\begin{proof}
    Let $\boldsymbol{v}=\Tilde{\boldsymbol{f}}^{-1}\circ\boldsymbol{f}:\mathcal{X}\to\mathcal{X}$. Since all assumptions in Theorem \ref{thm:4} hold, we have
    \begin{align}
        \boldsymbol{T}(\boldsymbol{X})=A\Tilde{\boldsymbol{T}}(\boldsymbol{v}(\boldsymbol{X}))+\boldsymbol{c},\label{eq:thm2_main}
    \end{align}
    where $A\in\mathbb{R}^{k\times k}$ is invertible. We want to show that $A$ is a block permutation matrix.
    \paragraph{Step I.} In this step, we show that $\boldsymbol{v}$ is a componentwise function. First we differentiate both sides of Eq. (\ref{eq:thm2_main}) with respect to $X_s$ and $X_t$ ($s\not=t$) to obtain
    \begin{align}
        \frac{\partial \boldsymbol{T}(\boldsymbol{X})}{\partial X_s}
        =& A \sum_{i=1}^n \frac{\partial \Tilde{\boldsymbol{T}}(\boldsymbol{v}(\boldsymbol{X}))}{\partial v_i(\boldsymbol{X})}\cdot\frac{\partial v_i(\boldsymbol{X})}{\partial X_s}\\
        \frac{\partial^2 \boldsymbol{T}(\boldsymbol{X})}{\partial X_s\partial X_t}
        =& A \sum_{i=1}^n\sum_{j=1}^n \frac{\partial^2 \Tilde{\boldsymbol{T}}(\boldsymbol{v}(\boldsymbol{X}))}{\partial v_i(\boldsymbol{X})\partial v_j(\boldsymbol{X})}\cdot\frac{\partial v_j(\boldsymbol{X})}{\partial X_t}\cdot\frac{\partial v_i(\boldsymbol{X})}{\partial X_s} \nonumber \\
        &+A \sum_{i=1}^n \frac{\partial \Tilde{\boldsymbol{T}}(\boldsymbol{v}(\boldsymbol{X}))}{\partial v_i(\boldsymbol{X})}\cdot\frac{\partial^2 v_i(\boldsymbol{X})}{\partial X_s\partial X_t}.
    \end{align}
    By construction, the second-order cross derivatives of $\boldsymbol{T}$ and $\Tilde{\boldsymbol{T}}$ are all zero. Therefore, we have
    \begin{align}
        \mathbf{0}
        &= A \sum_{i=1}^n \frac{\partial^2 \Tilde{\boldsymbol{T}}(\boldsymbol{v}(\boldsymbol{X}))}{\partial v_i(\boldsymbol{X})^2}\cdot\frac{\partial v_i(\boldsymbol{X})}{\partial X_t}\cdot\frac{\partial v_i(\boldsymbol{X})}{\partial X_s}+A \sum_{i=1}^n \frac{\partial \Tilde{\boldsymbol{T}}(\boldsymbol{v}(\boldsymbol{X}))}{\partial v_i(\boldsymbol{X})}\cdot\frac{\partial^2 v_i(\boldsymbol{X})}{\partial X_s\partial X_t}.
    \end{align}
    The above equation can be written in the matrix-vector form:
    \begin{align}
        \mathbf{0}&=A\Tilde{\boldsymbol{T}}''(\boldsymbol{X})\boldsymbol{v}_{s,t}'(\boldsymbol{X})+A\Tilde{\boldsymbol{T}}'(\boldsymbol{X})\boldsymbol{v}_{s,t}''(\boldsymbol{X}),
    \end{align}
    where
    \begin{align}
        \Tilde{\boldsymbol{T}}''(\boldsymbol{X}) = \left[\frac{\partial^2 \Tilde{\boldsymbol{T}}(\boldsymbol{v}(\boldsymbol{X}))}{\partial v_1(\boldsymbol{X})^2},\cdots,\frac{\partial^2 \Tilde{\boldsymbol{T}}(\boldsymbol{v}(\boldsymbol{X}))}{\partial v_n(\boldsymbol{X})^2}\right]\in\mathbb{R}^{k\times n}\\
        \boldsymbol{v}_{s,t}'(\boldsymbol{X}) = \left[ \frac{\partial v_1(\boldsymbol{X})}{\partial X_t}\cdot\frac{\partial v_1(\boldsymbol{X})}{\partial X_s},\cdots,\frac{\partial v_n(\boldsymbol{X})}{\partial X_t}\cdot\frac{\partial v_n(\boldsymbol{X})}{\partial X_s} \right]^T\in\mathbb{R}^{n},
    \end{align}
    and
    \begin{align}
        \Tilde{\boldsymbol{T}}'(\boldsymbol{X}) = \left[\frac{\partial \Tilde{\boldsymbol{T}}(\boldsymbol{v}(\boldsymbol{X}))}{\partial v_1(\boldsymbol{X})},\cdots,\frac{\partial \Tilde{\boldsymbol{T}}(\boldsymbol{v}(\boldsymbol{X}))}{\partial v_n(\boldsymbol{X})}\right]\in\mathbb{R}^{k\times n}\\
        \boldsymbol{v}_{s,t}''(\boldsymbol{X}) = \left[\frac{\partial^2 v_1(\boldsymbol{X})}{\partial X_s\partial X_t},\cdots,\frac{\partial^2 v_n(\boldsymbol{X})}{\partial X_s\partial X_t}\right]^T \in\mathbb{R}^{n}.
    \end{align}
    Now by concatenating
    \begin{align}
        \Tilde{\boldsymbol{T}}'''(\boldsymbol{X})=[\Tilde{\boldsymbol{T}}''(\boldsymbol{X}),\Tilde{\boldsymbol{T}}'(\boldsymbol{X})]\in\mathbb{R}^{k\times 2n}\quad\text{and} \nonumber \\ \boldsymbol{v}_{s,t}''(\boldsymbol{X})=[\boldsymbol{v}_{s,t}'(\boldsymbol{X})^T,\boldsymbol{v}_{s,t}''(\boldsymbol{X})^T]^T\in\mathbb{R}^{2n},
    \end{align}
    we obtain
    \begin{align}
        \mathbf{0}&=A\Tilde{\boldsymbol{T}}'''(\boldsymbol{X})\boldsymbol{v}_{s,t}'''(\boldsymbol{X}).
    \end{align}
    Finally, we take the rows of $\Tilde{\boldsymbol{T}}'''(\boldsymbol{X})$ that corresponds to the factorized strongly exponential family distribution part and denote them by $\Tilde{\boldsymbol{T}}_f'''(\boldsymbol{X})\in\mathbb{R}^{k'\times 2n}$. By Lemma 5 in the iVAE paper \citep{khemakhem2020variational} and the assumption that $k'\geq 2n$, we have that the rank of $\Tilde{\boldsymbol{T}}_f'''(\boldsymbol{X})$ is $2n$. Since $k\geq k'\geq 2n$, the rank of $\Tilde{\boldsymbol{T}}'''(\boldsymbol{X})$ is also $2n$. Since the rank of $A$ is $k$, the rank of $A\Tilde{\boldsymbol{T}}'''(\boldsymbol{X})\in\mathbb{R}^{k\times 2n}$ is $2n$. This implies that $\boldsymbol{v}_{s,t}'''(\boldsymbol{X})$ must be a zero vector. In particular, we have that $\boldsymbol{v}_{s,t}'(\boldsymbol{X})=\mathbf{0},~\forall s\not=t$. Therefore, we have shown that $\boldsymbol{v}$ is a componentwise function.
    \paragraph{Step II.} To complete the proof, we need to show that $A$ is a block permutation matrix. Without loss of generality, we assume that the permutation in $\boldsymbol{v}$ is the identity. That is $\boldsymbol{v}(\boldsymbol{X})=[v_1(X_1),\cdots,v_n(X_n)]^T$ for some nonlinear univariate scalar functions $v_1,\cdots,v_n$. Since $\boldsymbol{f}$ and $\Tilde{\boldsymbol{f}}$ are bijective, we have that $\boldsymbol{v}$ is also bijective and $\boldsymbol{v}^{-1}(\boldsymbol{X})=[v_1^{-1}(X_1),\cdots,v_n^{-1}(X_n)]^T$. We denote $\Bar{\boldsymbol{T}}(\boldsymbol{v}(\boldsymbol{X}))=\Tilde{\boldsymbol{T}}(\boldsymbol{v}(\boldsymbol{X}))+A^{-1}\boldsymbol{c}$ and plug it into Eq. (\ref{eq:thm2_main}) to obtain $\boldsymbol{T}(\boldsymbol{X})=A\Bar{\boldsymbol{T}}(\boldsymbol{v}(\boldsymbol{X}))$. Applying $\boldsymbol{v}^{-1}$ to the variables $\boldsymbol{X}$ at both sides gives 
    \begin{align}
        \boldsymbol{T}(\boldsymbol{v}^{-1}(\boldsymbol{X}))=A\Bar{\boldsymbol{T}}(\boldsymbol{X}).
    \end{align}
    Let $t$ be the index of an entry in the sufficient statistics $\boldsymbol{T}$ that corresponds to the the factorized strongly exponential family distribution part $\boldsymbol{T}_{f}$. For all $ s\not=t$, we have
    \begin{align}
        0=\frac{\partial \boldsymbol{T}(\boldsymbol{v}^{-1}(\boldsymbol{X}))_t}{\partial X_s}=\sum_{j=1}^k a_{tj}\frac{\partial\Bar{\boldsymbol{T}}(\boldsymbol{X})_j}{\partial X_s}.
    \end{align}
    Since the entries of $\Tilde{\boldsymbol{T}}$ are linearly independent (if they were not linearly independent, then $\Tilde{\boldsymbol{T}}$ can be compressed into a smaller vector by removing the redundant entries), we have that $a_{tj}$ is zero for any $j$ such that $\frac{\partial\Bar{\boldsymbol{T}}(\boldsymbol{X})_j}{\partial X_s}\not=0$. This includes the entries $j$ in the sufficient statistics $\Tilde{\boldsymbol{T}}$ that corresponds to 1) the factorized strongly exponential family distribution part which do not depend on $X_t$; and 2) the neural network part.\\
    
    \noindent Therefore, when $t$ is the index of an entry in the sufficient statistics $\boldsymbol{T}$ that corresponds to factor $i$ in the factorized strongly exponential family distribution part $\boldsymbol{T}_{f}$, the only non-zero $a_{tj}$ are the ones that map between $\boldsymbol{T}_{i}(X_i)$ and $\Bar{\boldsymbol{T}}_{i}(v_i(X_i))$, where $\boldsymbol{T}_{i}$ are the factors in $\boldsymbol{T}_f$ that only depend on $X_i$ and $\Bar{\boldsymbol{T}}_{i}$ is defined similarly. Therefore, we can construct an invertible submatrix $A_i'$ with all non-zero elements $a_{tj}$ for all $t$ that corresponds to factor $i$, such that
    \begin{align}
        \boldsymbol{T}_{i}(X_i)=A_i'\Bar{\boldsymbol{T}}_{i}(v_i(X_i))=A_i'\Tilde{\boldsymbol{T}}_{i}(v_i(X_i))+\boldsymbol{c}_i,\quad i=1,\cdots,n,
    \end{align}
    where $\Tilde{\boldsymbol{T}}_{i}$ are the factors in $\Tilde{\boldsymbol{T}}_f$ that only depends on $X_i$, and $\boldsymbol{c}_i$ are the corresponding elements of $\boldsymbol{c}$. This means that the matrix $A$ is a block permutation matrix. For each $i=1,\cdots,n$, the block $A_i'$ of $A$ affinely transforms $\boldsymbol{T}_{i}(X_i)$ into $\Tilde{\boldsymbol{T}}_{i}(v_i(X_i))$. There is also an additional block $A'$ which affinely transforms $\boldsymbol{T}_{NN}(\boldsymbol{X})$ into $\Tilde{\boldsymbol{T}}_{NN}(\boldsymbol{v}(\boldsymbol{X}))$. This completes the proof.
\end{proof}

\subsubsection{Part III}

Now we combine Theorem \ref{thm:4} in Part I and Theorem \ref{thm:5} in Part II into one theorem, which completes the proof of Theorem \ref{thm:1}.


\subsection{Proof of Theorem \ref{thm:2}}

We recall that the loss function in Phase 1 is as follows:
\begin{align}
\mathcal{L}_{\text{phase1}}(\boldsymbol{\theta}, \boldsymbol{\phi}) = \mathcal{L}^{\text{VAE}}_{\text{phase1}}(\boldsymbol{f}, \hat{\boldsymbol{T}}, \hat{\boldsymbol{\lambda}}, \boldsymbol{\phi}) - \mathcal{L}^{\text{SM}}_{\text{phase1}}(\hat{\boldsymbol{f}}, \boldsymbol{T}, \boldsymbol{\lambda}, \hat{\boldsymbol{\phi}}), \label{eq:appx_phase1_vaesm}
\end{align}
where 
{
\begin{multline}  
\mathcal{L}^{\text{VAE}}_{\text{phase1}}(\boldsymbol{\theta}, \boldsymbol{\phi}) 
:= \mathbb{E}_{p_D}\big[\mathbb{E}_{q_{\boldsymbol{\phi}}(\boldsymbol{X}|\boldsymbol{O}, \boldsymbol{Y}, \boldsymbol{E})}\left[\log p_{\boldsymbol{f}}(\boldsymbol{O}|\boldsymbol{X})\right. \\ + \left.\log p_{\boldsymbol{T}, \boldsymbol{\lambda}}(\boldsymbol{X}|\boldsymbol{Y}, \boldsymbol{E})-\log q_{\boldsymbol{\phi}}(\boldsymbol{X}|\boldsymbol{O}, \boldsymbol{Y}, \boldsymbol{E})\right]\big],  \label{eq:appx_phase1_loss} 
\end{multline}
\begin{multline}  
\mathcal{L}^{\text{SM}}_{\text{phase1}}(\boldsymbol{T}, \boldsymbol{\lambda}) 
:= \mathbb{E}_{p_D}\big[\mathbb{E}_{q_{\boldsymbol{\phi}}(\boldsymbol{X}|\boldsymbol{O}, \boldsymbol{Y}, \boldsymbol{E})}\left[||\nabla_{\boldsymbol{X}} \log q_{\boldsymbol{\phi}}(\boldsymbol{X}|\boldsymbol{O}, \boldsymbol{Y}, \boldsymbol{E})\right. \\ \left.- \nabla_{\boldsymbol{X}} \log p_{\boldsymbol{T}, \boldsymbol{\lambda}}(\boldsymbol{X}|\boldsymbol{Y}, \boldsymbol{E})||^2
\right]\big]. \label{eq:appx_phase1_sm}
\end{multline}

\begin{proof}
If the family of $q_{\boldsymbol{\phi}}(\boldsymbol{X}|\boldsymbol{O}, \boldsymbol{Y}, \boldsymbol{E})$ is flexible enough so that it contains $p_{\boldsymbol{\theta}}(\boldsymbol{X}|\boldsymbol{O}, \boldsymbol{Y}, \boldsymbol{E})$, then by optimizing the loss over its parameter $\boldsymbol{\phi}$, we will minimize the score matching term $\mathcal{L}^{\text{SM}}_{\text{phase1}}$ in Eq. (\ref{eq:appx_phase1_sm}), which will eventually reach zero. If we assume that the model is not degenerate and that $q_{\boldsymbol{\phi}}>0$ everywhere, then having that $\mathcal{L}^{\text{SM}}_{\text{phase1}} =0$ implies that $\nabla_{\boldsymbol{X}} \log q_{\boldsymbol{\phi}}(\boldsymbol{X}|\boldsymbol{O}, \boldsymbol{Y}, \boldsymbol{E})$ and $\nabla_{\boldsymbol{X}} \log p_{\boldsymbol{T}, \boldsymbol{\lambda}}(\boldsymbol{X}|\boldsymbol{Y}, \boldsymbol{E})$ are equal. This implies that $\log q_{\boldsymbol{\phi}}(\boldsymbol{X}|\boldsymbol{O}, \boldsymbol{Y}, \boldsymbol{E})=\log p_{\boldsymbol{T}, \boldsymbol{\lambda}}(\boldsymbol{X}|\boldsymbol{Y}, \boldsymbol{E})+c$ for some constant $c$. But $c$ is necessarily $0$ because both $q_{\boldsymbol{\phi}}(\boldsymbol{X}|\boldsymbol{O}, \boldsymbol{Y}, \boldsymbol{E})$ and $p_{\boldsymbol{T}, \boldsymbol{\lambda}}(\boldsymbol{X}|\boldsymbol{Y}, \boldsymbol{E})$ are pdf's. Therefore, the VAE term $\mathcal{L}^{\text{VAE}}_{\text{phase1}}$ in Eq. (\ref{eq:appx_phase1_loss}) will be equal to the log-likelihood, meaning that the loss $\mathcal{L}_{\text{phase1}}$ in Eq. (\ref{eq:appx_phase1_vaesm}) will be equal to the log-likelihood. Under this circumstance, the estimation in Eq. (\ref{eq:appx_phase1_vaesm}) inherits all the properties of maximum likelihood estimation (MLE). In this particular case, since our identifiability is guaranteed up to a permutation and componentwise transformation, the consistency of MLE means that we converge to the true parameter $\boldsymbol{\theta}^{\ast}$ up to a permutation and componentwise transformation in the limit of infinite data. Because true identifiability is one of the assumptions for MLE consistency, replacing it by identifiability up to a permutation and componentwise transformation does not change the proof but only the conclusion.
\end{proof}


\subsection{Proof of Theorem \ref{thm:3}}

\begin{proof}
Theorem \ref{thm:1} and Theorem \ref{thm:2} guarantee that in the limit of infinite data, iVAE with a general non-factorized prior can learn the true parameters $\boldsymbol{\theta}^{\ast}:=(\boldsymbol{f}^{\ast}, \boldsymbol{T}^{\ast}, \boldsymbol{\lambda}^{\ast})$ up to a permutation and componentwise transformation of the latent variables. Let $(\hat{\boldsymbol{f}}, \hat{\boldsymbol{T}}, \hat{\boldsymbol{\lambda}})$ be the parameters obtained by iVAE. We, therefore, have $(\hat{\boldsymbol{f}}, \hat{\boldsymbol{T}}, \hat{\boldsymbol{\lambda}}) \sim_P (\boldsymbol{f}^{\ast}, \boldsymbol{T}^{\ast}, \boldsymbol{\lambda}^{\ast})$, where $\sim_P$ denotes the equivalence up to a permutation and componentwise transformation. If there were no noise, this would mean that the learned $\hat{\boldsymbol{f}}$ transforms $\boldsymbol{O}$ into $\hat{\boldsymbol{X}}= \hat{\boldsymbol{f}}^{-1}(\boldsymbol{O})$ that are equal to $\boldsymbol{X}^{\ast}= \left(\boldsymbol{f}^{\ast}\right)^{-1}(\boldsymbol{O})$ up to a permutation and componentwise transformation (Definition \ref{dfn:P_idf}). If with noise, we obtain the posterior distribution of the latent variables up to an analogous indeterminacy.
\end{proof}


\subsection{Proof of Proposition \ref{prop:1}} \label{appendix:proof4}

\begin{proof} The following rules can be independently performed to distinguish all the 10 structures shown in Figs. \ref{fig:2c}-\ref{fig:2m}. For clarity, we divide them into three groups. Note that, since these rules rely on different algorithms of causal discovery and conditional independence tests, unless stated otherwise, we assume that the assumptions of these algorithms are satisfied during the proof process.

\paragraph{Group 1} All the six structures in this group can be discovered only by performing conditional independence tests. 

\begin{itemize}
\item \textbf{Rule 1.1 } If $X_i \indep \boldsymbol{Y}$, $X_i \notindep \boldsymbol{E}$, and $\boldsymbol{E} \indep \boldsymbol{Y}$, then Fig. \ref{fig:2e} is discovered.
\item \textbf{Rule 1.2 } If $X_i \notindep \boldsymbol{Y}$, $X_i \indep \boldsymbol{E}$, and $\boldsymbol{E} \notindep \boldsymbol{Y}$, then Fig. \ref{fig:2i} is discovered.
\item \textbf{Rule 1.3 } If $X_i \notindep \boldsymbol{Y}$, $X_i \notindep \boldsymbol{E}$, and $\boldsymbol{E} \indep \boldsymbol{Y}$, then Fig. \ref{fig:2g} is discovered.
\item \textbf{Rule 1.4 } If $X_i \notindep \boldsymbol{Y}$, $X_i \notindep \boldsymbol{E}$, $\boldsymbol{E} \notindep \boldsymbol{Y}$, and $X_i \indep \boldsymbol{Y}|\boldsymbol{E}$, then Fig. \ref{fig:2k} is discovered.
\item \textbf{Rule 1.5 } If $X_i \notindep \boldsymbol{Y}$, $X_i \notindep \boldsymbol{E}$, $\boldsymbol{E} \notindep \boldsymbol{Y}$, and $X_i \indep \boldsymbol{E} | \boldsymbol{Y}$, then Fig. \ref{fig:2j} is discovered.
\item \textbf{Rule 1.6 } If $X_i \notindep \boldsymbol{Y}$, $X_i \notindep \boldsymbol{E}$, $\boldsymbol{E} \notindep \boldsymbol{Y}$, and $\boldsymbol{Y} \indep \boldsymbol{E} | X_i$, then Fig. \ref{fig:2f} is discovered.
\end{itemize}

\paragraph{Group 2} If $X_i \notindep \boldsymbol{Y}$, $X_i \indep \boldsymbol{E}$, and $\boldsymbol{E} \indep \boldsymbol{Y}$, then we can discover both Fig. \ref{fig:2c} and Fig. \ref{fig:2d}. These two structures cannot be further distinguished only by conditional independence tests, because they come from the same Markov equivalence class. Fortunately, we can further distinguish them by running binary causal discovery algorithms \citep{peters2017elements}, e.g., ANM \citep{hoyer2009nonlinear} for continuous data and CDS \citep{fonollosa2019conditional} for continuous and discrete data.

\begin{itemize}
\item \textbf{Rule 2.1 } If $X_i \notindep \boldsymbol{Y}$, $X_i \indep \boldsymbol{E}$, and $\boldsymbol{E} \indep \boldsymbol{Y}$, and a chosen binary causal discovery algorithm prefers $X_i \rightarrow \boldsymbol{Y}$ to $X_i \leftarrow \boldsymbol{Y}$, then Fig. \ref{fig:2c} is discovered.
\item \textbf{Rule 2.2 } If $X_i \notindep \boldsymbol{Y}$, $X_i \indep \boldsymbol{E}$, and $\boldsymbol{E} \indep \boldsymbol{Y}$, and a chosen binary causal discovery algorithm prefers $X_i \leftarrow \boldsymbol{Y}$ to $X_i \rightarrow \boldsymbol{Y}$, then Fig. \ref{fig:2d} is discovered.
\end{itemize}

\paragraph{Group 3} If $X_i \notindep \boldsymbol{Y}$, $X_i \notindep \boldsymbol{E}$, $\boldsymbol{E} \notindep \boldsymbol{Y}$, $X_i \notindep \boldsymbol{Y}|\boldsymbol{E}$, $X_i \notindep \boldsymbol{E} | \boldsymbol{Y}$, and $\boldsymbol{Y} \notindep \boldsymbol{E} | X_i$, then we can discover both Fig. \ref{fig:2l} and Fig. \ref{fig:2m}. These two structures cannot be further distinguished only by conditional independence tests, because they come from the same Markov equivalence class. They also cannot be distinguished by any binary causal discovery algorithm, since both $X_i$ and $\boldsymbol{Y}$ are affected by $\boldsymbol{E}$. Fortunately, \citet{zhang2017causal} provided a heuristic solution to this case based on the invariance of causal mechanisms, i.e., $P(\text{cause})$ and $P(\text{effect}|\text{cause})$ change independently. The detailed description of their method is given in Section 4.2 of \citet{zhang2017causal}. For convenience, here we directly borrow their final result. \citet{zhang2017causal} states that determining the causal direction between $X_i$ and $\boldsymbol{Y}$ in Fig. \ref{fig:2l} and Fig. \ref{fig:2m} is finally reduced to calculating the following term:
\begin{align}
\Delta_{X_i \rightarrow \boldsymbol{Y}} = 
\left\langle
\log\frac{\bar{P}(\boldsymbol{Y}|X_i)}{\langle\hat{P}(\boldsymbol{Y}|X_i)\rangle}
\right\rangle,
\end{align}
where $\langle \cdot \rangle$ denotes the sample average, $\bar{P}(\boldsymbol{Y}|X_i)$ is the empirical estimate of $P(\boldsymbol{Y}|X_i)$ on all data points, and $\langle\hat{P}(\boldsymbol{Y}|X_i)\rangle$ denotes the sample average of $\hat{P}(\boldsymbol{Y}|X_i)$, which is the estimate of $P(\boldsymbol{Y}|X_i)$ in each environment. We take the direction for which $\Delta$ is smaller to be the causal direction.

\begin{itemize}
\item \textbf{Rule 3.1 } If $X_i \notindep \boldsymbol{Y}$, $X_i \notindep \boldsymbol{E}$, $\boldsymbol{E} \notindep \boldsymbol{Y}$, $X_i \notindep \boldsymbol{Y}|\boldsymbol{E}$, $X_i \notindep \boldsymbol{E} | \boldsymbol{Y}$, $\boldsymbol{Y} \notindep \boldsymbol{E} | X_i$, and $\Delta_{X_i \rightarrow \boldsymbol{Y}}$ is smaller than $\Delta_{\boldsymbol{Y} \rightarrow X_i}$, then Fig. \ref{fig:2l} is discovered.
\item \textbf{Rule 3.2 } If $X_i \notindep \boldsymbol{Y}$, $X_i \notindep \boldsymbol{E}$, $\boldsymbol{E} \notindep \boldsymbol{Y}$, $X_i \notindep \boldsymbol{Y}|\boldsymbol{E}$, $X_i \notindep \boldsymbol{E} | \boldsymbol{Y}$, $\boldsymbol{Y} \notindep \boldsymbol{E} | X_i$, and $\Delta_{\boldsymbol{Y} \rightarrow X_i}$ is smaller than $\Delta_{X_i \rightarrow \boldsymbol{Y}}$, then Fig. \ref{fig:2m} is discovered.
\end{itemize}

\end{proof}


\subsection{Proof of Proposition \ref{prop:2}}

\begin{proof}
Theorem A.1 in \citep{DBLP:journals/corr/abs-2103-02667} has showed that i) any predictor $w \circ \Phi$ with optimal OOD generalization uses only
$\text{\normalfont Pa}(\boldsymbol{Y})$ to compute $\Phi$; ii) the classifier $w$ in this optimal predictor can be estimated using data from any environment $e$ for which the distribution of $\text{\normalfont Pa}(\boldsymbol{Y}^e)$ has full support; iii) the optimal predictor will be invariant across $\mathcal{E}_{all}$. In iCaRL, the hypotheses of Theorems \ref{thm:1} and \ref{thm:2} and Assumption \ref{eq:assumption1} guarantee that $\text{\normalfont Pa}(\boldsymbol{Y})$ can be recovered by first identifying the latent variables $\boldsymbol{X}$ from $\boldsymbol{O}$, $\boldsymbol{Y}$ and $\boldsymbol{E}$ and then discovering the direct causes of $\boldsymbol{Y}$ through solving Eq. (\ref{eq:phase3_phi}). Furthermore, since the conditional prior in Eq. (\ref{eq:asm4_prior}) of the main text has full support, the distribution of $\text{\normalfont Pa}(\boldsymbol{Y}^e)$ always has full support. Also, under Assumption \ref{eq:assumption1} we have that $p(\boldsymbol{Y}|\text{\normalfont Pa}(\boldsymbol{Y}))$ is invariant across $\mathcal{E}_{all}$. Hence, the classifier $w$ in this optimal predictor can be estimated using data from any environment $e$. We therefore have that the resulting optimal predictor will be invariant across $\mathcal{E}_{all}$. This completes the proof.
\end{proof}


\section{Datasets} \label{appendix:datasets}

For convenience and completeness, we provide descriptions of Colored MNIST Digits and Colored Fashion MNIST here. Please refer to the original papers \citep{arjovsky2019invariant,ahuja2020invariant,gulrajani2020search,venkateswara2017deep} for more details. 


\subsection{Synthetic Data}

For the nonlinear transformation, we use the MLP:
\begin{itemize}
\item Input layer: Input batch \textit{(batch size, input dimension)}
\item Layer 1: Fully connected layer, output size = 6, activation = ReLU
\item Output layer: Fully connected layer, output size = 10
\end{itemize}


\subsection{Colored MNIST Digits}

We use the exact same environment as in \citet{arjovsky2019invariant}. \citet{arjovsky2019invariant} propose to create an environment for training to classify digits in MNIST  data\footnote{\url{https://www.tensorflow.org/api_docs/ python/tf/keras/datasets/mnist/load_data}}, where the images in MNIST are now colored in such a way that the colors spuriously correlate with the labels. The task is to classify whether the digit is less than 5 (not including 5) or more than 5. There are three environments (two training containing 30,000 points each, one test containing 10,000 points) We add noise to the preliminary label ($\tilde{y} = 0$ if the digit is between 0-4 and $\tilde{y} = 1$ if the digit is between 5-9) by flipping it with 25 percent probability to construct the final labels. We sample the color id $z$ by flipping the final labels with probability $p_e$, where $p_e$ is $0.2$ in the first environment, $0.1$ in the second environment, and $0.9$ in the third environment. The third environment is the testing environment. We color the digit red if $z = 1$ or green if $z = 0$.


\subsection{Colored Fashion MNIST}

We modify the fashion MNIST dataset\footnote{\url{https://www.tensorflow.org/api_docs/ python/tf/keras/datasets/fashion_mnist/
load_data}} in a manner similar to the MNIST digits dataset. Fashion MNIST data has images from different categories: ``t-shirt'', ``trouser'', ``pullover'', ``dress'', ``coat'', ``sandal'', ``shirt'', ``sneaker'', ``bag'', ``ankle boots''. We add colors to the images in such a way that the colors correlate with the labels. The task is to classify whether the image is that of foot wear or a clothing item. There are three environments (two training, one test) We add noise to the preliminary label ($\tilde{y} = 0$: ``t-shirt'', ``trouser'', ``pullover'', ``dress'', ``coat'', ``shirt'' and $\tilde{y} = 1$: ``sandal'', ``sneaker'', ``ankle boots'') by flipping it with 25 percent probability to construct the final label. We sample the color id $z$ by flipping the noisy label with probability $p_e$, where $p_e$ is $0.2$ in the first environment, $0.1$ in the second environment, and $0.9$ in the third environment, which is the test environment. We color the object red if $z = 1$ or green if $z = 0$.


\section{Implementation Details}
\label{appendix:implementation}

\subsection{Joint Training}

As described in Section \ref{sect:phase1}, we can jointly learn $(\boldsymbol{\theta}, \boldsymbol{\phi})$ by optimizing the following objective:
{\small
\begin{align}
&\mathcal{L}_{\text{phase1}}(\boldsymbol{\theta}, \boldsymbol{\phi}) \\
=& \mathcal{L}^{\text{VAE}}_{\text{phase1}}(\boldsymbol{f}, \hat{\boldsymbol{T}}, \hat{\boldsymbol{\lambda}}, \boldsymbol{\phi}) - \mathcal{L}^{\text{SM}}_{\text{phase1}}(\hat{\boldsymbol{f}}, \boldsymbol{T}, \boldsymbol{\lambda}, \hat{\boldsymbol{\phi}}) \\
=& \mathbb{E}_{p_D}\left[\mathbb{E}_{q_{\boldsymbol{\phi}}(\boldsymbol{X}|\boldsymbol{O}, \boldsymbol{Y}, \boldsymbol{E})}\left[\log p_{\boldsymbol{f}}(\boldsymbol{O}|\boldsymbol{X}) + \log p_{\hat{\boldsymbol{T}}, \hat{\boldsymbol{\lambda}}}(\boldsymbol{X}|\boldsymbol{Y}, \boldsymbol{E})-\log q_{\boldsymbol{\phi}}(\boldsymbol{X}|\boldsymbol{O}, \boldsymbol{Y}, \boldsymbol{E})\right]\right] \nonumber \\
&- 
\mathbb{E}_{p_D}\left[\mathbb{E}_{q_{\hat{\boldsymbol{\phi}}}(\boldsymbol{X}|\boldsymbol{O}, \boldsymbol{Y}, \boldsymbol{E})}\left[||\nabla_{\boldsymbol{X}} \log q_{\hat{\boldsymbol{\phi}}}(\boldsymbol{X}|\boldsymbol{O}, \boldsymbol{Y}, \boldsymbol{E}) 
- \nabla_{\boldsymbol{X}} \log p_{\boldsymbol{T}, \boldsymbol{\lambda}}(\boldsymbol{X}|\boldsymbol{Y}, \boldsymbol{E})||^2
\right]\right] \\
=& \mathbb{E}_{p_D}\left[\mathbb{E}_{q_{\boldsymbol{\phi}}(\boldsymbol{X}|\boldsymbol{O}, \boldsymbol{Y}, \boldsymbol{E})}\left[\log p_{\boldsymbol{f}}(\boldsymbol{O}|\boldsymbol{X}) + \log p_{\hat{\boldsymbol{T}}, \hat{\boldsymbol{\lambda}}}(\boldsymbol{X}|\boldsymbol{Y}, \boldsymbol{E})-\log q_{\boldsymbol{\phi}}(\boldsymbol{X}|\boldsymbol{O}, \boldsymbol{Y}, \boldsymbol{E})\right]\right] \nonumber \\
&-
\mathbb{E}_{p_D}\left[\mathbb{E}_{q_{\hat{\boldsymbol{\phi}}}(\boldsymbol{X}|\boldsymbol{O}, \boldsymbol{Y}, \boldsymbol{E})}\left[
\sum_{j=1}^n \left[
\frac{\partial^2 p_{\boldsymbol{T}, \boldsymbol{\lambda}}(\boldsymbol{X}| \boldsymbol{Y}, \boldsymbol{E})}{\partial X_j^2} 
+ \frac{1}{2}\left(
\frac{\partial p_{\boldsymbol{T}, \boldsymbol{\lambda}}(\boldsymbol{X}| \boldsymbol{Y}, \boldsymbol{E})}{\partial X_j}\right)^2
\right]
\right]\right] \nonumber \\
& + const.
\end{align}
}
where the last equality is due to the equation in Appendix \ref{appendix:dev_sm}, and $\hat{\boldsymbol{f}}, \hat{\boldsymbol{T}}, \hat{\boldsymbol{\lambda}}, \hat{\boldsymbol{\phi}}$ are
copies of $\boldsymbol{f}, \boldsymbol{T}, \boldsymbol{\lambda}, \boldsymbol{\phi}$
that are treated as constants and whose
gradient is not calculated during learning. In practice, $\hat{\boldsymbol{f}}, \hat{\boldsymbol{T}}, \hat{\boldsymbol{\lambda}}, \hat{\boldsymbol{\phi}}$ can be easily implemented through either ``\texttt{detach}'' in PyTorch \citep{NEURIPS2019_9015} or ``\texttt{stop\_gradient}'' in TensorFlow \citep{tensorflow2015-whitepaper}.


\subsection{The General Non-factorized Prior}

In the experiments, the general non-factorized prior in Assumption \ref{eq:assumption2} is implemented as follows:
{
\begin{multline}
p_{\boldsymbol{T},\boldsymbol{\lambda}}(\boldsymbol{X}| \boldsymbol{Y}, \boldsymbol{E})
= \big\langle \underbrace{\texttt{NN}(\boldsymbol{X}; \texttt{param1})}_{\boldsymbol{T}_{NN}(\boldsymbol{X})},  \underbrace{\texttt{NN}(\boldsymbol{Y}, \boldsymbol{E};\texttt{param2})}_{\boldsymbol{\lambda}_{NN}(\boldsymbol{Y}, \boldsymbol{E})}\big\rangle \\
+ \big\langle
\underbrace{\texttt{concat}(\boldsymbol{X}, \boldsymbol{X}^2)}_{\boldsymbol{T}_{f}(\boldsymbol{X})}, \underbrace{\texttt{NN}(\boldsymbol{Y}, \boldsymbol{E};\texttt{param3})}_{\boldsymbol{\lambda}_{f}(\boldsymbol{Y}, \boldsymbol{E})}
\big\rangle, \nonumber
\end{multline}
}
where $\langle \cdot, \cdot \rangle$ is the dot product of two vectors, and $\texttt{concat}(\cdot, \cdot)$ means the concatenation of two vectors. Now let us explain each term in details. 

Firstly, $\texttt{concat}(\boldsymbol{X}, \boldsymbol{X}^2)$ is a vector of the latent variables and their squared values, and $\texttt{NN}(\boldsymbol{Y}, \boldsymbol{E};\texttt{param3})$ is a deep neural network parameterized by $\texttt{param3}$ that computes a vector of natural parameters as a function of $\boldsymbol{Y}$ and $\boldsymbol{E}$. Hence, the term $\big\langle\texttt{concat}(\boldsymbol{X}, \boldsymbol{X}^2), \texttt{NN}(\boldsymbol{Y}, \boldsymbol{E};\texttt{param3})
\big\rangle$ is equivalent to the factorized exponential family, which also satisfies that each $\boldsymbol{T}_i(X_i)$ has dimension larger or equal to 2.

Secondly, $\texttt{NN}(\boldsymbol{X}; \texttt{param1})$ is a neural network that receives as input a vector of latent variables and outputs another vector representing complicated nonlinear transformations of those variables. $\texttt{NN}(\boldsymbol{Y}, \boldsymbol{E};\texttt{param2})$ is another neural network that generates a corresponding vector of natural parameters. Hence, the term $\big\langle \texttt{NN}(\boldsymbol{X}; \texttt{param1}),  \texttt{NN}(\boldsymbol{Y}, \boldsymbol{E};\texttt{param2})\big\rangle$ will allow this prior to capture the dependencies between the latent variables $\boldsymbol{X}$.


\section{Hyperparameters and Architectures} \label{appendix:architectures}

In this section, we describe the hyperparameters and architectures of different models used in different experiments. 


\subsection{Synthetic Data}

We used Adam optimizer for training with learning rate set to 1e-3 and batch size set to $128$.


\subsubsection{ERM}
\paragraph{Linear ERM}
\begin{itemize}
\item Input layer: Input batch \textit{(batch size, input dimension)}
\item Output layer: Fully connected layer, output size = 1
\end{itemize}

\paragraph{Nonlinear ERM}
\begin{itemize}
\item Input layer: Input batch \textit{(batch size, input dimension)}
\item Layer 1: Fully connected layer, output size = 6, activation = ReLU
\item Output layer: Fully connected layer, output size = 1
\end{itemize}


\subsubsection{IRM}
\paragraph{Linear Data Representation $\Phi$}
\begin{itemize}
\item Input layer: Input batch \textit{(batch size, input dimension)}
\item Output layer: Fully connected layer, output size = 1
\end{itemize}

\paragraph{Nonlinear Data Representation $\Phi$}
\begin{itemize}
\item Input layer: Input batch \textit{(batch size, input dimension)}
\item Layer 1: Fully connected layer, output size = 6, activation = ReLU
\item Output layer: Fully connected layer, output size = 1
\end{itemize}


\subsubsection{F-IRM GAME}
\paragraph{Linear Classifier $w$}
\begin{itemize}
\item Input layer: Input batch \textit{(batch size, input dimension)}
\item Output layer: Fully connected layer, output size = 1
\end{itemize}

\paragraph{Nonlinear Classifier $w$}
\begin{itemize}
\item Input layer: Input batch \textit{(batch size, input dimension)}
\item Layer 1: Fully connected layer, output size = 6, activation = ReLU
\item Output layer: Fully connected layer, output size = 1
\end{itemize}


\subsubsection{V-IRM GAME}

\paragraph{Linear Data Representation $\Phi$}
\begin{itemize}
\item Input layer: Input batch \textit{(batch size, input dimension)}
\item Output layer: Fully connected layer, output size = 2
\end{itemize}

\paragraph{Nonlinear Data Representation $\Phi$}
\begin{itemize}
\item Input layer: Input batch \textit{(batch size, input dimension)}
\item Layer 1: Fully connected layer, output size = 6, activation = ReLU
\item Output layer: Fully connected layer, output size = 2
\end{itemize}

\paragraph{Linear Classifier $w$}
\begin{itemize}
\item Input layer: Input batch \textit{(batch size, 2)}
\item Output layer: Fully connected layer, output size = 1
\end{itemize}

\paragraph{Nonlinear Classifier $w$}
\begin{itemize}
\item Input layer: Input batch \textit{(batch size, 2)}
\item Layer 1: Fully connected layer, output size = 6, activation = ReLU
\item Output layer: Fully connected layer, output size = 1
\end{itemize}


\subsubsection{iCaRL}

\paragraph{NF-iVAE $\boldsymbol{\lambda}_{f}$-Linear Prior} 
\begin{itemize}
\item Input layer: Input batch \textit{(batch size, input dimension)}
\item Output layer: Fully connected layer, output size = 2
\end{itemize}

\paragraph{NF-iVAE $\boldsymbol{\lambda}_{f}$-Nonlinear Prior} 
\begin{itemize}
\item Input layer: Input batch \textit{(batch size, input dimension)}
\item Layer 1: Fully connected layer, output size = 6, activation = ReLU
\item Output layer: Fully connected layer, output size = 2
\end{itemize}

\paragraph{NF-iVAE Linear Encoder}
\begin{itemize}
\item Input layer: Input batch \textit{(batch size, input dimension)}
\item Mean Output layer: Fully connected layer, output size = 2
\item Log Variance Output layer: Fully connected layer, output size = 2
\end{itemize}

\paragraph{NF-iVAE Nonlinear Encoder}
\begin{itemize}
\item Input layer: Input batch \textit{(batch size, input dimension)}
\item Layer 1: Fully connected layer, output size = 6, activation = ReLU
\item Mean Output layer: Fully connected layer, output size = 2
\item Log Variance Output layer: Fully connected layer, output size = 2
\end{itemize}

\paragraph{NF-iVAE Linear Decoder}
\begin{itemize}
\item Input layer: Input batch \textit{(batch size, 2)}
\item Mean Output layer: Fully connected layer, output size = output dimension
\item Variance Output layer: $0.01 \times \boldsymbol{1}$, where $\boldsymbol{1}$ is a vector full of $1$ with the length of output dimension
\end{itemize}

\paragraph{NF-iVAE Nonlinear Decoder}
\begin{itemize}
\item Input layer: Input batch \textit{(batch size, 2)}
\item Layer 1: Fully connected layer, output size = 6, activation = ReLU
\item Mean Output layer: Fully connected layer, output size = output dimension
\item Variance Output layer: $0.01 \times \boldsymbol{1}$, where $\boldsymbol{1}$ is a vector full of $1$ with the length of output dimension
\end{itemize}

\paragraph{Linear Classifier $w$}
\begin{itemize}
\item Input layer: Input batch \textit{(batch size, 1)}
\item Output layer: Fully connected layer, output size = 1
\end{itemize}

\paragraph{Nonlinear Classifier $w$}
\begin{itemize}
\item Input layer: Input batch \textit{(batch size, 1)}
\item Layer 1: Fully connected layer, output size = 6, activation = ReLU
\item Output layer: Fully connected layer, output size = 1
\end{itemize}


\subsection{CMNIST and CFMNIST}

Considering that the results of most baselines come from IRMG \citep{ahuja2020invariant}, for a fair comparison, we follow the same setting of IRMG in terms of hyper-parameters and validation considerations. For example, the batch size is set to 256, and the learning rate is $10^{-4}$. We also did not use the test environment data for validation. Please find more details in \citet{ahuja2020invariant}.

\paragraph{NF-iVAE $\boldsymbol{T}_{NN}$-Prior} 
\begin{itemize}
\item Input layer: Input batch \textit{(batch size, input dimension)}
\item Layer 1: Fully connected layer, output size = 50, activation = ReLU
\item Output layer: Fully connected layer, output size = 45
\end{itemize}

\paragraph{NF-iVAE $\boldsymbol{\lambda}_{NN}$-Prior} 
\begin{itemize}
\item Input layer: Input batch \textit{(batch size, input dimension)}
\item Layer 1: Fully connected layer, output size = 50, activation = ReLU
\item Output layer: Fully connected layer, output size = 45
\end{itemize}

\paragraph{NF-iVAE $\boldsymbol{\lambda}_{f}$-Prior} 
\begin{itemize}
\item Input layer: Input batch \textit{(batch size, input dimension)}
\item Layer 1: Fully connected layer, output size = 50, activation = ReLU
\item Output layer: Fully connected layer, output size = 20
\end{itemize}

\paragraph{NF-iVAE $\boldsymbol{O}$-Encoder} 
\begin{itemize}
\item Input layer: Input batch \textit{(batch size, 2, 28, 28)}
\item Layer 1: Convolutional layer, output channels = 32, kernel size = 3, stride = 2, padding = 1, activation = ReLU
\item Layer 2: Convolutional layer, output channels = 32, kernel size = 3, stride = 2, padding = 1, activation = ReLU
\item Layer 3: Convolutional layer, output channels = 32, kernel size = 3, stride = 2, padding = 1, activation = ReLU
\item Output layer: Flatten
\end{itemize}

\paragraph{NF-iVAE $(\boldsymbol{Y}, \boldsymbol{E})$-Encoder} 
\begin{itemize}
\item Input layer: Input batch \textit{(batch size, input dimension)}
\item Output layer: Fully connected layer, output size = 100, activation = ReLU
\end{itemize}

\paragraph{NF-iVAE $(\boldsymbol{O}, \boldsymbol{Y}, \boldsymbol{E})$-Merger/Encoder} 
\begin{itemize}
\item Input layer: Input batch \textit{(batch size, input dimension)}
\item Layer 1: Fully connected layer, output size = 100, activation = ReLU
\item Mean Output layer: Fully connected layer, output size = 10
\item Log Variance Output layer: Fully connected layer, output size = 10
\end{itemize}

\paragraph{NF-iVAE Decoder}
\begin{itemize}
\item Input layer: Input batch \textit{(batch size, 10)}
\item Layer 1: Fully connected layer, output size = $32 \times 4 \times 4$, activation = ReLU
\item Layer 2: Reshape to \textit{(batch size, 32, 4, 4)}
\item Layer 3: Deconvolutional layer, output channels = 32, kernel size = 3, stride = 2, padding = 1, outpadding = 0, activation = ReLU
\item Layer 4: Deconvolutional layer, output channels = 32, kernel size = 3, stride = 2, padding = 1, outpadding = 1, activation = ReLU
\item Layer 5: Deconvolutional layer, output channels = 2, kernel size = 3, stride = 2, padding = 1, outpadding = 1
\item Mean Output layer: activation = Sigmoid
\item Variance Output layer: $0.01 \times \boldsymbol{1}$, where $\boldsymbol{1}$ is a matrix full of $1$ with the size of $2 \times 28 \times 28$. 
\end{itemize}

\paragraph{Classifier $w$}
\begin{itemize}
\item Input layer: Input batch \textit{(batch size, 50)}
\item Layer 1: Fully connected layer, output size = 100, activation = ReLU
\item Output layer: Fully connected layer, output size = 1, activation = Sigmoid
\end{itemize}


\subsection{VLCS}

We used the exact experimental setting that is described in \citet{gulrajani2020search}. Specifically, we trained our model over all possible train and test environment combination for one of the commonly used hyper-parameter tuning procedure: train domain validation. We use ResNet-50 as an encoder and reverse the architecture of ResNet-50 as a decoder. We set the number of the latent variables to $n=50$. We do the hyperparameter search by exactly following the guides given in \citet{gulrajani2020search}.

\end{document}